\documentclass[journal,12pt,draftclsnofoot,onecolumn]{IEEEtran}
\usepackage{verbatim}
\usepackage{amsmath,graphicx}

\usepackage{mathrsfs}
\usepackage{amsopn}
\usepackage{cite}
\usepackage{amsthm}
\usepackage{epsfig,psfrag}
\usepackage{amsfonts,amssymb}
\usepackage{subfigure}
\usepackage{algorithm}
\usepackage{algorithmic}
\usepackage{graphics}
\usepackage{epsfig}
\usepackage{epstopdf}
\usepackage{amsbsy}
\usepackage{tabularx}
\usepackage{amssymb} 
\usepackage{multirow}
\usepackage{xcolor}
\usepackage{array}
\usepackage{arydshln}
\usepackage{color}
\usepackage{caption}
\usepackage{xcolor}
\usepackage{nicefrac}
\usepackage{hyperref}
\usepackage{cleveref}

\DeclareGraphicsExtensions{.pdf,.jpeg,.png}
\newcommand{\tabincell}[2]{\begin{tabular}{@{}#1@{}}#2\end{tabular}}

\begin{document}
\title{Multi-Focus Image Fusion using Sparse Representation and Coupled Dictionary Learning}
\author{Farshad~G.~Veshki 
and Sergiy~A.~Vorobyov, {\it Fellow, IEEE} 

\IEEEaftertitletext{\vspace{0\baselineskip}}
\thanks{
The authors are with Aalto University, Dept. Signal Processing and Acoustics, FI-00076, AALTO, Finland. E-mails: farshad.ghorbaniveshki@aalto.fi and svor@ieee.org}
}

\markboth{}
{Shell \MakeLowercase{\textit{et al.}}: Bare Demo of IEEEtran.cls
for Journals}
\maketitle
\begin{abstract}
We address the multi-focus image fusion problem, where multiple images captured with different focal settings are to be fused into an all-in-focus image of higher quality. Algorithms for this problem necessarily admit the source image characteristics along with focused and blurred features. However, most sparsity-based approaches use a single dictionary in focused feature space to describe multi-focus images, and ignore the representations in blurred feature space. We propose a multi-focus image fusion approach based on sparse representation using a coupled dictionary. It exploits the observations that the patches from a given training set can be sparsely represented by a couple of overcomplete dictionaries related to the focused and blurred categories of images and that a sparse approximation based on such coupled dictionary leads to a more flexible and therefore better fusion strategy than the one based on just selecting the sparsest representation in the original image estimate. In addition, to improve the fusion performance, we employ a coupled dictionary learning approach that enforces pairwise correlation between atoms of dictionaries learned to represent the focused and blurred feature spaces. We also discuss the advantages of the fusion approach based on coupled dictionary learning, and present efficient algorithms for fusion based on coupled dictionary learning. Extensive experimental comparisons with state-of-the-art multi-focus image fusion algorithms validate the effectiveness of the proposed approach. 
\end{abstract}


\IEEEpeerreviewmaketitle
\section{Introduction}
\label{Introduction}
\IEEEPARstart{O}ver the last several decades, considerable attention has been given to the multi-focus image fusion problem~\cite{1, 2, 3, 4, 5}. Multi-focus image fusion is an effective post-processing technique for combining multiple images captured with different focal distances into an all-in-focus image, without sacrificing image quality, and at the same time without using specialized optic sensors~\cite{6, 7, 8}. The problem is of high importance in many fields, ranging from remote sensing to medical imaging~\cite{9, 10, 11,12}, especially for addressing the demand for cost minimization of optical sensors/cameras.

Looking at recent approaches, sparsity and overcompleteness have been successfully used for computational image fusion~\cite{ 13, 14, 15, 16, 17, 18, 19}. The methods exploit the fact that patches of natural image can be compactly represented using an overcomplete dictionary as a linear combination of only few \emph{atoms}. It means that the vector of weighting coefficients for the atoms is sparse. 

Many image processing applications have benefited remarkably by using the above approach with a single learned overcomplete dictionary. A coupled dictionary-based approach~\cite{20, 21, 22, 23,24} has not been however used before for multi-focus image fusion.\footnote{Except for our preliminary conference contribution \cite{26}.} Indeed, the forenamed image fusion methods~\cite{13, 14, 15, 16, 17, 18} directly learn and exploit a single overcomplete dictionary in a single feature space in order to describe multiple images which contain both the focused and blurred categories of image features. Hence, these methods ignore the sparse representations in blurred feature space, and set limits on sparsity of the vector of coefficients. The latter consequently leads to a limited accuracy of fusing coefficients. These disadvantages, which are associated with a limitation of single feature space only, motivate us to perform fusion in double feature spaces. In particular, instead of learning a single overcomplete dictionary from the focused features only, this paper suggests to learn two dictionaries over focused and blurred feature spaces, and then use the pair of dictionaries to perform fusion via sparse representation in both spaces. In this way, we exploit the existing structure shared by all available multi-focus images, correlate the representations over double feature spaces, and improve the fusion performance. We use the method proposed in our previous work~\cite{25} for learning two correlated dictionaries representing focused and blurred feature spaces. Our approach, based on dictionaries learned from double feature spaces and fusion via sparse representations over both spaces proves to be more accurate than traditional methods based on single learned dictionary.

In this paper, we extend the coupled dictionary learning approach based on sparse and redundant representations to the problem of fusing multi-focus images. Coupled overcomplete dictionary is expected to lead to more compact representation of the focused and blurred categories of images. Then the weighted max-$l1$-norm strategy can be used while seeking for the focused patches needed to reconstruct an all-in-focus image. 

The paper presents both algorithmic developments and simulation results for multi-focus image fusion. We formulate the multi-focus image fusion problem as a problem of obtaining an all-in-focus image from multi-focus input images based on their sparse representations over a coupled dictionary (of focused and blurred dictionaries). Moreover, we develop a fusion procedure that finds an accurate decision map based on the sparse representations of multi-focus source images, which are obtained using a coupled dictionary found by our novel coupled dictionary learning algorithm. The all-in-focus image is then reconstructed using such decision map and the source image patches.

There are major differences between the approach in this paper and the conference paper \cite{26}. It includes the way of obtaining sparse representation. Averaging operator is used for fusing sparse coefficients in \cite{26}, while here it is omitted since a coupled dictionary is used for fusion instead of two dictionaries separately. Also, the all-in-focus image is reconstructed here using the exact patches from source images that are found focused by our selection operator instead of using their sparse representations as in \cite{26}.

We use bold capital letters for matrices, for example, we denote $k$-th multi-focus and all-in-focus images as the following matrices of pixels $\boldsymbol{I}_k$ and $\boldsymbol{I}^{\rm F}$, respectively. Similarly, the matrices $\boldsymbol{D}^{\rm F}$ and $\boldsymbol{D}^{\rm B}$ denote the focus and blurred dictionaries, respectively. The operators and sets are denoted using calligraphic letters, for instance, $\mathcal{F}$ represents the fusion operator. 

Images are processed by patches.\footnote{It is because adapting a dictionary to large size images is impractical.} Image patches are extracted using the sliding window technique (moving through images starting from the left-top corner to the right-bottom corner). Then the image patch of the size $d \times d$ pixels e.g., $\boldsymbol{I}^{\rm F}_{i,j} \in \mathbb{R}^{d \times d}$ is ordered lexicographically as a column vector and denoted as $\boldsymbol{i}^{\rm F}_{i,j} \in \mathbb{R}^{d^2}$. For notation simplicity and without loss of generality, we drop hereafter the indices $(i,j)$ marking the patch position in an image, and we denote patches of $k$-th multi-focus image $\boldsymbol{I}_k$ and all-in-focus image $\boldsymbol{I}^{\rm F}$ just as $\boldsymbol{i}_k \in \mathbb{R}^{d^2}$ and $\boldsymbol{i}^{\rm F} \in \mathbb{R}^{d^2}$, respectively.

The vector norm $\|\cdot\|_p$ for $p\geq 1$ is the standard \mbox{$l_p$-norm}, and $\|\cdot\|_0$ denotes the operator that counts the number of non-zero entries in a vector. For a matrix $\boldsymbol{X}$, we define the Frobenius norm as $\|\boldsymbol{X}\|_{\rm F} = ( \sum_{i,j} ([\boldsymbol{X}]_{i,j})^2 )^{\frac{1}{2}}$. The symbol $\odot$ represents element-wise product of matrices, $\nabla(\cdot)$ denotes the forward finite difference operator on the vertical and horizontal directions, and $(\cdot)^T$ stands for the transpose operation.

The remainder of the paper is organized as follows. Section~\ref{Problem} gives the problem description and summarizes some general assumptions and a solution approach. 
Section~\ref{fusion} gives a detailed explanation of our fusion procedure. Simulation results are provided in Section~\ref{EXPERIMENTS}. Finally, we conclude the paper in Section~\ref{Conclusion}.

\section{Problem Description and Solution Approach}
\label{Problem}
Consider the problem of constructing/reconstructing a high quality all-in-focus image $\boldsymbol{I}^{\rm F}$ from a set of multi-focus source images $\left\{ \boldsymbol{I}_k \right\}_{k=1}^K$, which can be abstractly written in the form of the following fusing process
\begin{equation}\label{eq0000}
\begin{split}
    \begin{aligned}
    \boldsymbol{I}^{\rm F} = \mathcal{F} \left\{ \boldsymbol{I}_k \right\}_{k=1}^K + \boldsymbol{V}
        \end{aligned}
    \end{split}
\end{equation}
where $\mathcal{F} \{ \cdot \}$ stands for a fusing operator and $\boldsymbol{V}$ is the noise. The fusion goal is to obtain $\boldsymbol{I}^{\rm F}$ from $\left\{\boldsymbol{I}_k\right\}_{k=1}^K$. It is assumed here for simplicity that each multi-focus image $\boldsymbol{I}_k$ is captured for the same scene and all multi-focus images are properly aligned.\footnote{Note that this assumption is typical in the literature with a focus on image fusion algorithms design, but proper alignment of images is also an important practical issue.} One more typical assumption is that at each position, one of the input image patches is (the most) in-focus. Hence, the problem is to find the most focused patch in a set of all available corresponding patches, extracted from the multi-focus inputs. 
\begin{figure}[!htb]
	\centerline{\includegraphics[width=0.57\textwidth]{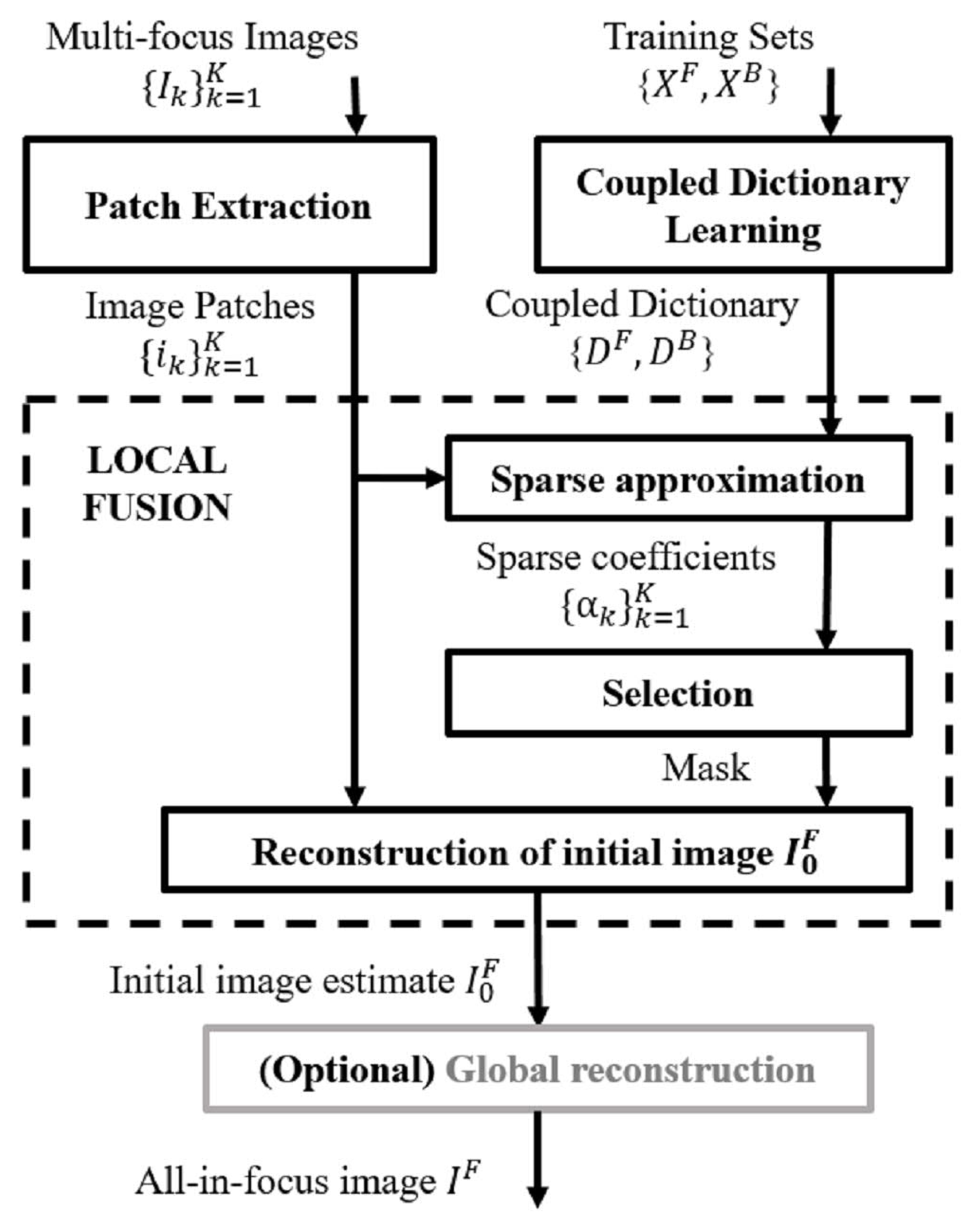}}
	\centering
	\caption{Block-diagram of the procedure for reconstructing an all-in-focus image $\boldsymbol{I}^{\rm F}$ from the given set of multi-focus images $\left\{\boldsymbol{I}_k\right\}_{k=1}^K$ via coupled sparse representations in focus and blurred spaces and coupled dictionary learning.}
	\label{fig:framework}
\end{figure}

As a solution approach, Fig.~\ref{fig:framework} shows the block-diagram of the proposed procedure for constructing/reconstructing an all-in-focus image $\boldsymbol{I}^{\rm F}$ from a given set of multi-focus images $\left\{\boldsymbol{I}_k\right\}_{k=1}^K$. In this block-diagram, the block ``Patch extraction'' represents the above described simple process of extracting patches $\{\boldsymbol{i}_k\}_{k=1}^K$ from the available multi-focus images $\left\{\boldsymbol{I}_k\right\}_{k=1}^K$. Due to the fact that visible artifacts may occur on patch boundaries, overlapping patches that include pixels of neighboring patches are typically used to suppress such artifacts. The input to the block ``Coupled dictionary learning'' is a set of two available subsets of training patches, i.e., $\boldsymbol{X} \triangleq \{ \boldsymbol{X}^{\rm F}, \ \boldsymbol{X}^{\rm B} \}$. Here $\boldsymbol{X}^{\rm F}$ and $\boldsymbol{X}^B $ are the subsets of manually labeled (classified) in-focus and blurred training patches, which are extracted from corresponding parts of available multi-focus images.\footnote{Note that the method proposed in the paper can be straightforwardly extended also to the case when image patches are labeled at different focal depths to more than two sets of patches, but then the main cost will be labeling the patches.} 

The output of the block ``Coupled dictionary learning'' is the coupled dictionary $\boldsymbol{D} \triangleq \left[ \boldsymbol{D}^{\rm F}, \ \boldsymbol{D}^{\rm B} \right]$ representing the in-focus and blurred image feature spaces. The method proposed in our previous work~\cite{25} for learning two correlated dictionaries representing focused and blurred feature spaces suites here perfectly. Indeed, the existing coupled dictionary learning algorithms have been developed for solving another inverse problem -- super resolution, where the objective is to find an accurate transform from blurred inputs to unknown focused image patches. 

The procedure of fusing multi-focus images is presented in Fig.~\ref{fig:framework} by three sub-blocks. The sub-block ``Sparse approximation'' represents a procedure of finding sparse representations $\{\boldsymbol{\alpha}_k\}_{k=1}^K$ of image patches $\{\boldsymbol{i}_k\}_{k=1}^K$ over the coupled dictionary $\boldsymbol{D}$. The selection operation, represented by sub-block ``Selection'', finds the focused patches and mask (a matrix containing the indix of in-focus patch $k^F$ at each position). Having the image patches and the mask, the in-focus patches are found, then the initial all-in-focus image is constructed, that is shown by the sub-block ``Reconstruction of initial image $\boldsymbol{I}^{\rm F}_0$''.

The averaging nature of such all-in-focus image reconstruction from its overlapping patches may cause some blurring, mostly around the edges, where each of the source images is focused on one side and is blurred on the other side. Indeed, the patches that cross the focus boundaries and scattered noise in the decision map introduce some blurredness around those edges and fade the small details. Thus, a global reconstruction may need to be performed optionally in order to restore the contrast resolution of the initial estimation~$\boldsymbol{I}^{\rm F}_0$. 

\section{Fusion via Coupled Sparse Representation}
\label{fusion}
Using the coupled dictionary learning method in \cite{25} the two dictionaries $\boldsymbol{D}^{\rm F}$ and $\boldsymbol{D}^{\rm B}$, which are better representatives for, respectively, the focused and blurred feature spaces, can be obtained. Considering the characteristics of learned dictionaries $\boldsymbol{D}^{\rm F}$ and $\boldsymbol{D}^{\rm B}$, we can infer that for the residuals $\boldsymbol{r}^F$ and $\boldsymbol{r}^B$, representing a pair of corresponding focused and blurred features, the inequalities 

\begin{equation}\label{eq9901}
\begin{split}
\begin{aligned}
  \left\{
\begin{array}{l}
\underset{ \boldsymbol{d}^F}{\mathrm{max}} 
\mid (\boldsymbol{d}^F)^T \cdot \boldsymbol{r}^F \mid \geq 
\underset{ \boldsymbol{d}^B}{\mathrm{max}} 
\mid (\boldsymbol{d}^B)^T\cdot \boldsymbol{r}^F \mid\\
\underset{ \boldsymbol{d}^B}{\mathrm{max}} 
\mid (\boldsymbol{d}^B)^T \cdot \boldsymbol{r}^B \mid \geq 
\underset{\boldsymbol{d}^F}{\mathrm{max}} 
\mid (\boldsymbol{d}^F)^T\cdot \boldsymbol{r}^B \mid,\\
    \end{array}
\right.
\\
\boldsymbol{d}^F \in \boldsymbol{D}^{\rm F} , \boldsymbol{d}^B \in \boldsymbol{D}^{\rm B}\qquad \qquad \qquad
\end{aligned}
\end{split}
\end{equation}
hold. We next propose a greedy method to find the sparse representations $\left\{ \boldsymbol{\alpha}_k \right\}_{k=1}^K$ representing corresponding patches $\left\{ \boldsymbol{i}_k \right\}_{k=1}^K$ over the coupled dictionary $\boldsymbol{D}$ (horizontal concatenation of $\boldsymbol{D}^{\rm F}$ and $\boldsymbol{D}^{\rm B}$), knowing that the greedy methods solve the problem
\begin{equation}\label{eq9902}
\begin{split}
\begin{aligned}	
{{\boldsymbol{d}}_i}= \mathrm{arg}\underset{ {\boldsymbol{d}}_t}{\mathrm{max}} 
\mid {\boldsymbol{d}}_t^T\cdot \boldsymbol{r}_i \mid
; \quad \boldsymbol{d}_t \in \boldsymbol{D} \triangleq \left[ \boldsymbol{D}^{\rm F} \ \boldsymbol{D}^{\rm B} \right]
\end{aligned}
\end{split}
\end{equation}
to find the best matching atom ${{\boldsymbol{d}}_i}$ approximating the residual ${{\boldsymbol{r}}_i}$ at iteration $i$. Then based on \eqref{eq9901} and \eqref{eq9902}, we can instantly deduce that
\begin{equation}\label{eq9903}
\begin{split}
\begin{aligned}
\left\{
\begin{array}{l}
\boldsymbol{r}_i \text{ is focused},\qquad \text{if } \boldsymbol{d}_i \in \boldsymbol{D}^{\rm F}\\
\boldsymbol{r}_i \text{ is blurred},\qquad\, \text{if }  \boldsymbol{d}_i \in \boldsymbol{D}^{\rm B}.\\
\end{array} \ 
\right.  		
\end{aligned}
\end{split}
\end{equation}

In this section, a fusion approach based on \eqref{eq9903} and using a coupled learned dictionary is proposed. 

\subsection{Proposed Local Fusion Method}
\label{localfusion}

The local fusion operation as presented in Fig.~\ref{fig:framework} consists of three operations, namely, the sparse approximation, selection, and mask generation. Then the proposed local fusion approach can be mathematically formulated as 
\begin{equation}\label{eq02}
\begin{aligned}
{\boldsymbol{i}}^{\rm F} &\triangleq \mathcal{F} \left\{ \left\{\boldsymbol{i}_k \right\}_{k=1}^K \right\}\\ &= \mathcal{M} \left\{ \mathcal{R} \left\{ \left\{ \mathcal{L} \left\{ \boldsymbol{\alpha}_k;\boldsymbol{i}_k , \boldsymbol{D}\right\} \right\}_{k=1}^K \right\}\right\} \cdot \left\{\boldsymbol{i}_k \right\}_{k=1}^K \\
\end{aligned}
\end{equation}
where $\mathcal{M}\{ \cdot \}$ is the mask operator, $\mathcal{R}\{ \cdot \}$ is the selection operator,  and $\mathcal{L} \{ \cdot \}$ is the sparse approximation operator.

The operator $\mathcal{L} \{ \cdot \}$ can be formulated as the following $l_1$-norm minimization problem 
\begin{equation}\label{eq01}    
\begin{split}
\begin{aligned}
\mathcal{L} \left\{ \boldsymbol{\alpha}_k;\boldsymbol{i}_k , \boldsymbol{D}\right\} = &\mathrm{arg} \underset{{\boldsymbol{\alpha}_k}} {\mathrm{min}} \left\| \boldsymbol{\alpha}_k \right \|_1  \\
\mathrm{s.t.}\  &\left \| \boldsymbol{D} \cdot   \boldsymbol{\alpha}_k  - \boldsymbol{i}_k \right\|_2^2 \leq \epsilon .
\end{aligned}
\end{split}
\end{equation}
Note that problem~\eqref{eq01} can be efficiently solved by many existing greedy methods, e.g., conventional orthogonal matching pursuit (OMP) algorithm~\cite{41}.

To deal with the effect of luminance, the mean intensities are removed from all image patches before sparse approximation. Moreover, all vectors in $\left\{\boldsymbol{i}_k \right\}_{k=1}^K$, $\boldsymbol{D}^{\rm F}$ and $\boldsymbol{D}^{\rm B}$ are normalized to have Frobenius norm of one. After obtaining each vector of sparse representation $\boldsymbol{\alpha}_k$ for the corresponding patch $\boldsymbol{i}_k$, the selection operator $\mathcal{R} \left\{ \{ \boldsymbol{\alpha}_k \}_{k=1}^K \right\}$ is applied in accordance with \eqref{eq02} to find the sparse vector of coefficients that represents the most focused image patch $\boldsymbol{i}^{\rm F}$.

According to the max-$l_1$-norm rule~\cite{16}, the most focused sparse vector of coefficients $\boldsymbol{\alpha}^{\rm F}$ from the set of vectors $\{\boldsymbol{\alpha}_k\}_{k=1}^K$ is the vector with the highest activity level, where the activity level is measured by $l_1$-norm. In the coupled dictionary framework, based on \eqref{eq9903}, we extend the max-$l_1$-norm rule to the following weighted $l_1$-norm, where the weight for the $l_1$-norm of coefficients that correspond to $\boldsymbol{D}^{\rm F}$ is greater than the weight for the $l_1$-norm of coefficients that correspond to $\boldsymbol{D}^{\rm B}$. Thus, the selection operator finds the focused patch and its corresponding index $k^F$ by solving the following problem
\begin{equation} \label{eq014}	
\begin{split}
\begin{aligned}
k^F&=\mathcal{R} \left\{ \left\{ \boldsymbol{\alpha}_k\right\}_{k=1}^K \right\}\\
&=\underset{{k}}{\mathrm{argmax}}\! \left\{{\omega}.\left\|{\boldsymbol{\alpha}_k^F} \right\|_{\rm 1}+ \boldsymbol(1-{\omega}).\left\|{\boldsymbol{\alpha}_k^B}\right \|_{\rm 1}\right\}\\
\end{aligned}
\end{split}
\end{equation}
where $\omega$ and $1-\omega$ are the weights corresponding to the focused and blurred subspaces, respectively, $0.5\!\leqslant\!{\omega}\!<\!\rm 1 $, and ${\boldsymbol{\alpha}_k^F}$ and ${\boldsymbol{\alpha}_k^B}$ denote two segments of ${\boldsymbol{\alpha}_k}$ that correspond to ${\boldsymbol{D}^{\rm F}}$ and ${\boldsymbol{D}^{\rm B}}$.\footnote{Note that for the cases when there are more than two learned dictionaries, the proposed selection operator can be extended by segmenting $\boldsymbol{\alpha}_k$  based on the number of dictionaries and weighting those segments proportional to the focus level of dictionaries.} 

Then, the all-in-focus patch ${\boldsymbol{i}}^{\rm F}$ is found by applying the linear mask operator $\mathcal{M}$ as follows
\begin{equation} \label{allfuse}
\begin{split}
\begin{aligned}
\boldsymbol{i}^{\rm F}=
\mathcal{M} \left(  k^F, \{\boldsymbol{i}_k \}_{k=1}^K    \right)= 
\boldsymbol{i}_{k^F }.       
\end{aligned}
\end{split}
\end{equation}

Going across the whole image, all in-focus image patches ${\boldsymbol{i}}^{\rm F}$ are found separately one by one, then by placing them at their positions and averaging the overlapping pixels, the initial approximation of all-in-focus image ${\boldsymbol{I}}_0^{\rm F}$ is reconstructed. 
\subsection{Global Reconstruction}
A global reconstruction, as optionally suggested at the end of Section~\ref{Problem}, can be achieved by applying the total variation (TV) regularization, which is commonly used in the natural image analysis. Thus, applying the TV prior on the image gradients magnitude, the global reconstruction problem can be written as
\begin{equation}\label{eq016}
\begin{split}
\begin{aligned}
\boldsymbol{I}^{\rm F}\!=\!\mathrm{arg} \underset{\boldsymbol{I}^{\rm F}}{\mathrm{min}} \frac{1}{2} \left \| \boldsymbol{I}^{\rm F} - \boldsymbol{I}^{\rm F}_0 \right\|_2^2 + \eta \sum_{i,j} \rho \left( \left \| [ \nabla\boldsymbol{I}^{\rm F} ]_{i,j} \right\|_2 \right)
\end{aligned}
\end{split}
\end{equation}
where $\rho(\cdot)$ takes the form of TV, $[ \nabla\boldsymbol{I}^{\rm F} ]_{i,j}$ denotes the discretization of the gradient for $(i,j)$-th element, defined as 
$\left \| [ \nabla\boldsymbol{I}^{\rm F} ]_{i,j} \right\|_2 = \sqrt{(D_h \nabla \boldsymbol{I}^{\rm F})_{i,j}^2 + (D_v \nabla \boldsymbol{I}^{\rm F})_{i,j}^2}$
with linear operators $D_h$ and $D_v$ representing finite difference approximations of the first-order horizontal and vertical partial derivatives~\cite{30}. Similar to the approaches in~\cite{30,31,32}, optimization problem~\eqref{eq016} can be efficiently solved by the alternating directions method of multipliers (ADMM)~\cite{33,35}, which decomposes a large scale global problem into a series of smaller local subproblems. The resulting $\boldsymbol{I}^{\rm F}$ after applying global reconstruction is then taken as the final estimate of the all-in-focus image.

\subsection{Algorithm of Fusion via Coupled Sparse Representation}
In summary, when the underlying dictionaries $\boldsymbol{D}^{\rm F}$ and $\boldsymbol{D}^{\rm B}$ are known, the fusion via local sparse representations is first calculated. Then the global reconstruction is employed for enhancing the contrast resolution of the reconstructed all-in-focus image. The overall algorithm for multi-focus image fusion via sparse representation is summarized as Algorithm~\ref{alg:Steps}.

\begin{algorithm}[htb]    
	\renewcommand{\algorithmicrequire}{\textbf{Input:}}
	\renewcommand{\algorithmicensure}{\textbf{Output:}} 
	\caption{Image fusion via sparse representation.}   
	\label{alg:Steps}      
	\begin{algorithmic}[1]    
		\REQUIRE                
               
Multi-focus source images $\left\{\boldsymbol{ I}_k\right\}_{k=1}^K$ and learned coupled dictionary $\boldsymbol{D} = [\boldsymbol{D}^{\rm F}, \boldsymbol{D}^{\rm B}]  $.\	
\STATE  Obtain vectorized image patches $\left\{\boldsymbol{ i}_k\right\}_{k=1}^K$;\\	
\STATE  Remove the mean intensities and normalize all image patches as:\\
$\hat{\boldsymbol{ i}}_k=\boldsymbol{i}_k-\textrm{mean} \left( \boldsymbol{ i}_k \right)$, $\hat{\boldsymbol{ i}}_k=\dfrac{\hat{\boldsymbol{ i}}_k}{\|\hat{\boldsymbol{ i}}_k\|_F}$, $\forall k$;\\				
\STATE \textbf{for} each set of $\left\{\hat{\boldsymbol{ i}}_k\right\}_{k=1}^K$:\\		
\STATE  \quad Find $\left\{\boldsymbol{ \alpha}_k\right\}_{k=1}^K$ by solving \eqref{eq01}; \\
\STATE   \quad Find $k^F$ by applying the selection operator \eqref{eq014};\\
\STATE  \quad  Find the in-focus patch $\boldsymbol{i}^{\rm F}$ using \eqref{allfuse};\\	
\STATE \textbf{end for}		
\STATE Form the initial estimate of all-in-focus image ${\boldsymbol{ I}_0^F}$;
\STATE (Optional) Perform the global reconstruction using \eqref{eq016}.
\ENSURE The all-in-focus image  $\boldsymbol{I}^{\rm F}$. \
	\end{algorithmic}
\end{algorithm}

\section{Experimental Results}
\label{EXPERIMENTS}
\subsection{Experimental Setup}\label{Exp}
In this section, we evaluate the proposed approach for image fusion and compare it to some existing state-of-the-art approaches in terms of visual and quantitative comparison. Next, we discuss also various factors that influence the performance. 

The quantitative assessments are based on two non-reference-based image fusion performance metrics, namely, normalized mutual information (NMI) and $Q_{A\!B/\!F}$~\cite{37}, and two reference-based metrics, namely, structural similarity index (SSIM)~\cite{39} and mean square error (MSE).

\begin{figure}[!htb]
	\begin{center}
		\begin{minipage}{0.2\linewidth}
			\centerline{\includegraphics[width=3.5cm,height=3.5cm]{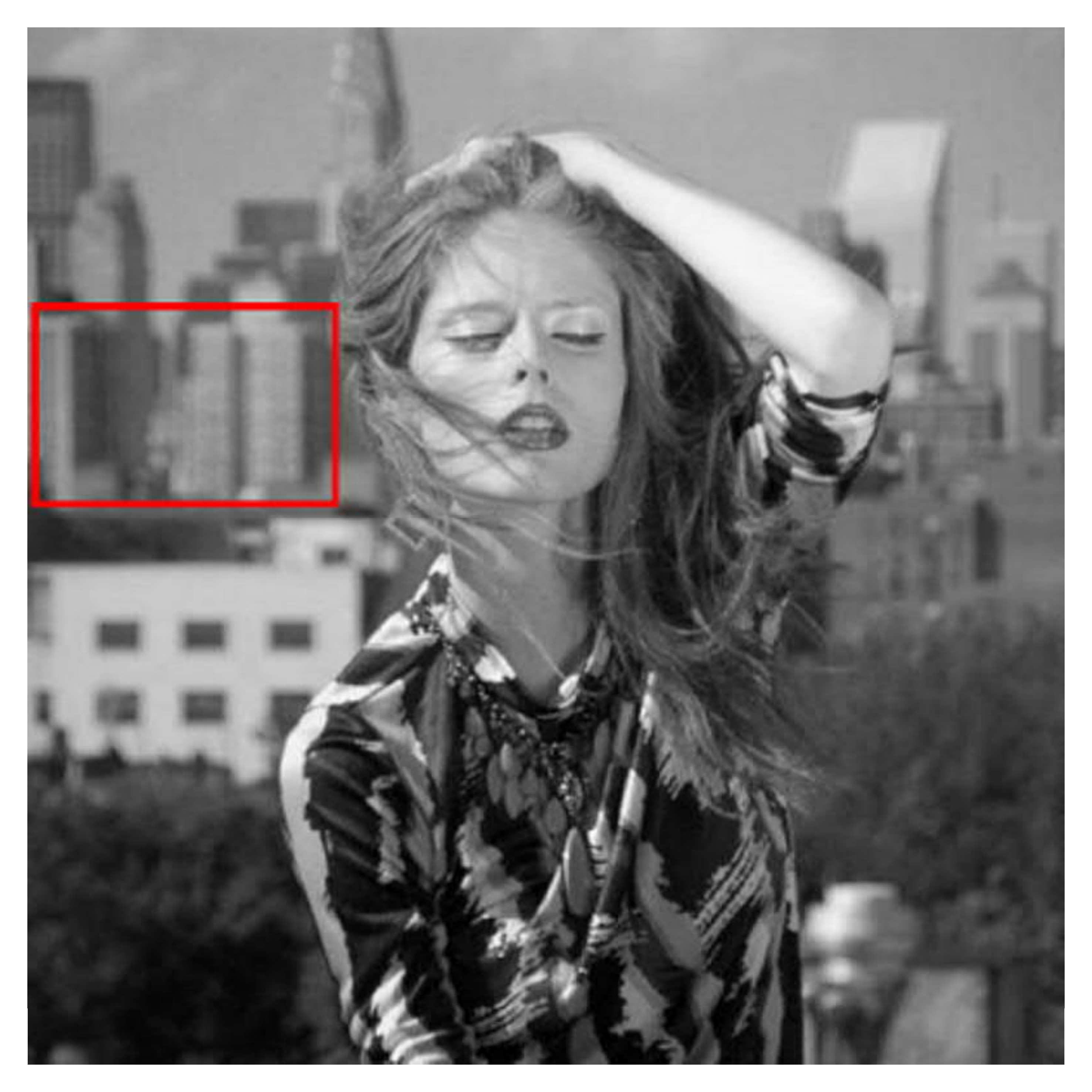}}	
		\end{minipage}%
		\;
		\begin{minipage}{0.2\linewidth}
			\centerline{\includegraphics[width=3.5cm,height=3.5cm]{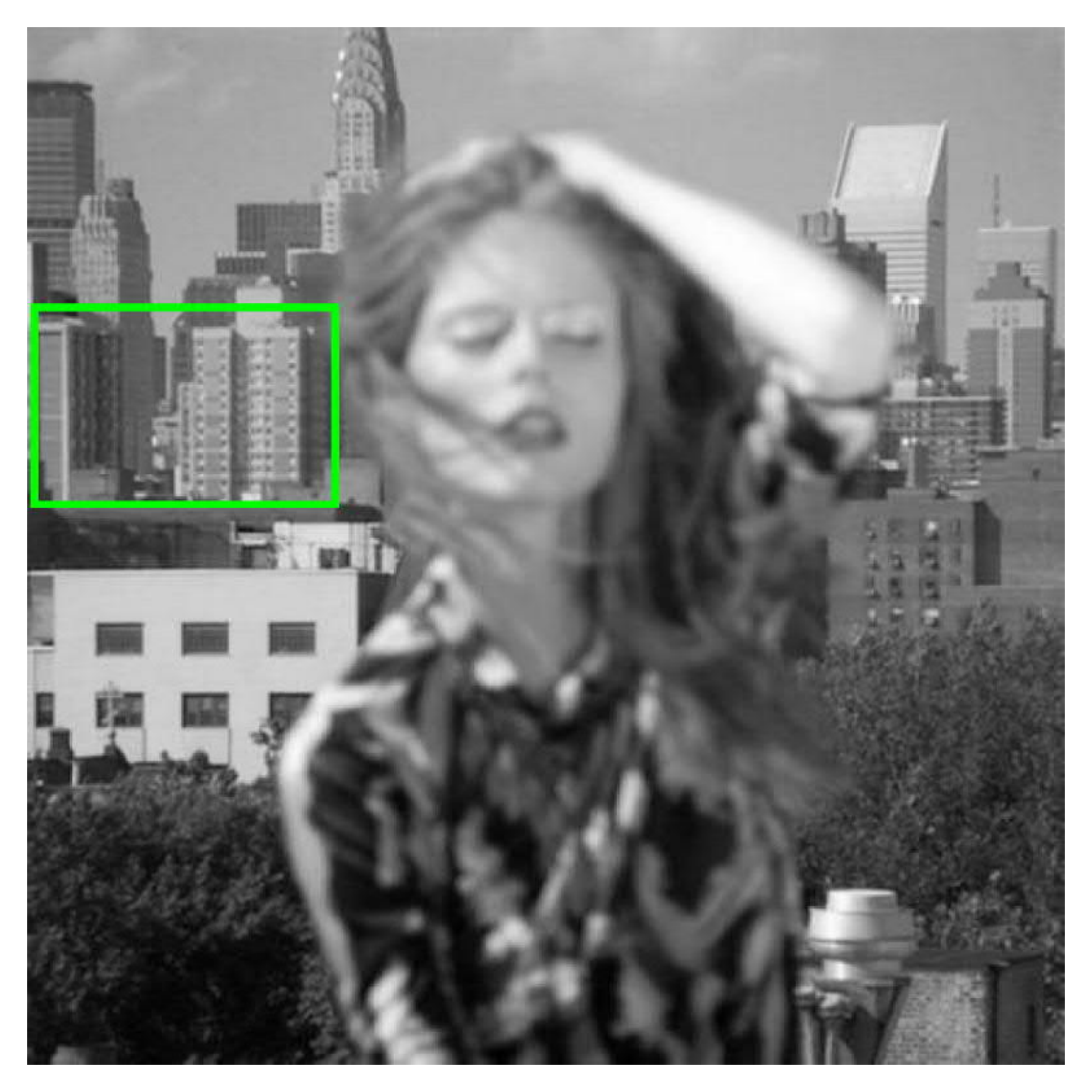}}
		\end{minipage}%
		\quad
		\begin{minipage}{0.2\linewidth}
			\centerline{\includegraphics[width=3.5cm,height=3.5cm]{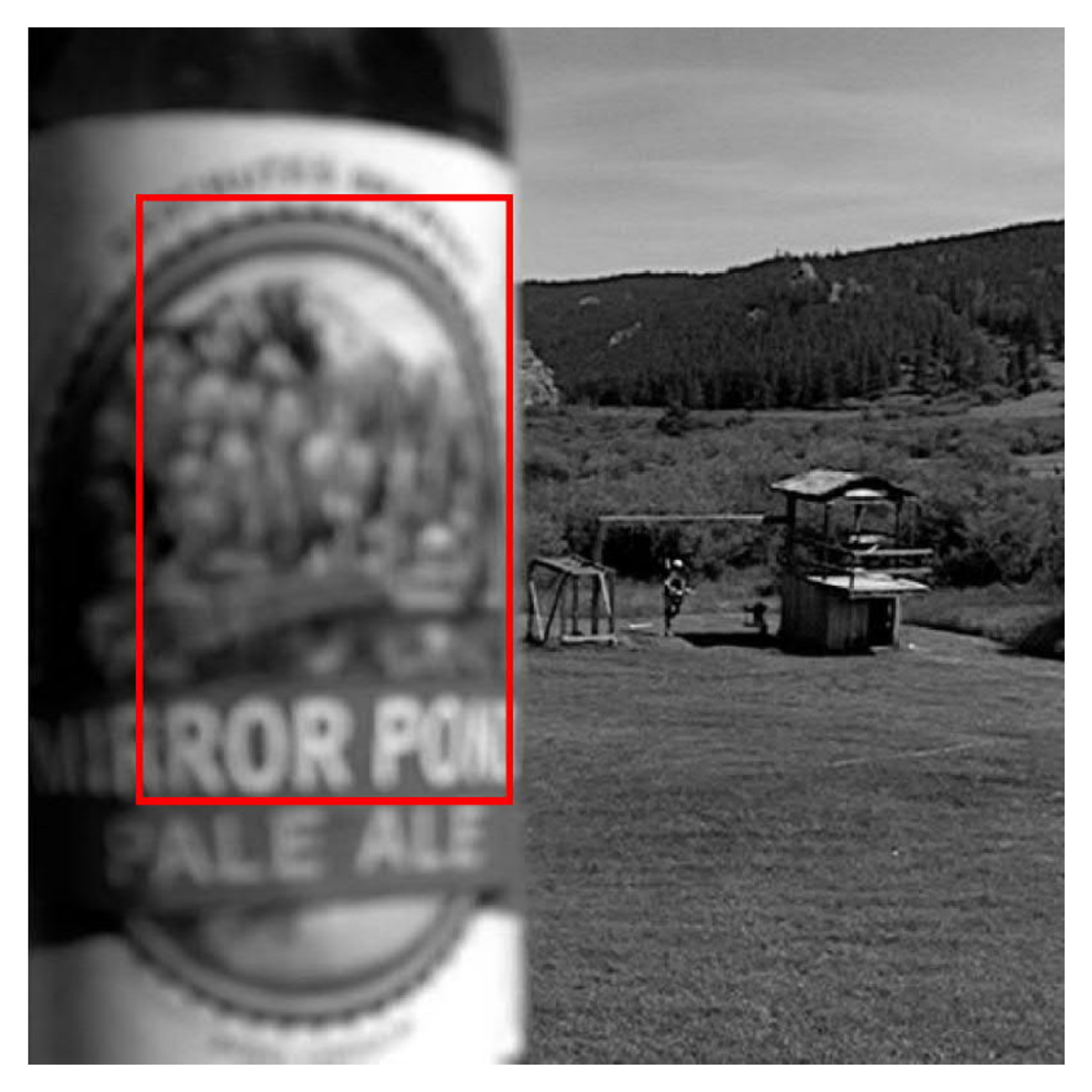}}	
		\end{minipage}%
		\;
		\begin{minipage}{0.2\linewidth}
			\centerline{\includegraphics[width=3.5cm,height=3.5cm]{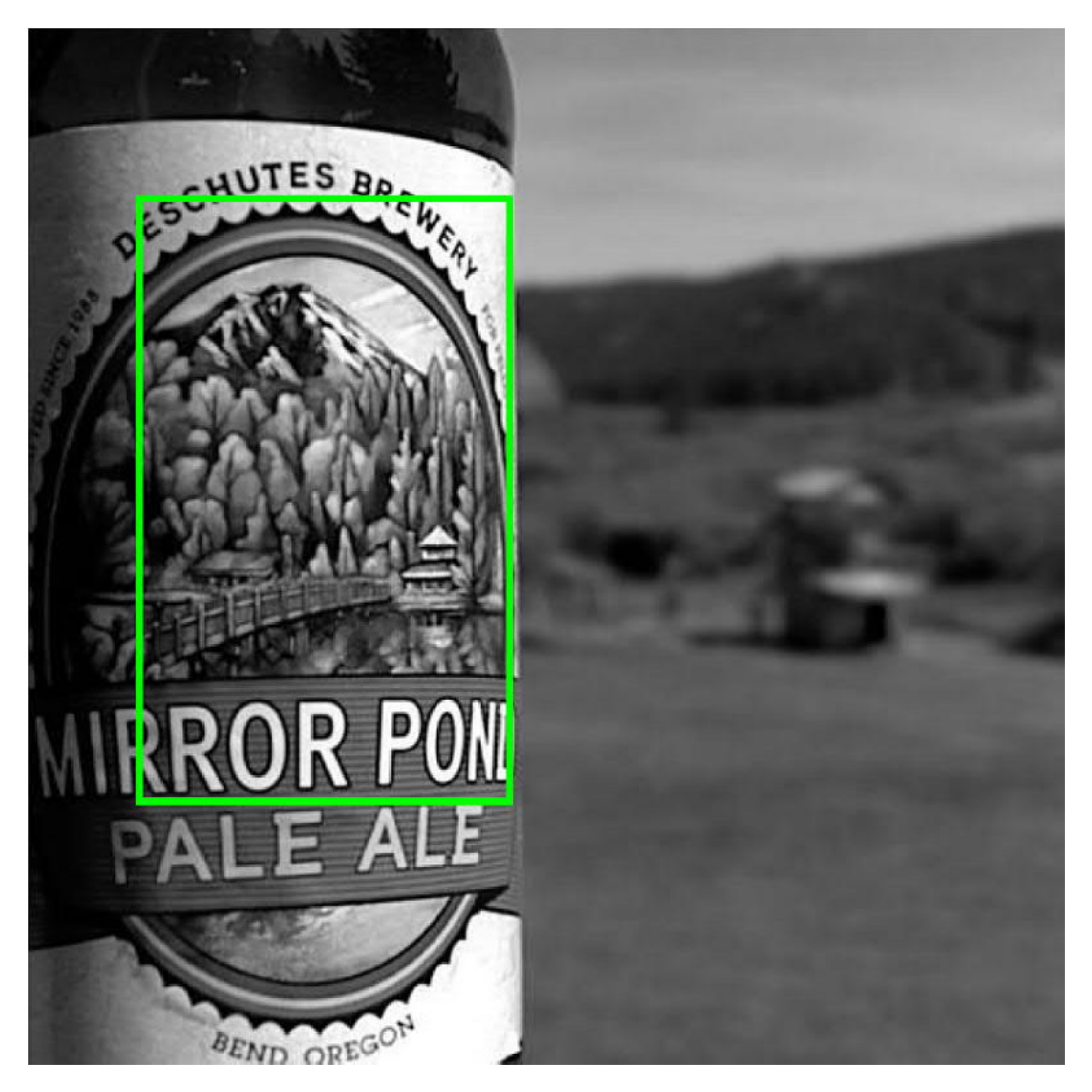}}
		\end{minipage}%
		\vfill
		\medskip
		\begin{minipage}{0.2\linewidth}
			\centerline{\includegraphics[width=3.5cm,height=3.5cm]{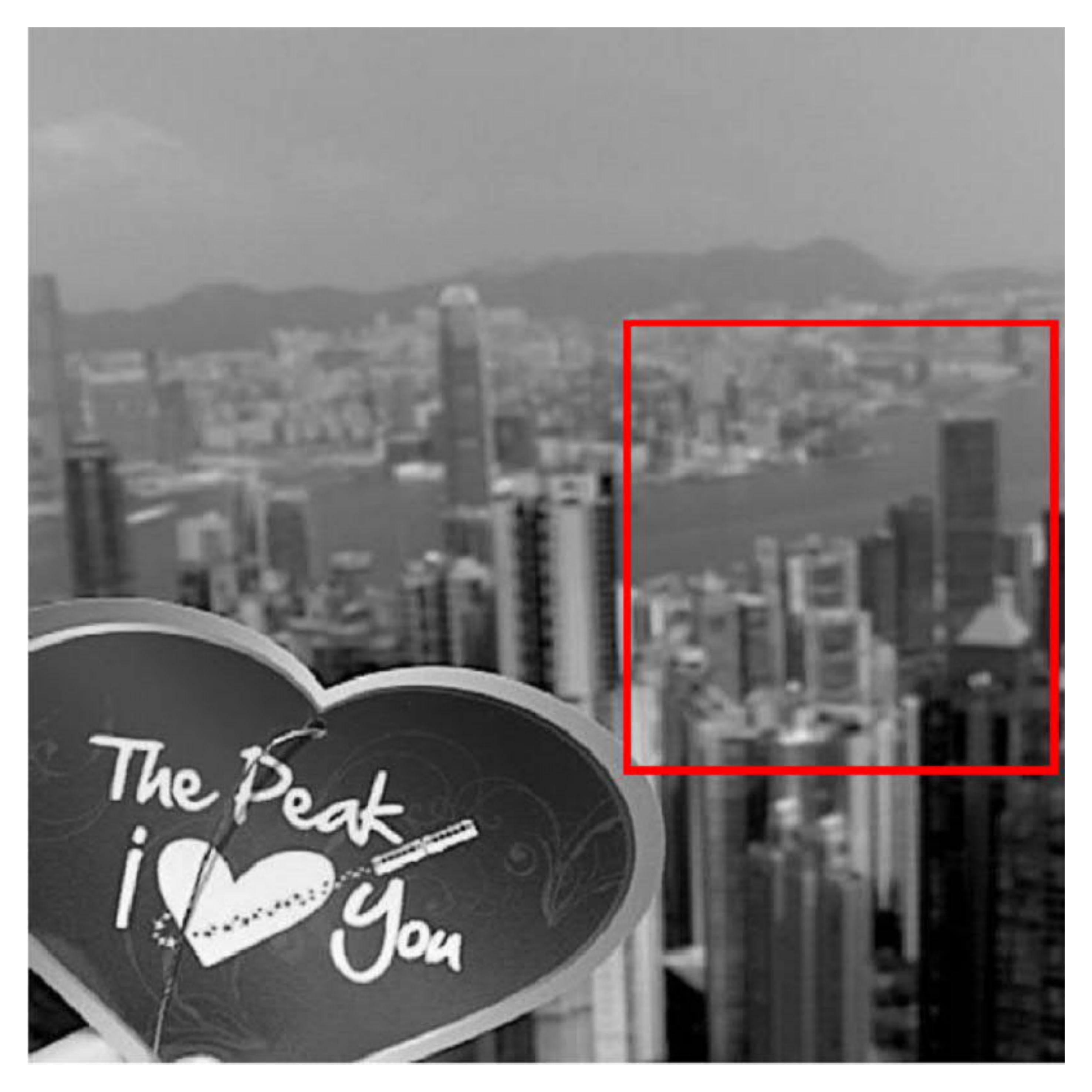}}
			
		\end{minipage}%
		\;
		\begin{minipage}{0.2\linewidth}
			\centerline{\includegraphics[width=3.5cm,height=3.5cm]{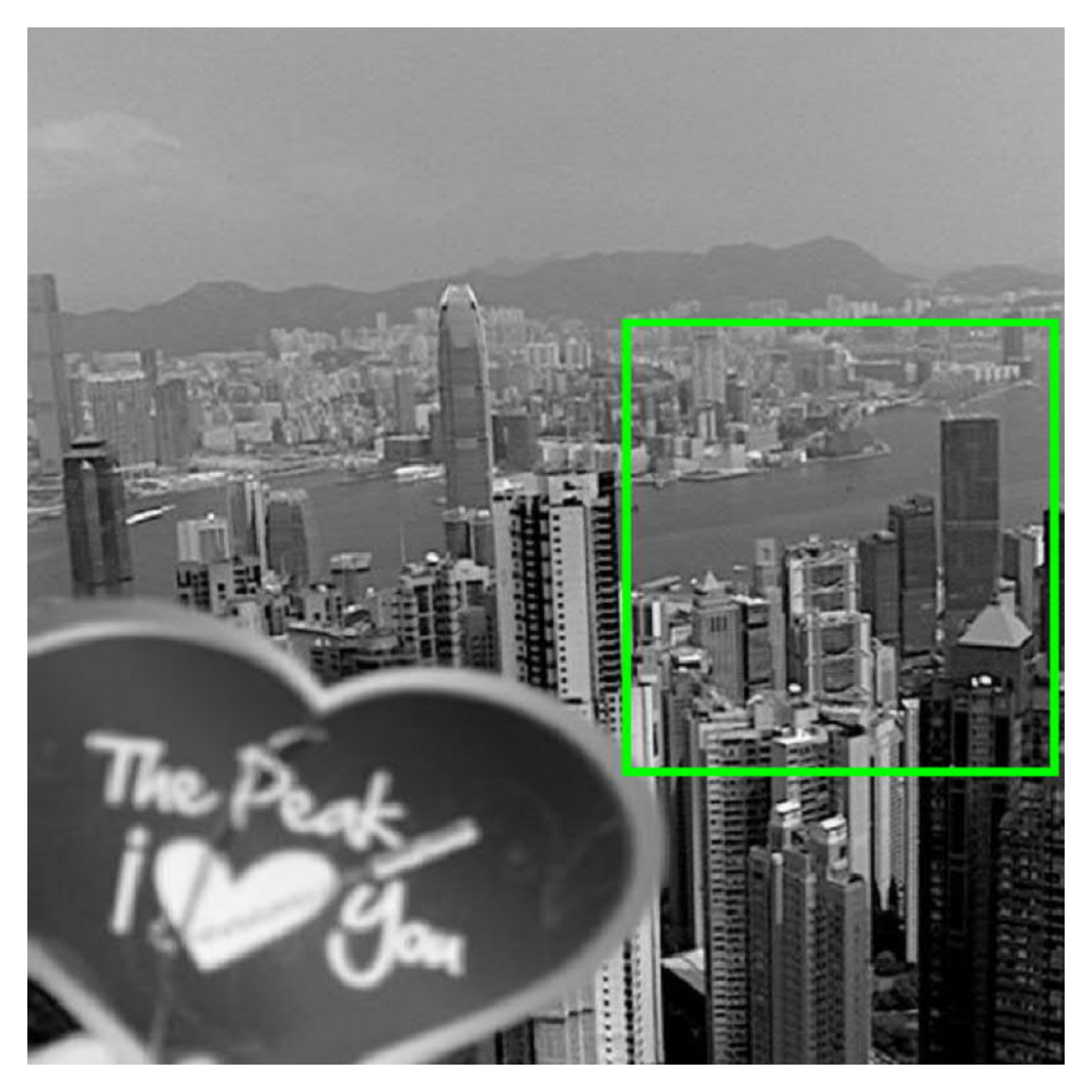}}
		\end{minipage}%
		\quad
		\begin{minipage}{0.2\linewidth}
			\centerline{\includegraphics[width=3.5cm,height=3.5cm]{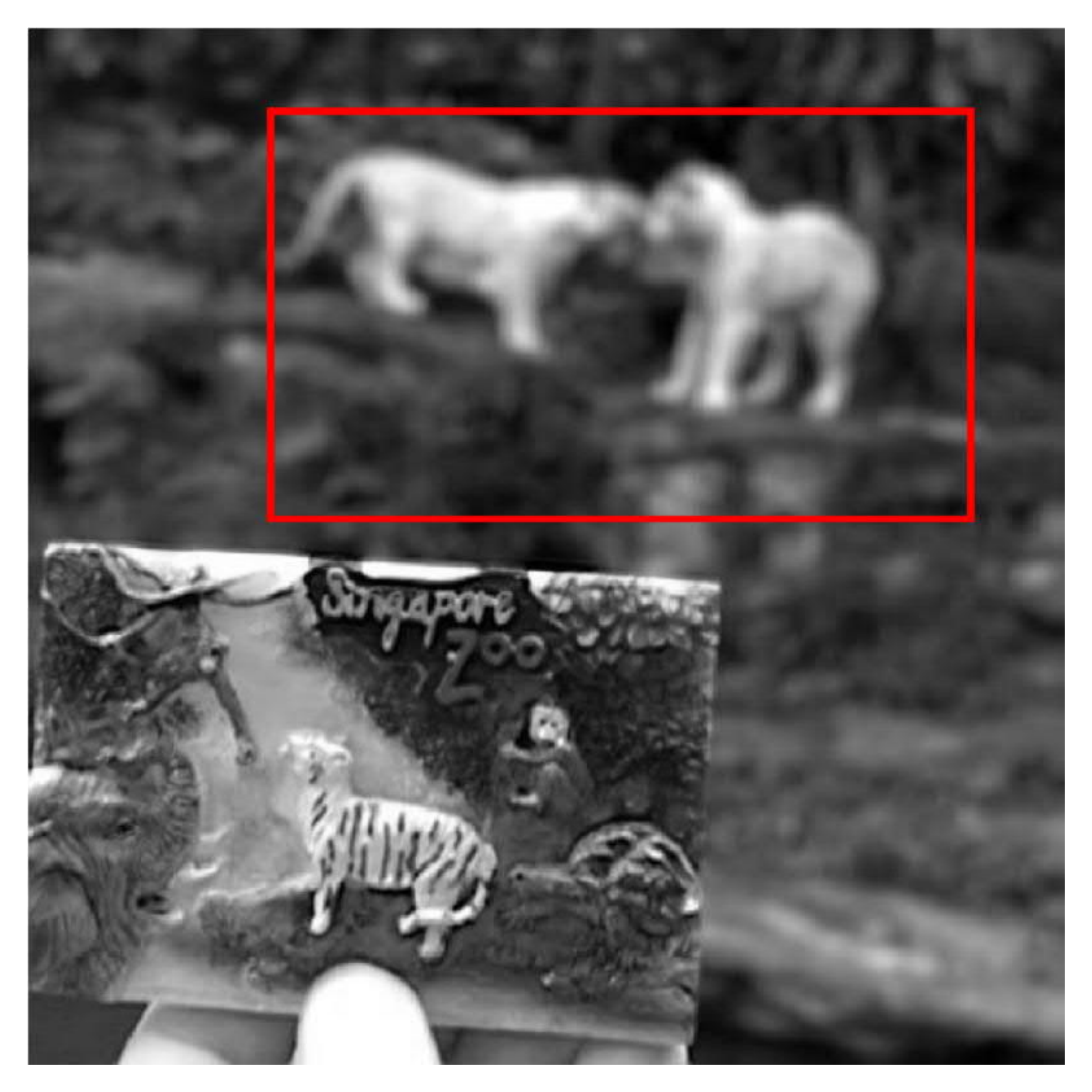}}
		\end{minipage}%
		\;
		\begin{minipage}{0.2\linewidth}
			\centerline{\includegraphics[width=3.5cm,height=3.5cm]{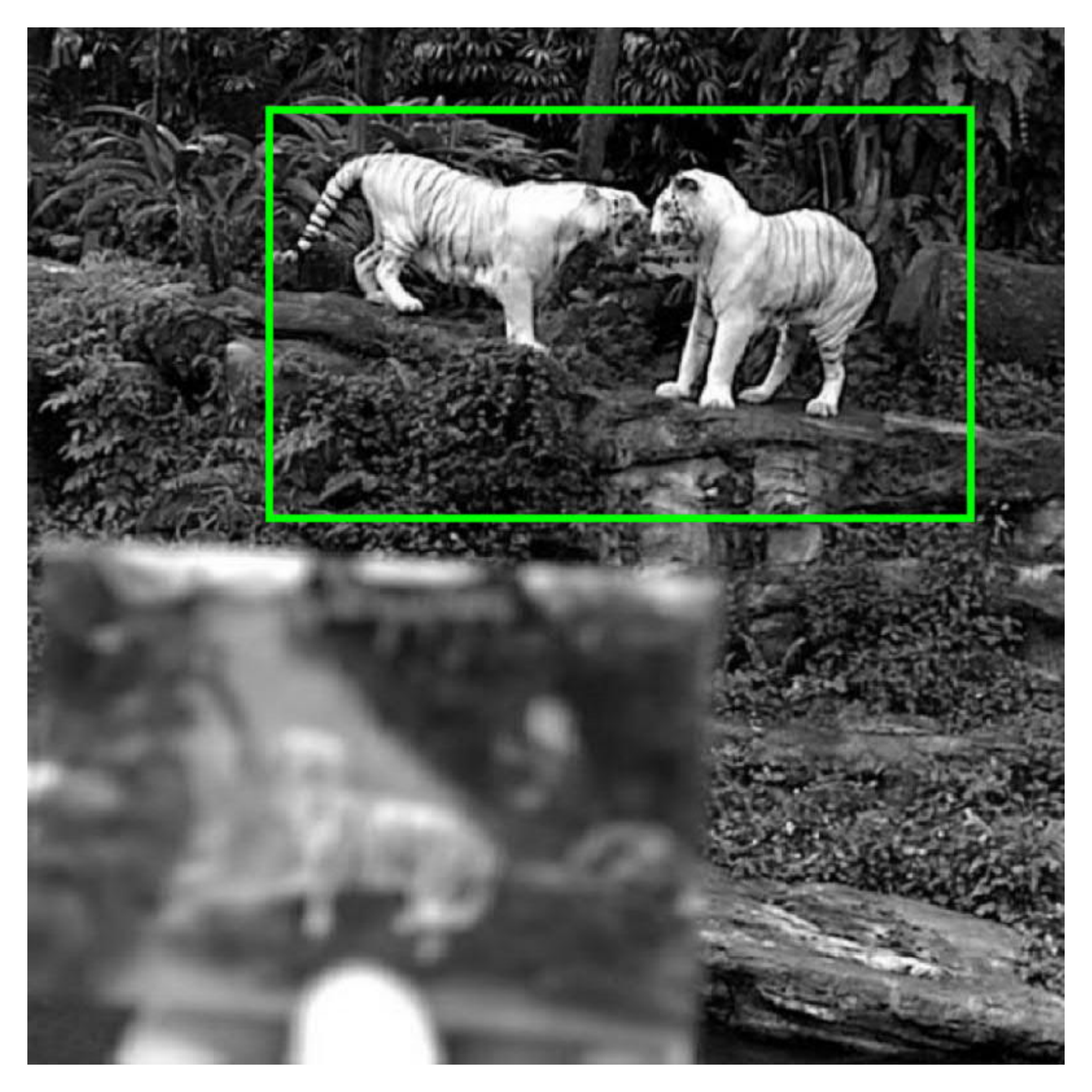}}
		\end{minipage}%
		\vfill
		\medskip
		\begin{minipage}{0.2\linewidth}
			\centerline{\includegraphics[width=3.5cm,height=3.5cm]{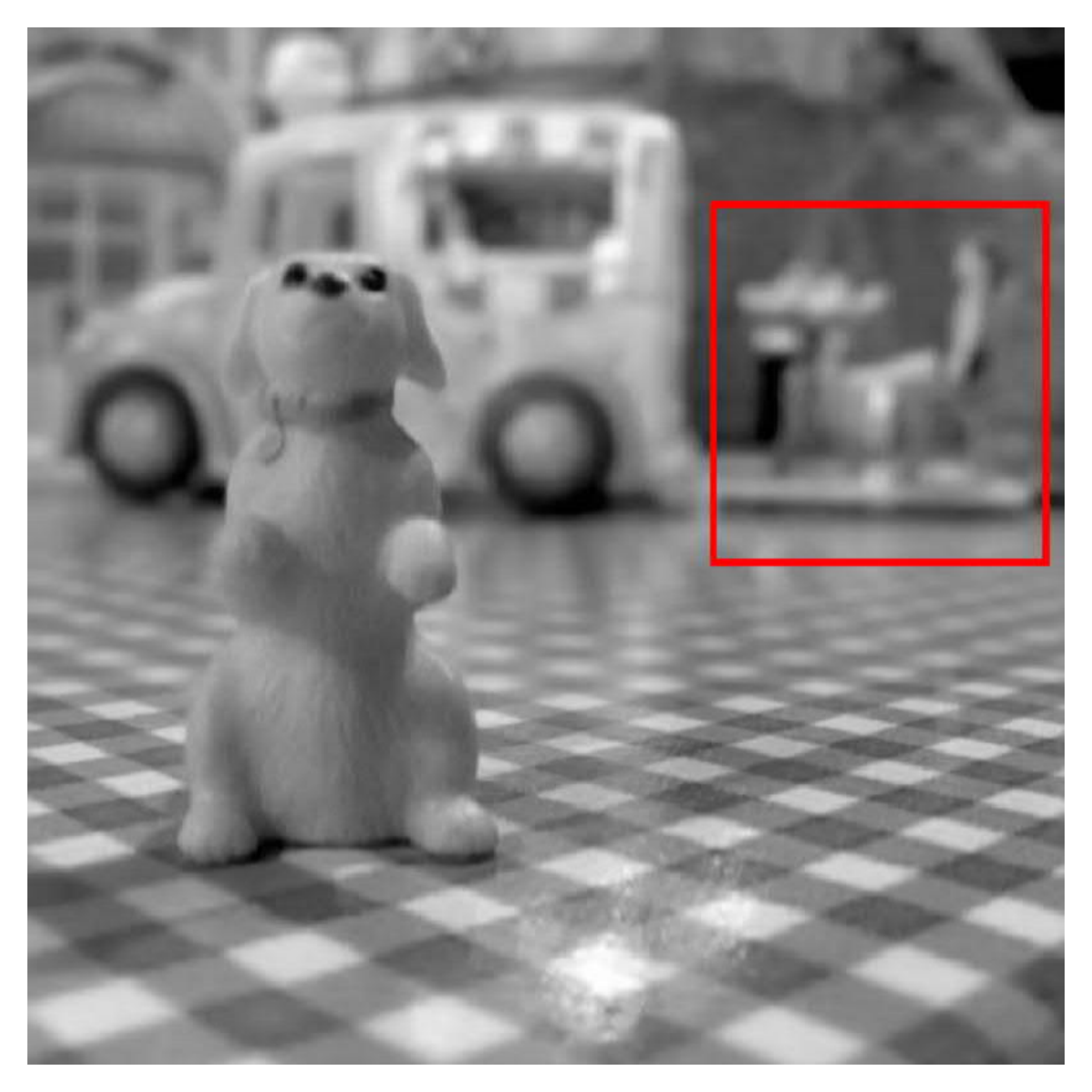}}
		\end{minipage}%
		\;
		\begin{minipage}{0.2\linewidth}
			\centerline{\includegraphics[width=3.5cm,height=3.5cm]{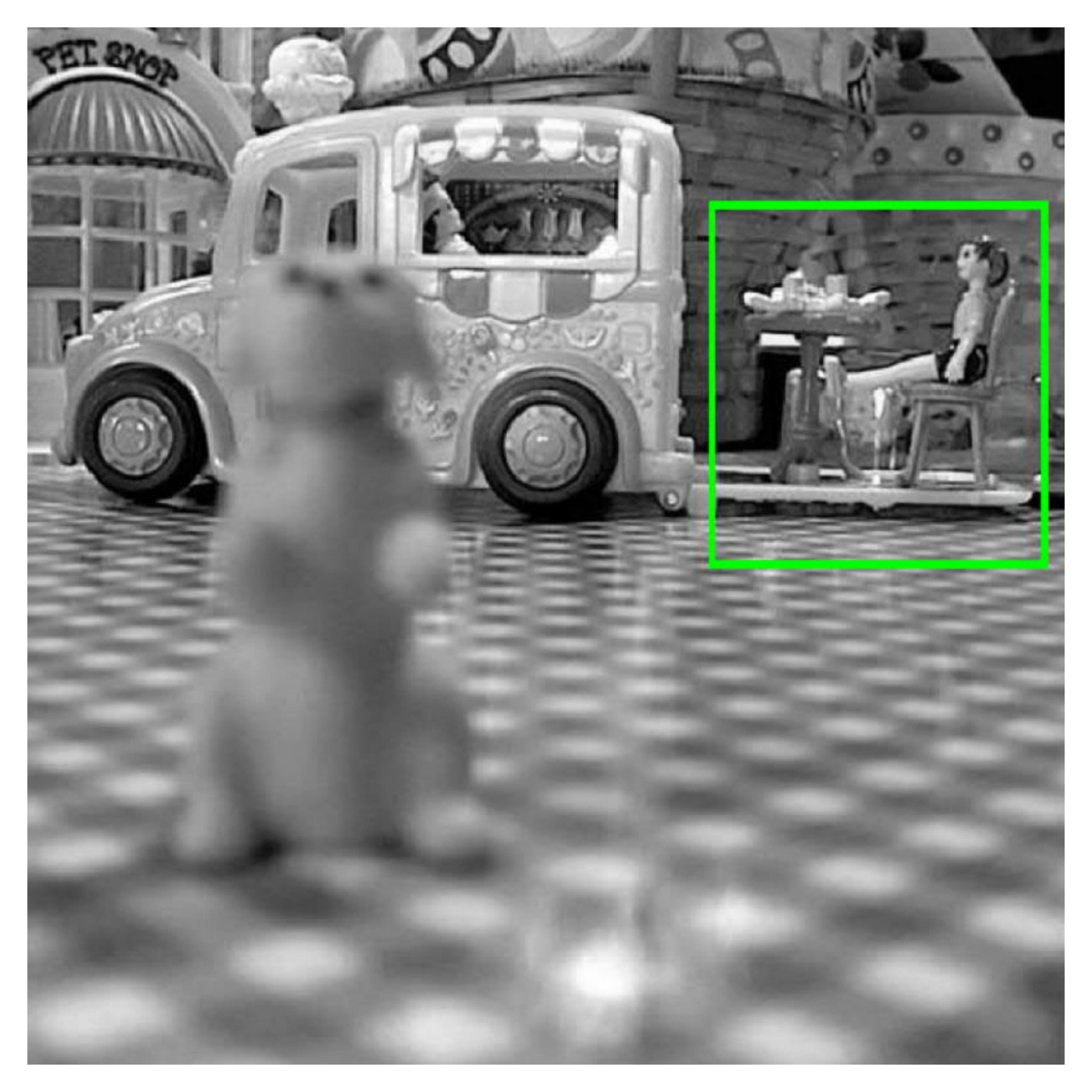}}
			
		\end{minipage}%
		
		\centering
		\caption{Learning data~\cite{29}: The red and green rectangles show parts of images used as blurred and focused learning data, respectively.}
		\label{fig:trainimages}
		\vspace*{-5pt}
	\end{center}	
\end{figure}

The proposed approach is compared to the following existing state-of-the-art multi-focus image fusion algorithms:
discrete wavelet transform-based image fusion approach (DWT)~\cite{34}, sparse representation ``choose-max''-based image fusion approach (SR-CM)~\cite{16}, sparse representation ``choose-max''-based image fusion via trained dictionary using K-SVD (SR-KSVD), multi-focus image fusion based on principal component analysis (PCA)~\cite{40}, multi-focus image fusion using dictionary-based sparse representation of focus measures (SR-FM)~\cite{17}, and multi-focus image fusion using dense SIFT (DSIFT)~\cite{38}.

\begin{figure}[!htb]
	\begin{center}
		\begin{minipage}{0.05\linewidth}
			\centerline{a}	
		\end{minipage}%
		\begin{minipage}{0.2\linewidth}
			\centerline{\includegraphics[width=3.5cm,height=3.5cm]{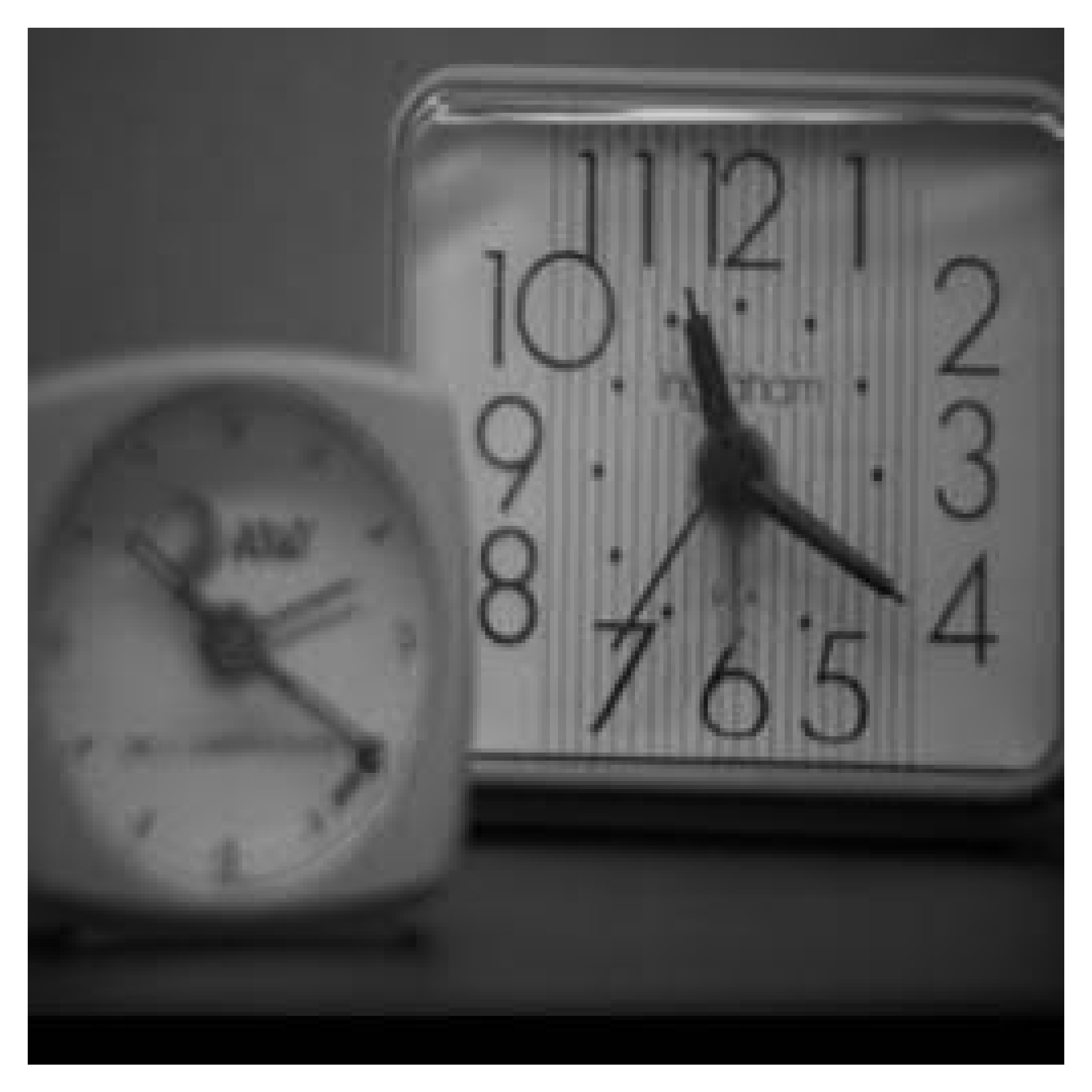}}	
		\end{minipage}%
		\;
		\begin{minipage}{0.2\linewidth}
			\centerline{\includegraphics[width=3.5cm,height=3.5cm]{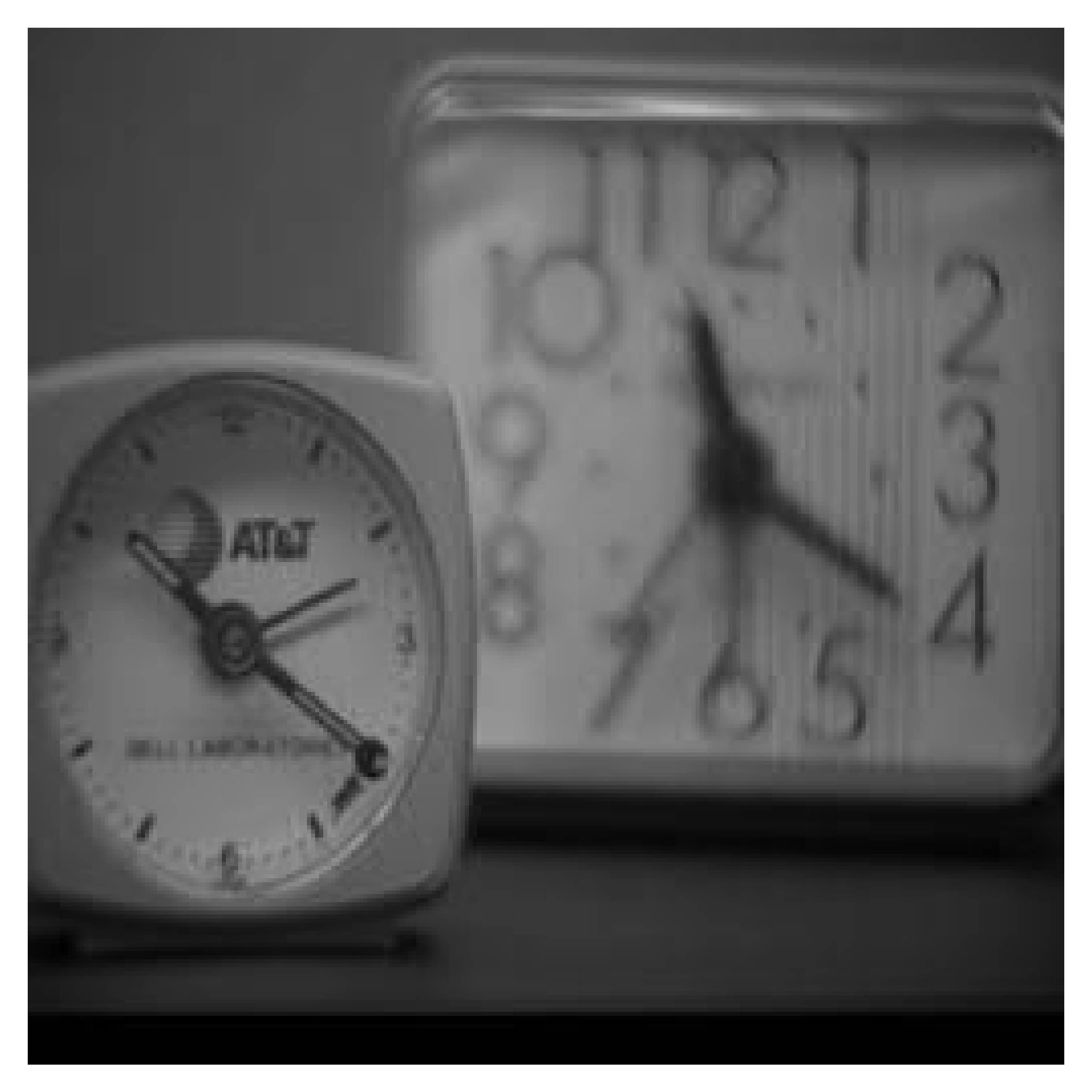}}
		\end{minipage}%
		\begin{minipage}{0.05\linewidth}
			\centerline{b}	
		\end{minipage}%
		\begin{minipage}{0.2\linewidth}
			\centerline{\includegraphics[width=3.5cm,height=3.5cm]{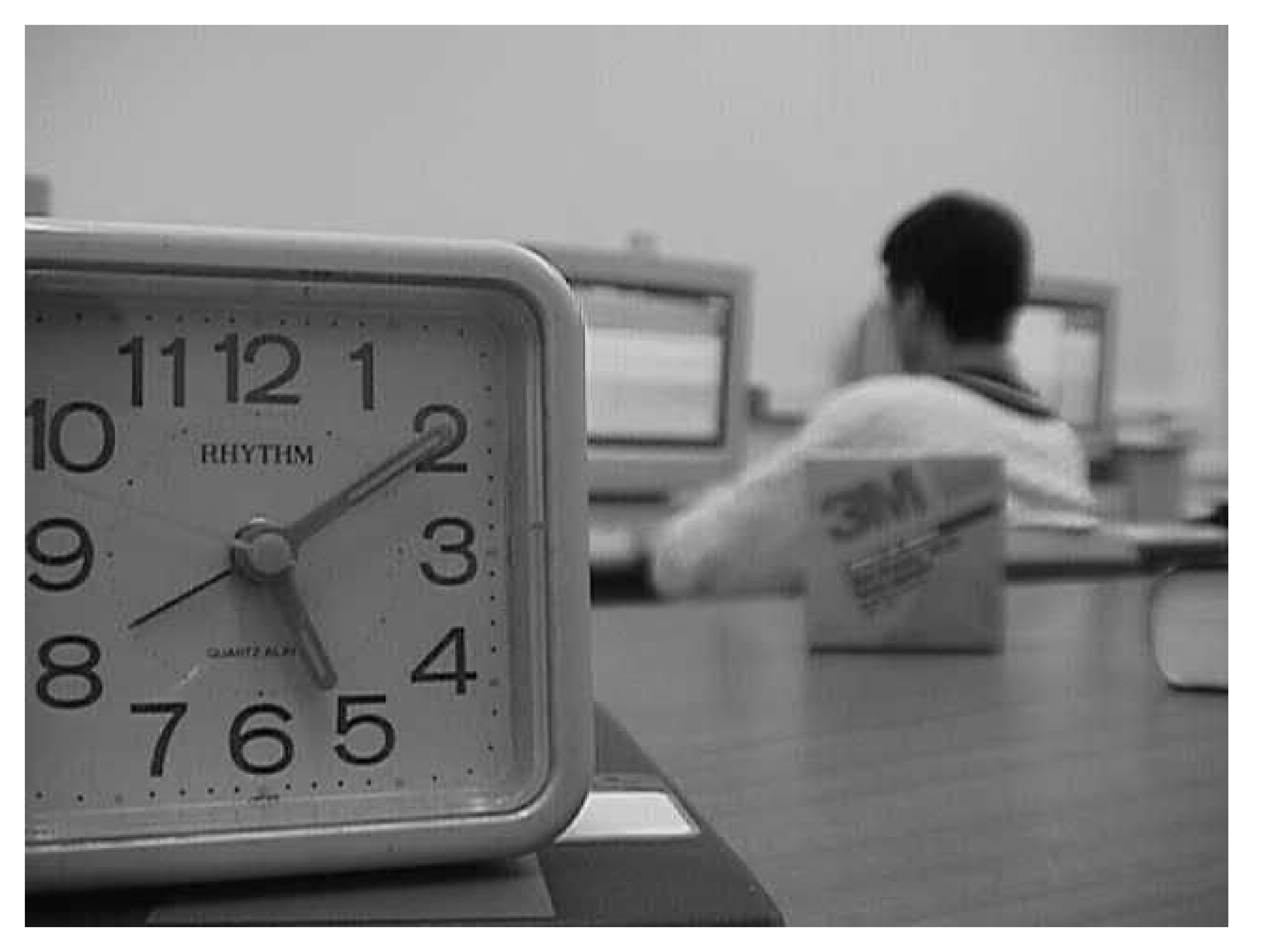}}	
		\end{minipage}%
		\;
		\begin{minipage}{0.2\linewidth}
			\centerline{\includegraphics[width=3.5cm,height=3.5cm]{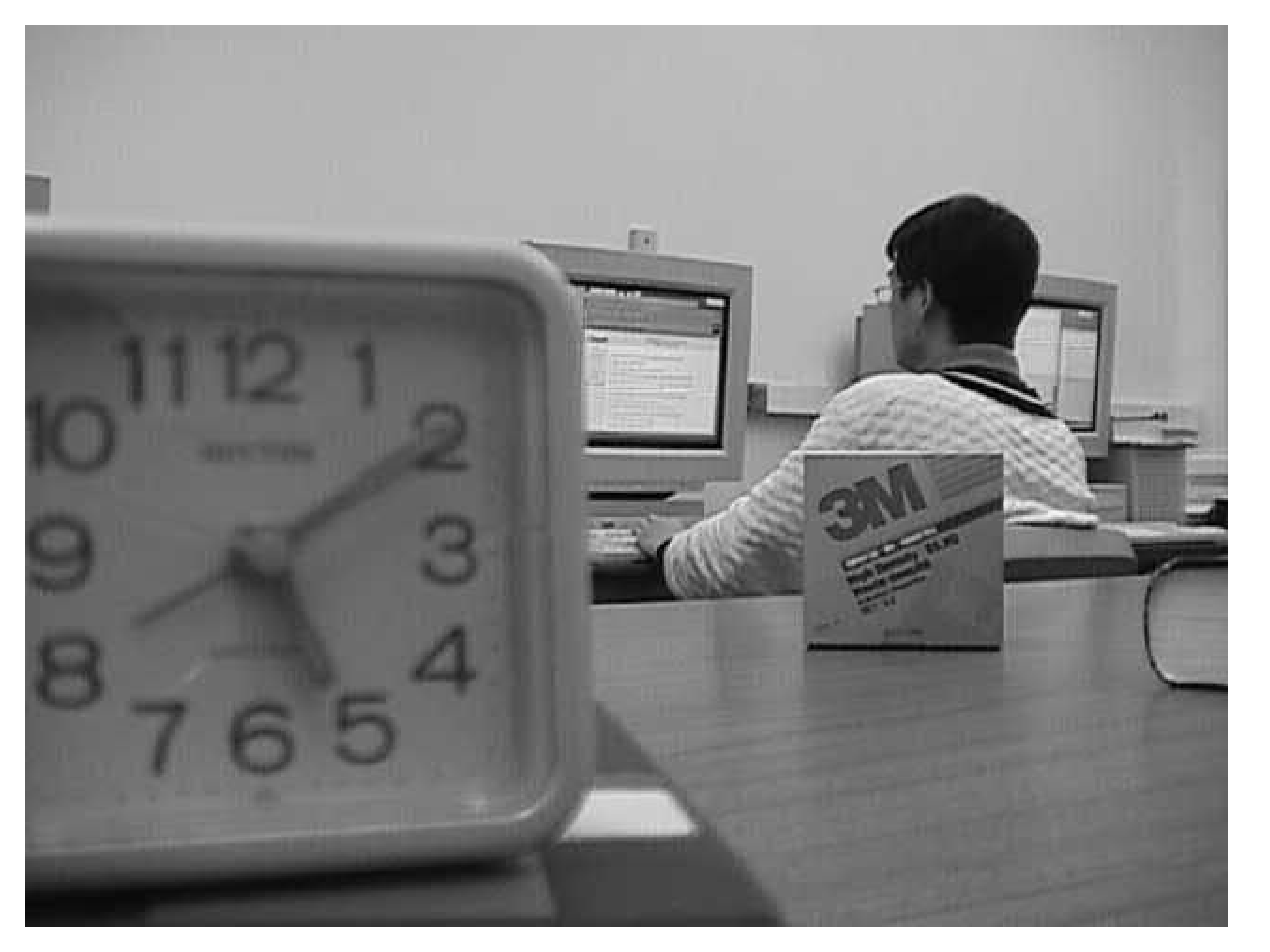}}	
		\end{minipage}%
		
		\vfill
		\begin{minipage}{0.05\linewidth}
			\centerline{c}	
		\end{minipage}%
		\begin{minipage}{0.2\linewidth}
			\centerline{\includegraphics[width=3.5cm,height=3.5cm]{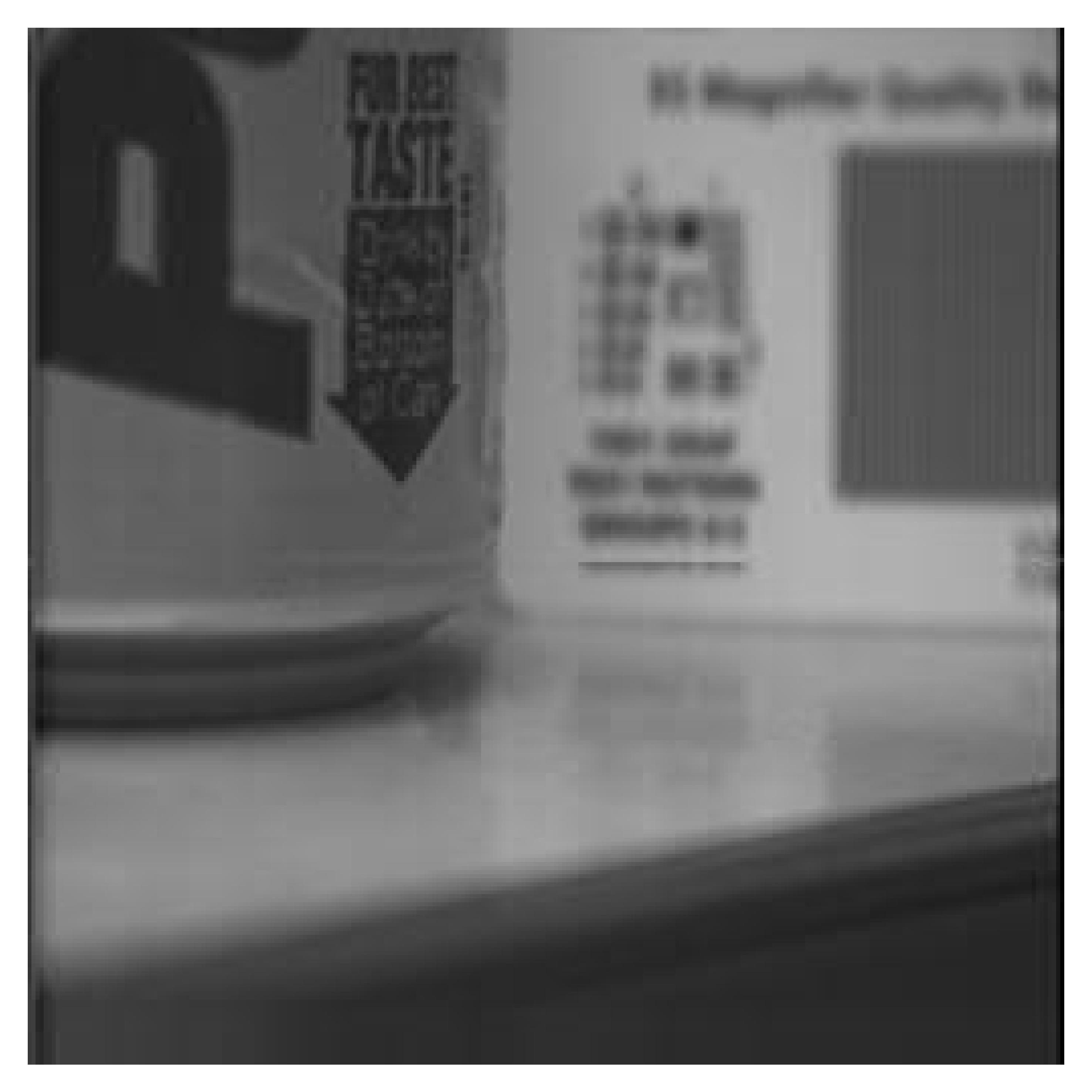}}	
		\end{minipage}%
		\;
		\begin{minipage}{0.2\linewidth}
			\centerline{\includegraphics[width=3.5cm,height=3.5cm]{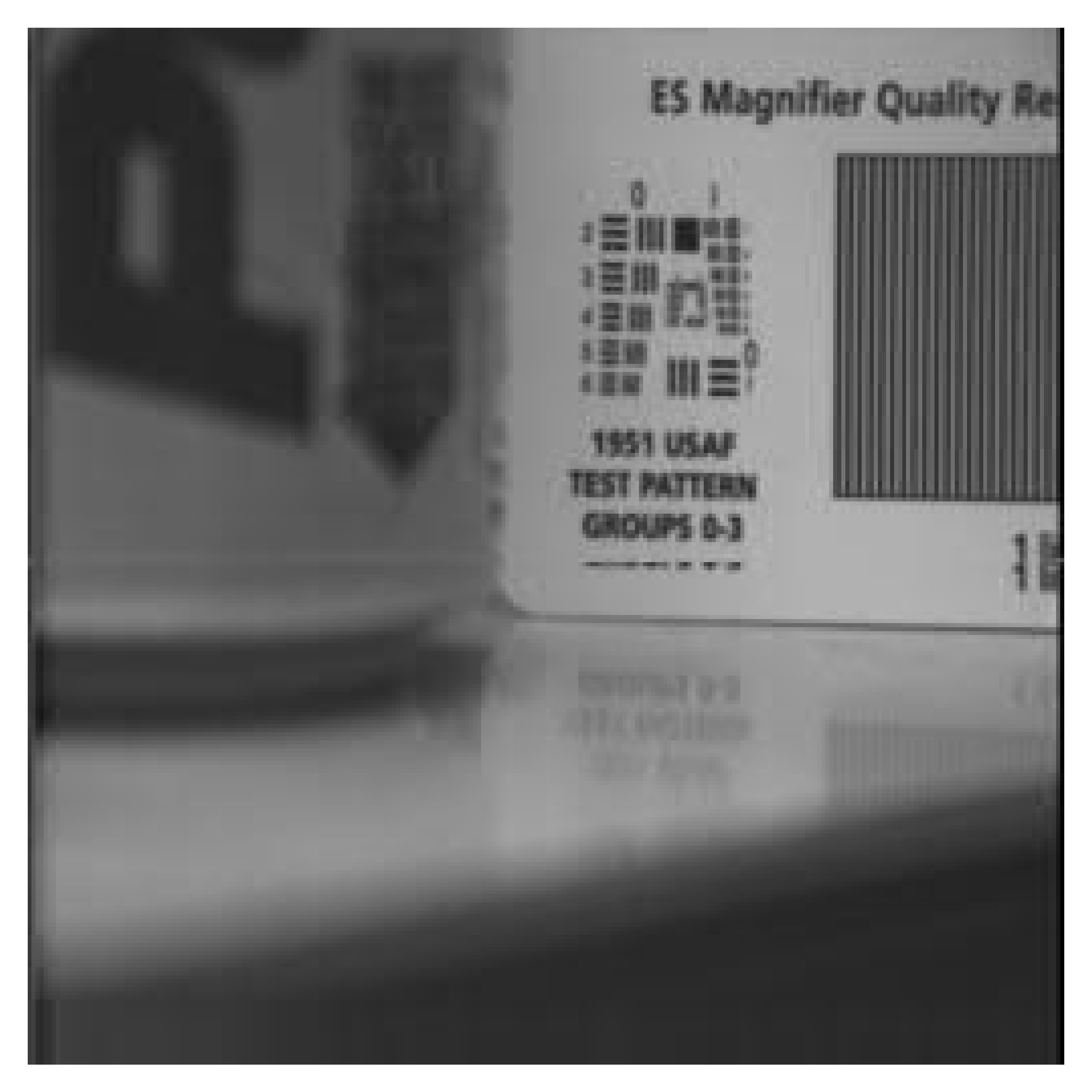}}	
		\end{minipage}%
		\begin{minipage}{0.05\linewidth}
			\centerline{d}	
		\end{minipage}%
		\begin{minipage}{0.2\linewidth}
			\centerline{\includegraphics[width=3.5cm,height=3.5cm]{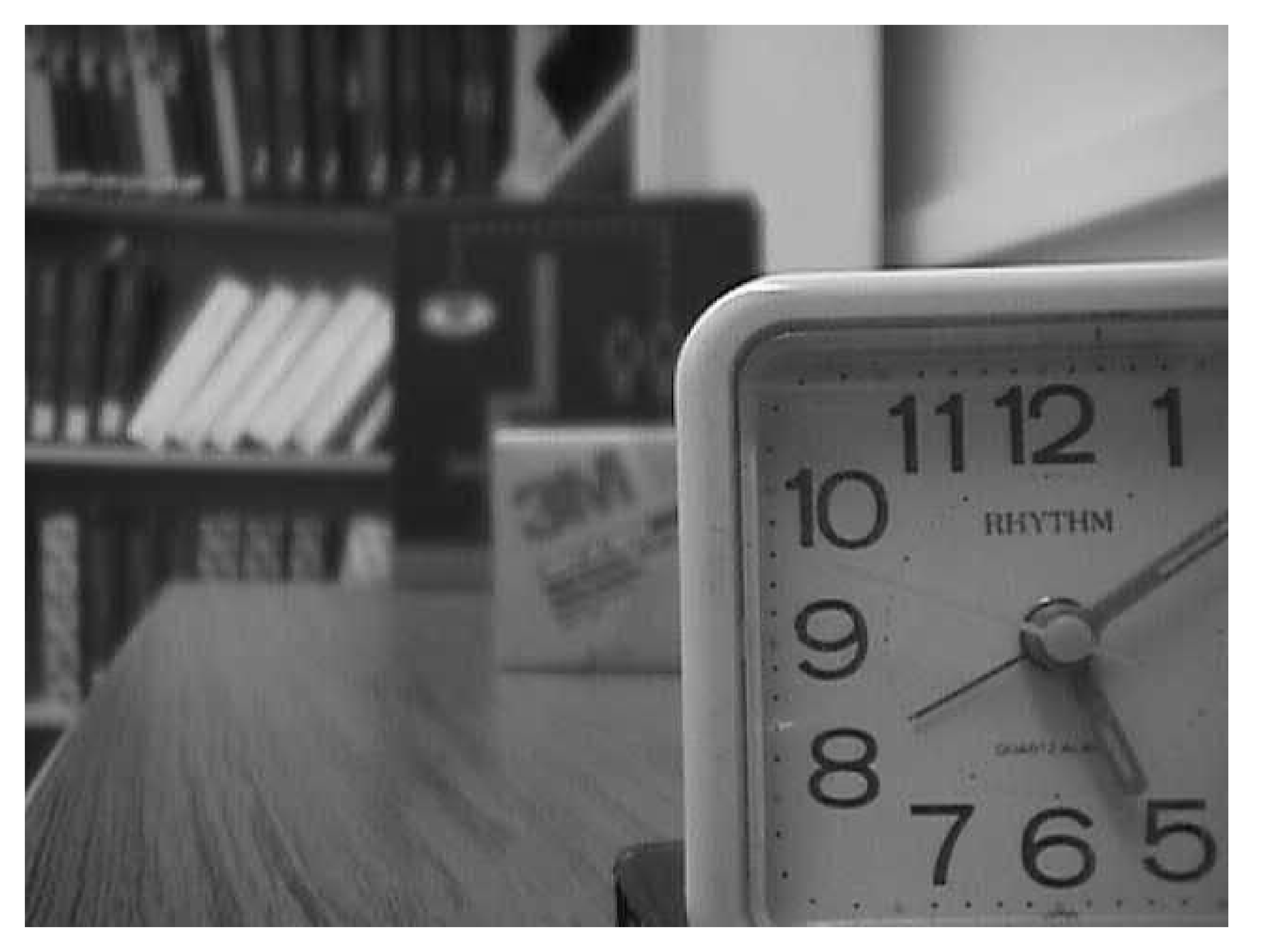}}	
		\end{minipage}%
		\;
		\begin{minipage}{0.2\linewidth}
			\centerline{\includegraphics[width=3.5cm,height=3.5cm]{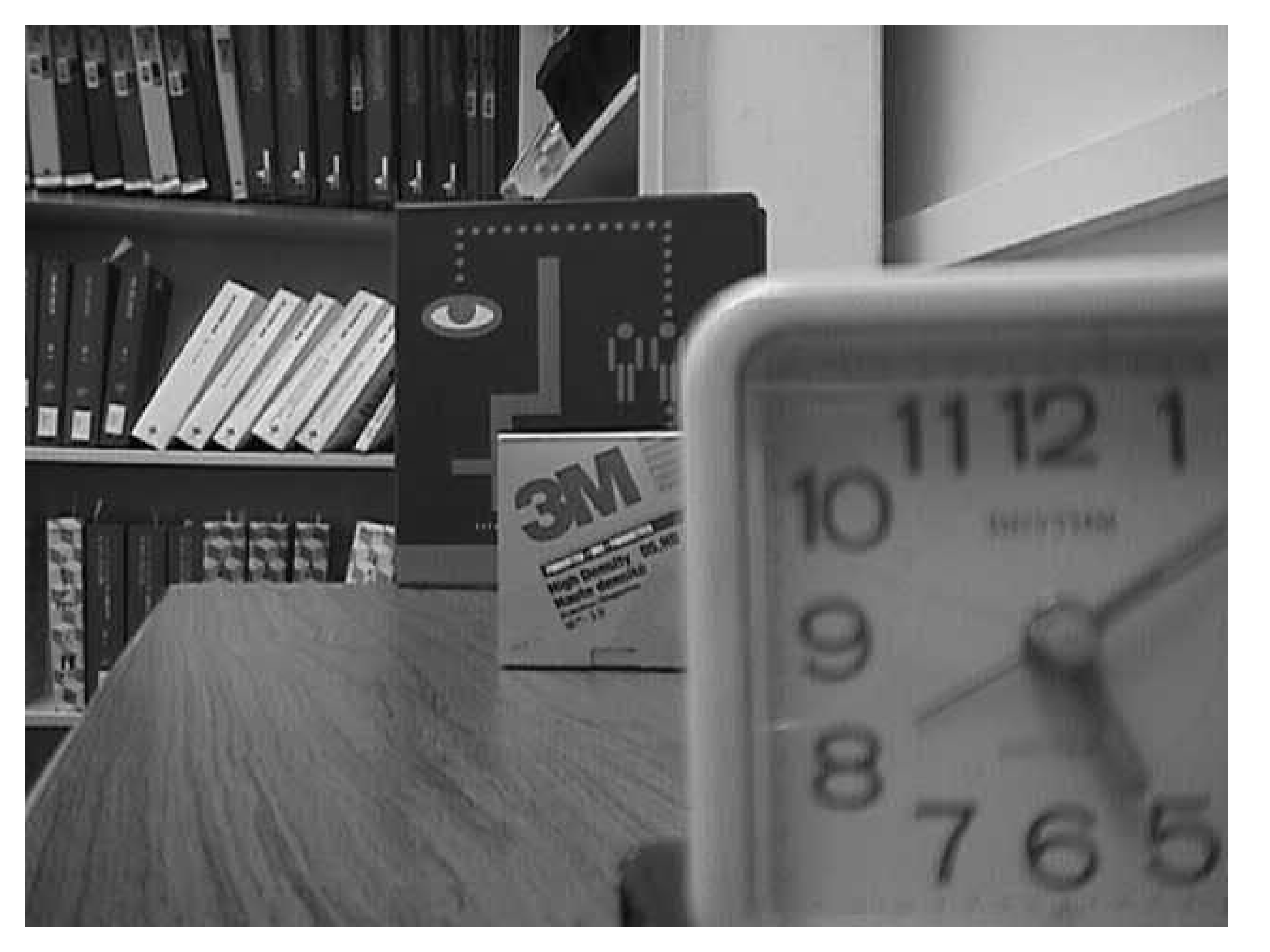}}
		\end{minipage}%
		\vfill
		\begin{minipage}{0.05\linewidth}
			\centerline{e}	
		\end{minipage}%
		\begin{minipage}{0.2\linewidth}
			\centerline{\includegraphics[width=3.5cm,height=3.5cm]{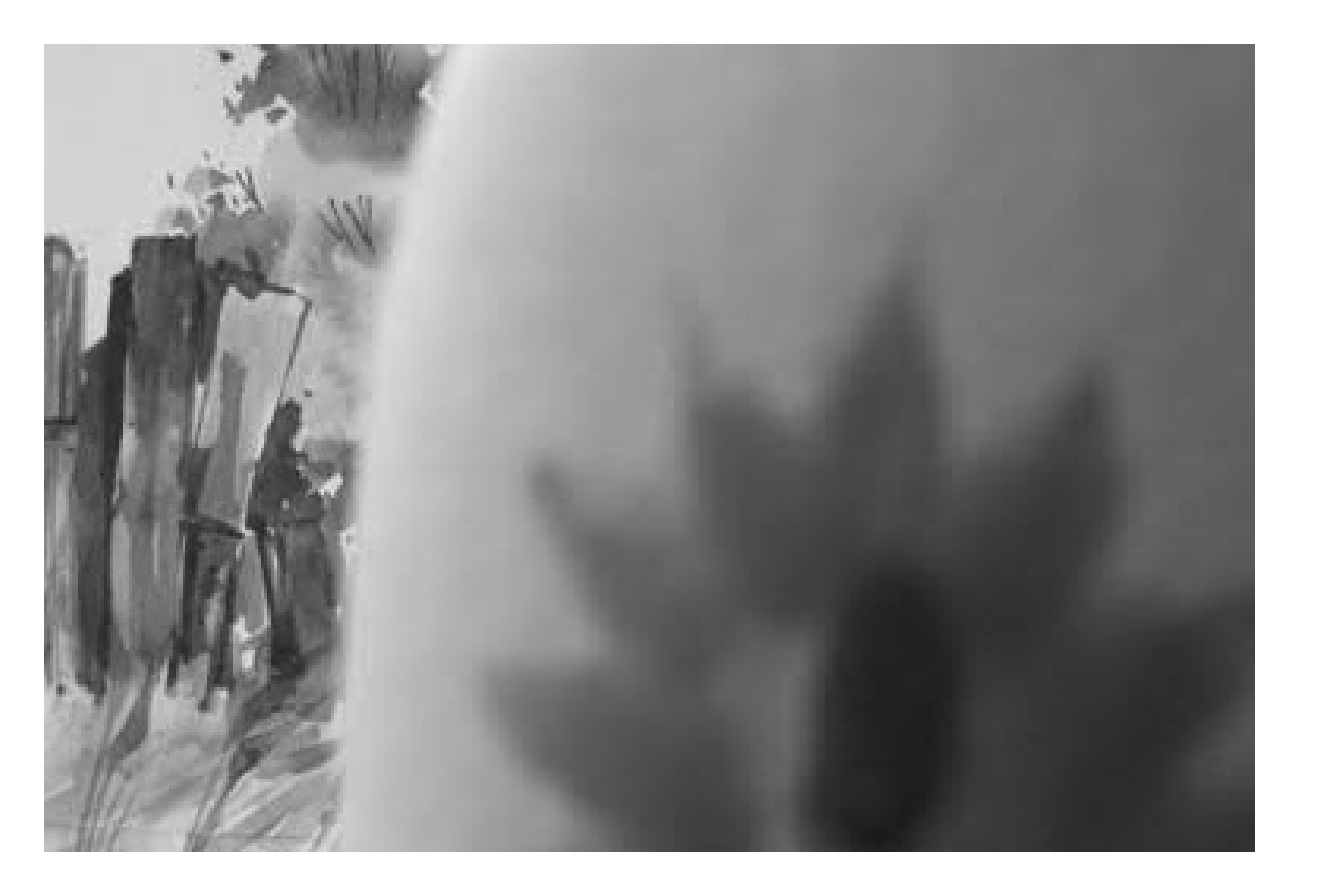}}	
		\end{minipage}%
		\;
		\begin{minipage}{0.2\linewidth}
			\centerline{\includegraphics[width=3.5cm,height=3.5cm]{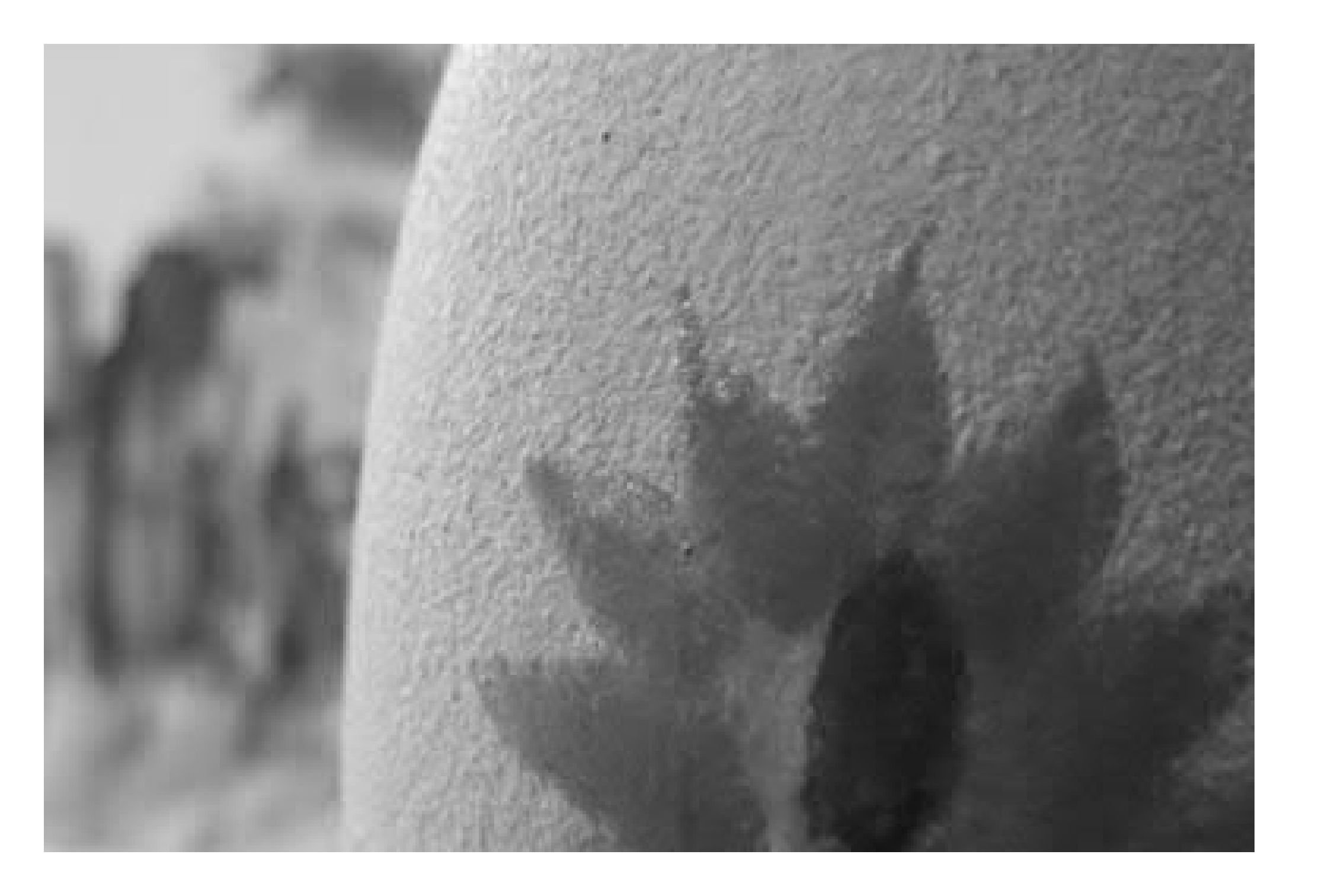}}	
		\end{minipage}%
		\begin{minipage}{0.05\linewidth}
			\centerline{f}	
		\end{minipage}%
		\begin{minipage}{0.2\linewidth}
			\centerline{\includegraphics[width=3.5cm,height=3.5cm]{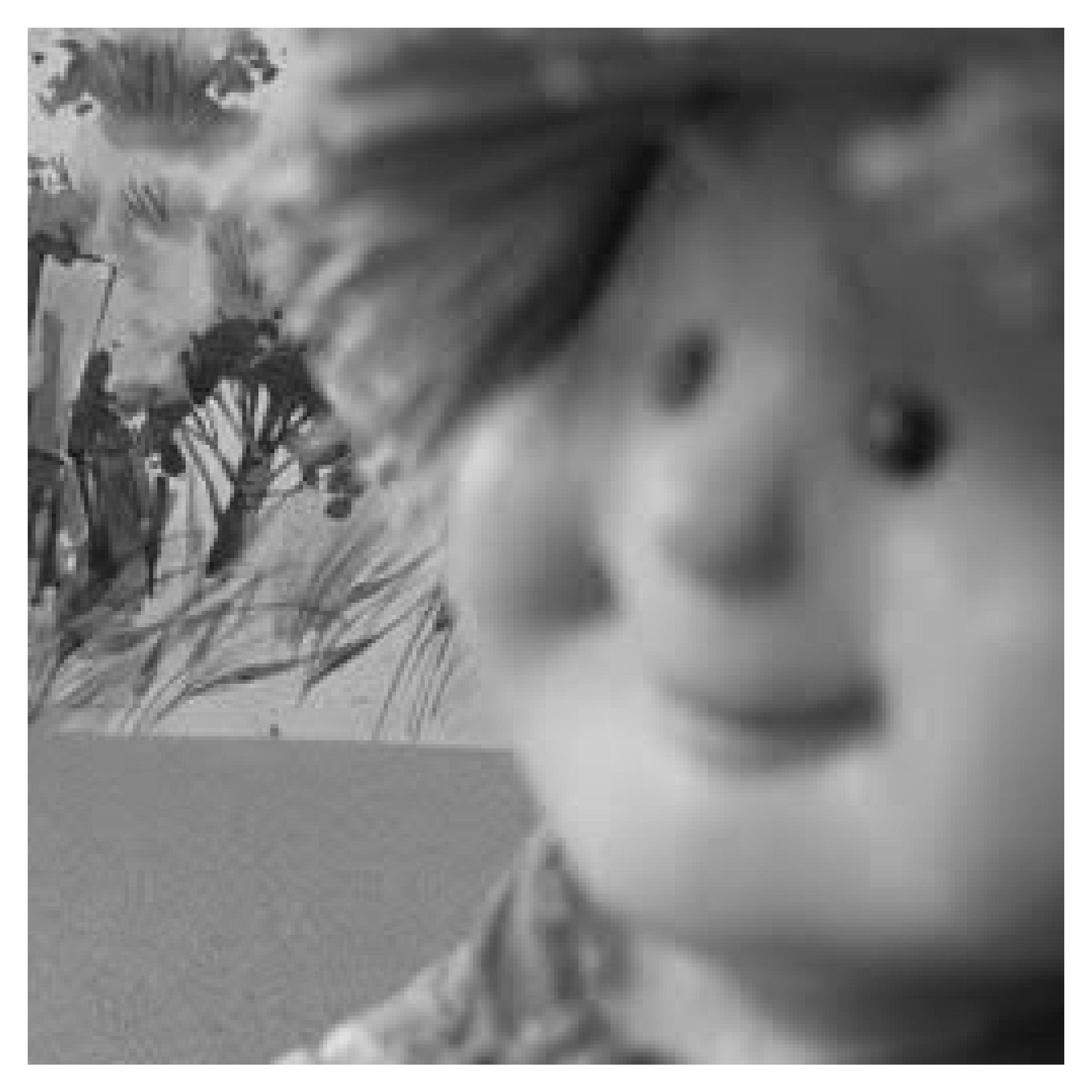}}	
		\end{minipage}%
		\;
		\begin{minipage}{0.2\linewidth}
			\centerline{\includegraphics[width=3.5cm,height=3.5cm]{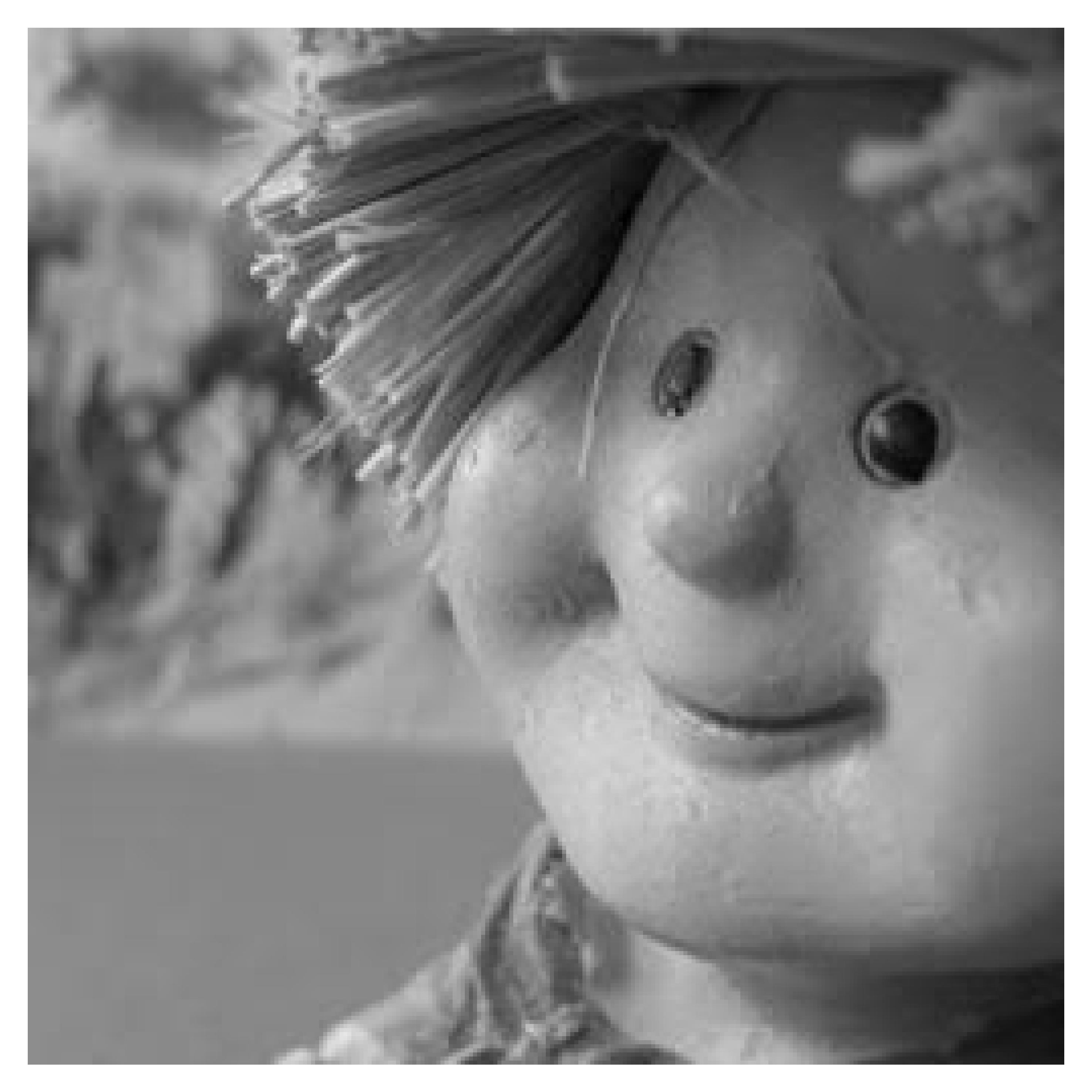}}	
		\end{minipage}%
		\centering
		\caption{Input 1: gray-scale multi-focus images\cite{17}: (a) Clocks, (b) Lab, (c) Pepsi, (d) Disk, (e) Jug, and (f) Doll.}
		\label{fig:sourceimages}
		\vspace*{-5pt}
	\end{center}
\end{figure}
\begin{figure}[!htb]
	\begin{center}
		\begin{minipage}{0.05\linewidth}
			\centerline{a}	
		\end{minipage}%
		\begin{minipage}{0.3\linewidth}
			\centerline{\includegraphics[width=4cm,height=4cm]{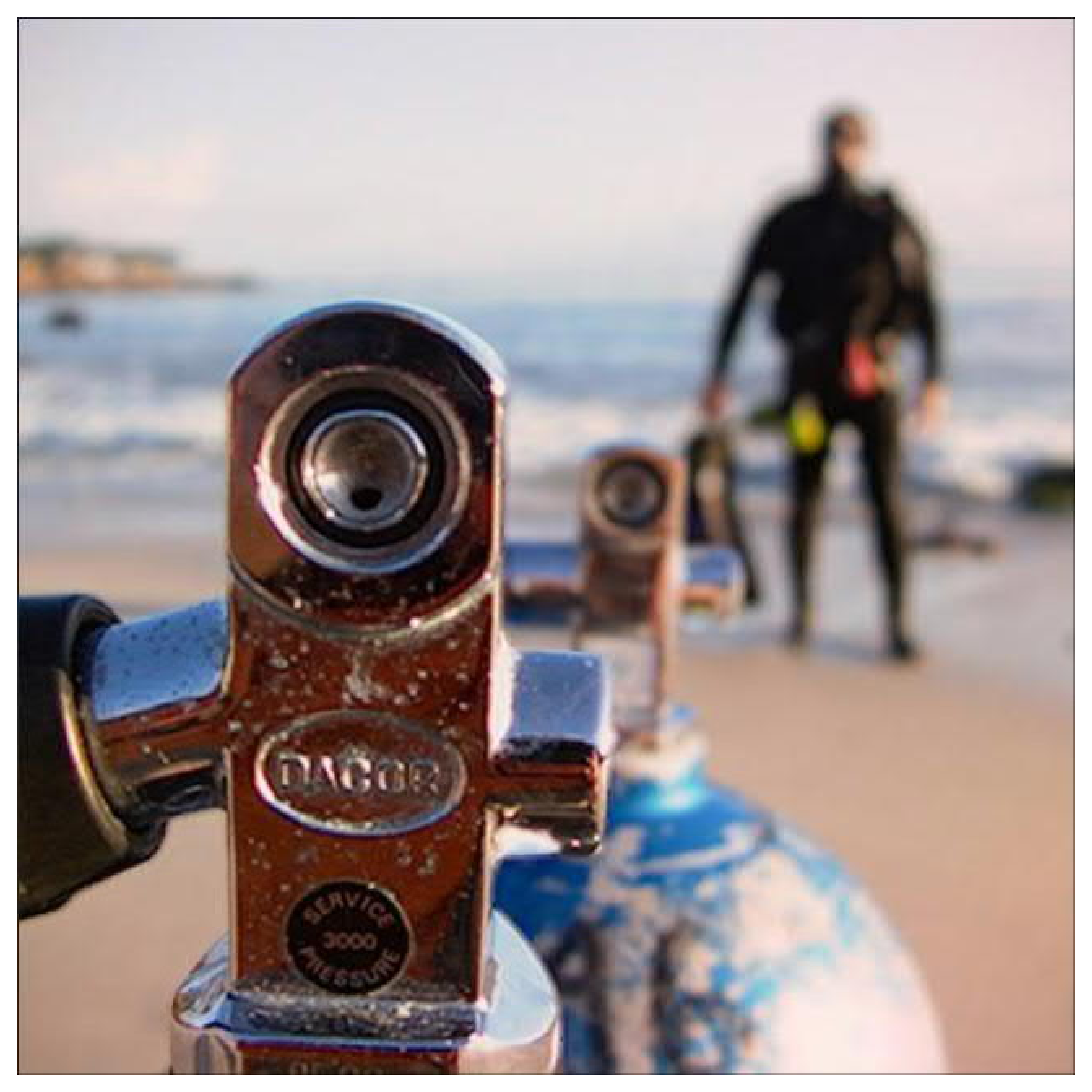}}	
		\end{minipage}%
		\begin{minipage}{0.3\linewidth}
			\centerline{\includegraphics[width=4cm,height=4cm]{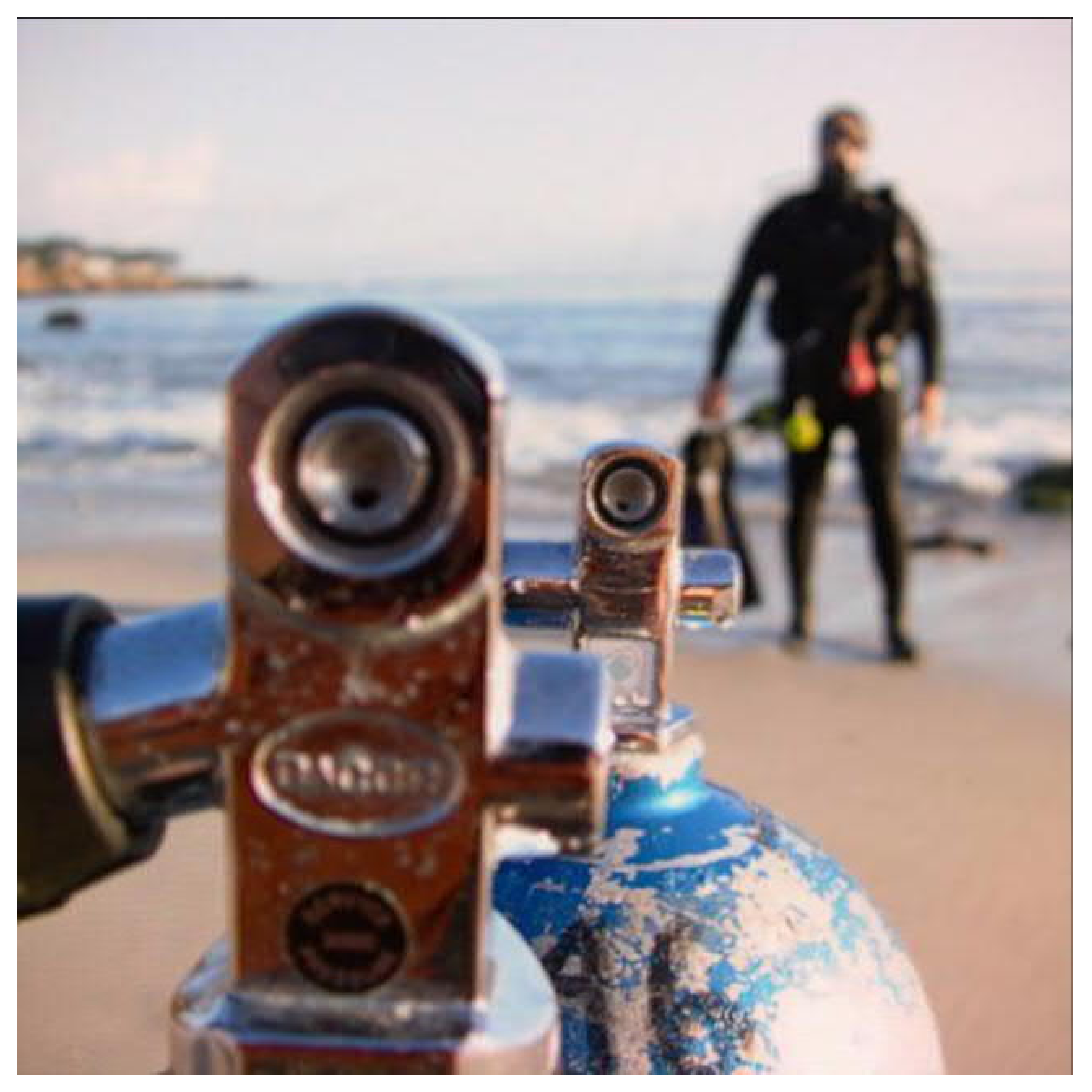}}	
		\end{minipage}%
		\begin{minipage}{0.3\linewidth}
			\centerline{\includegraphics[width=4cm,height=4cm]{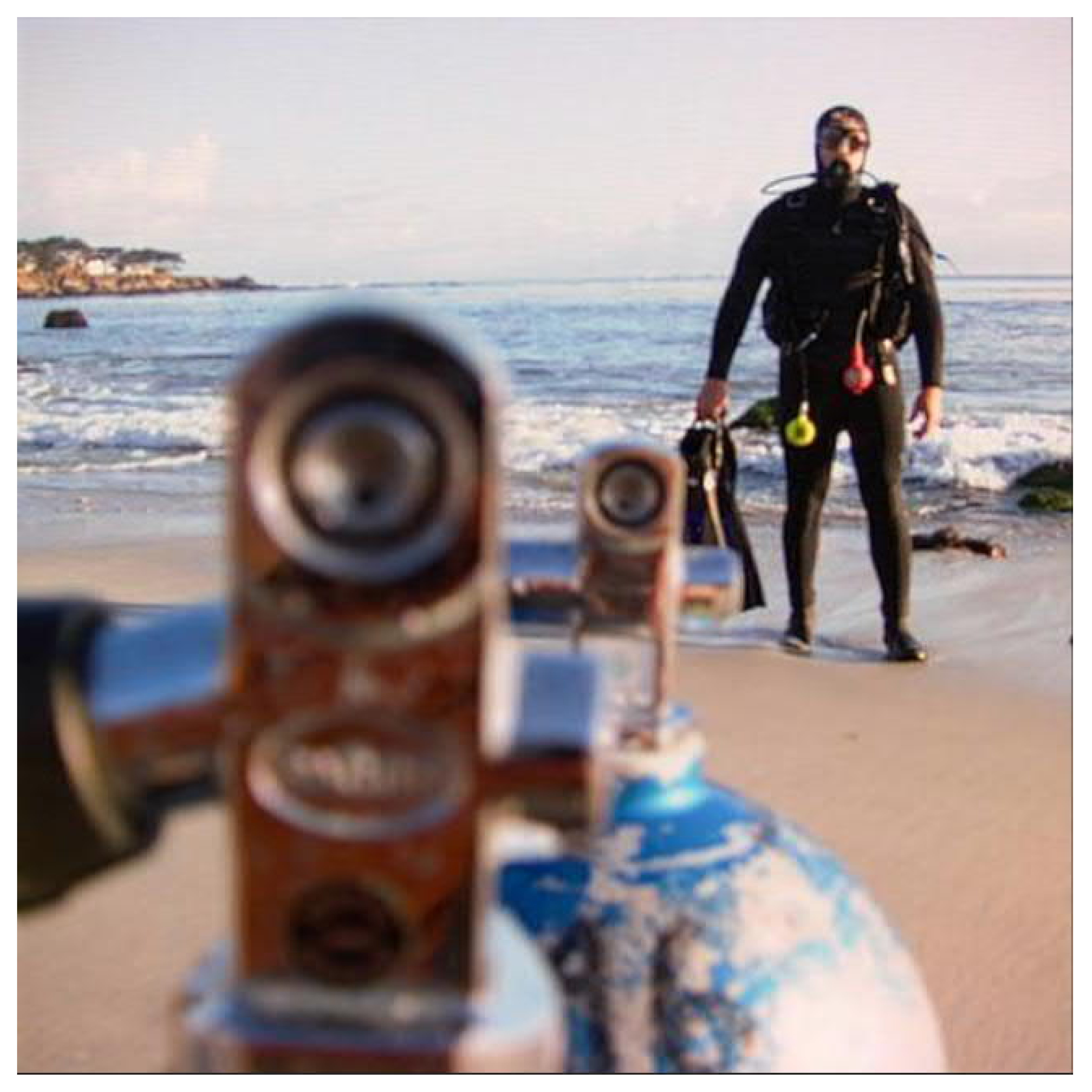}}
		\end{minipage}%
		\vfill
		\medskip
		\begin{minipage}{0.05\linewidth}
			\centerline{b}	
		\end{minipage}%
		\begin{minipage}{0.3\linewidth}
			\centerline{\includegraphics[width=4cm,height=4cm]{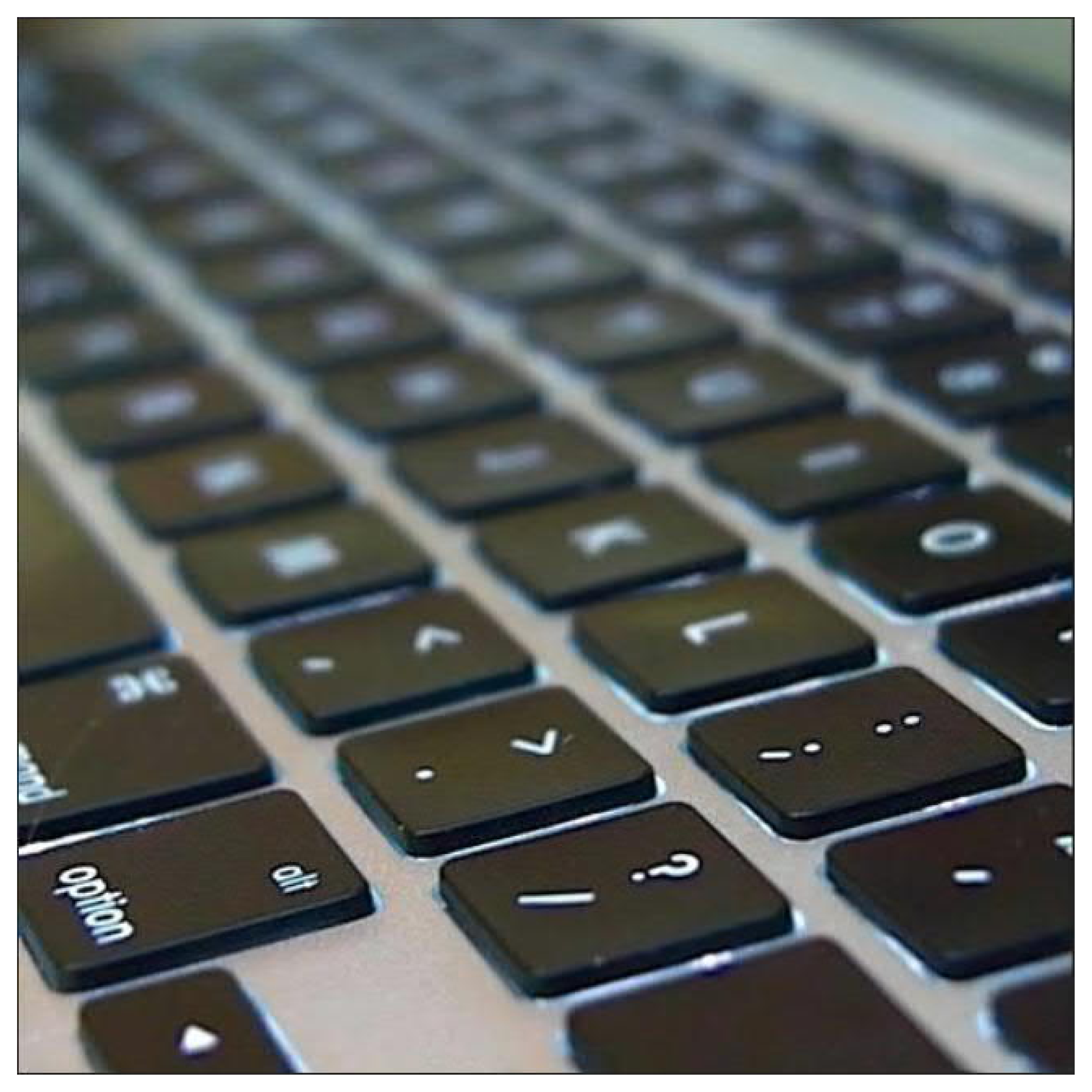}}
		\end{minipage}%
		\begin{minipage}{0.3\linewidth}
			\centerline{\includegraphics[width=4cm,height=4cm]{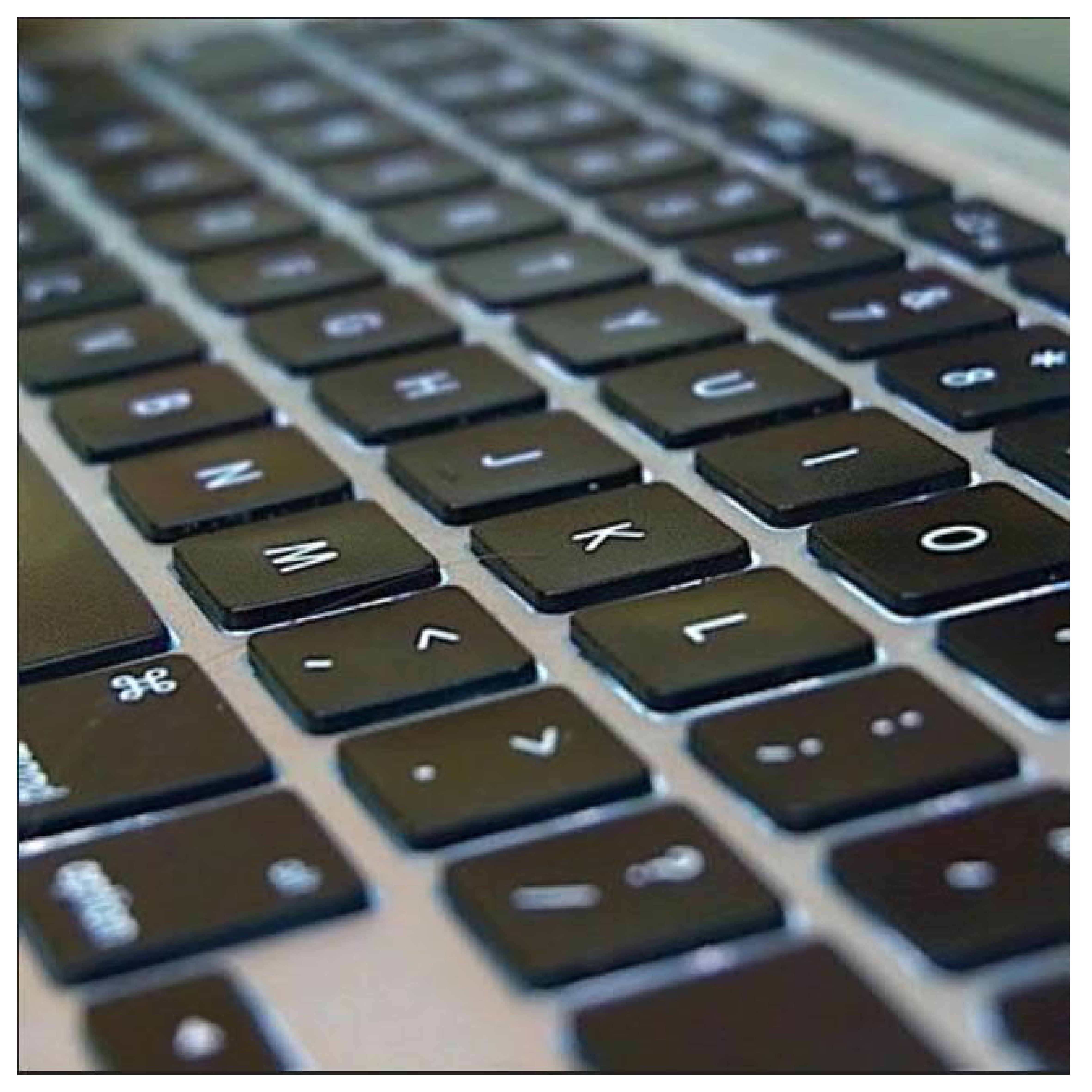}}	
		\end{minipage}%
		\begin{minipage}{0.3\linewidth}
			\centerline{\includegraphics[width=4cm,height=4cm]{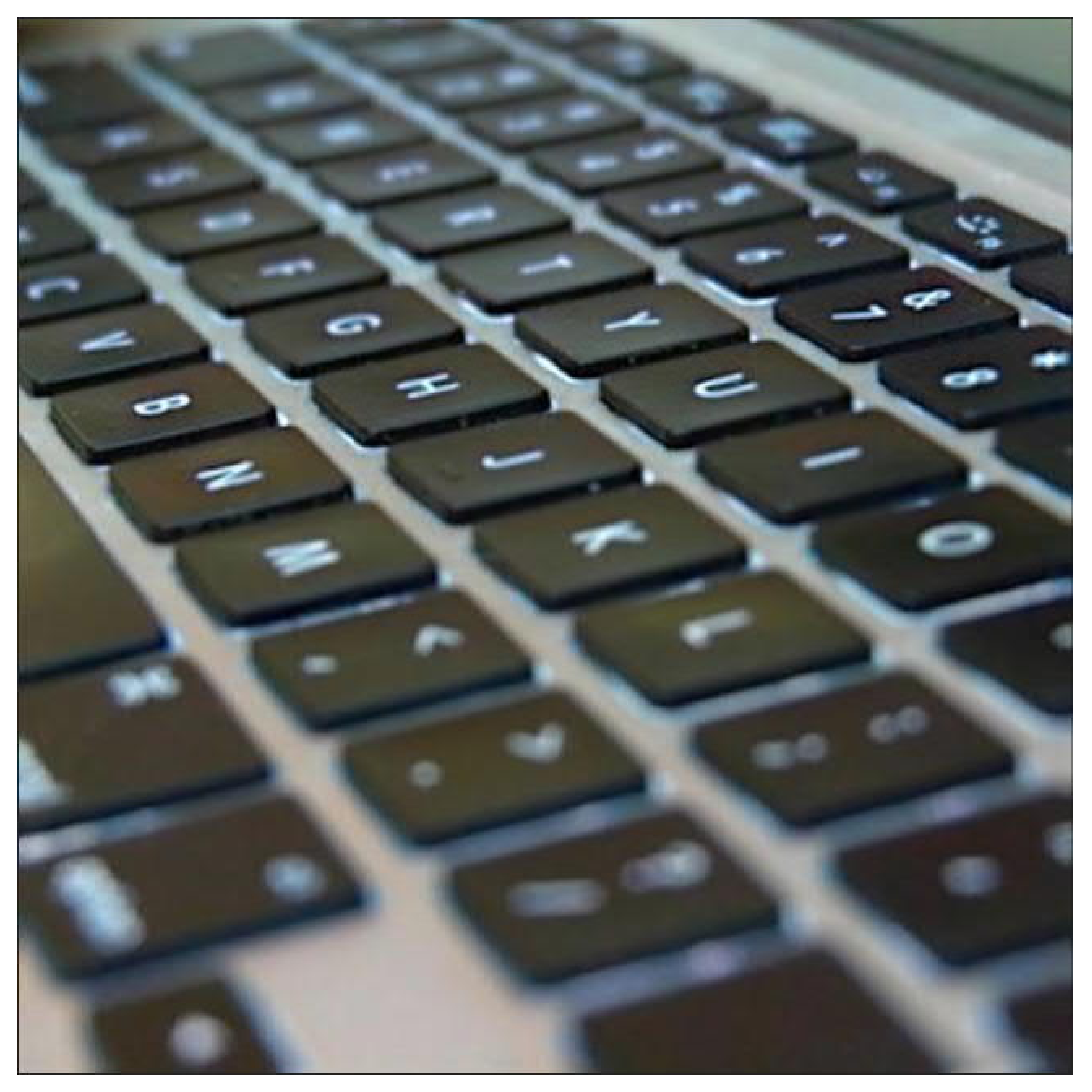}}
		\end{minipage}%
				\vfill
		\medskip
		\begin{minipage}{0.05\linewidth}
			\centerline{c}	
		\end{minipage}%
		\begin{minipage}{0.3\linewidth}
			\centerline{\includegraphics[width=4cm,height=4cm]{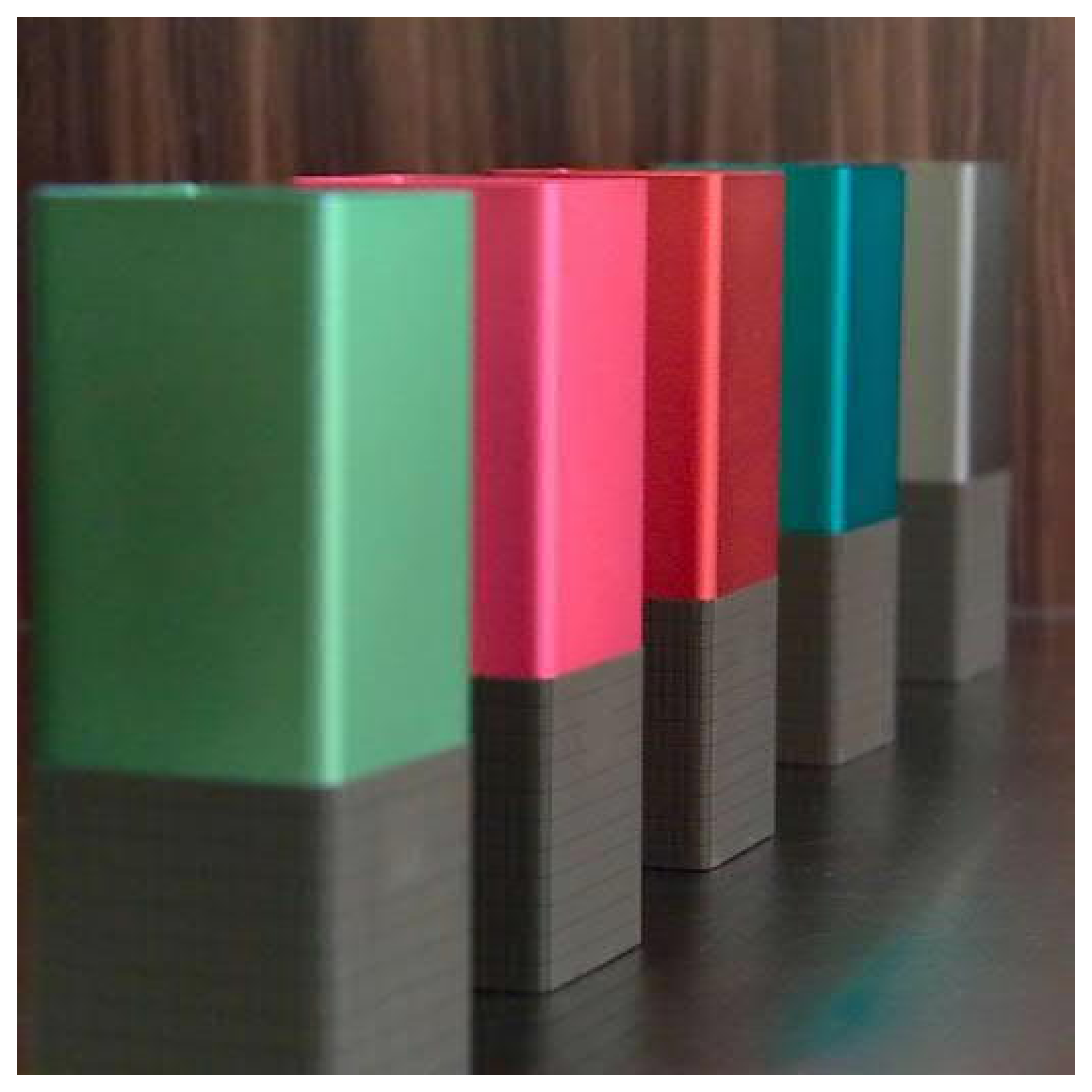}}
		\end{minipage}%
		\begin{minipage}{0.3\linewidth}
			\centerline{\includegraphics[width=4cm,height=4cm]{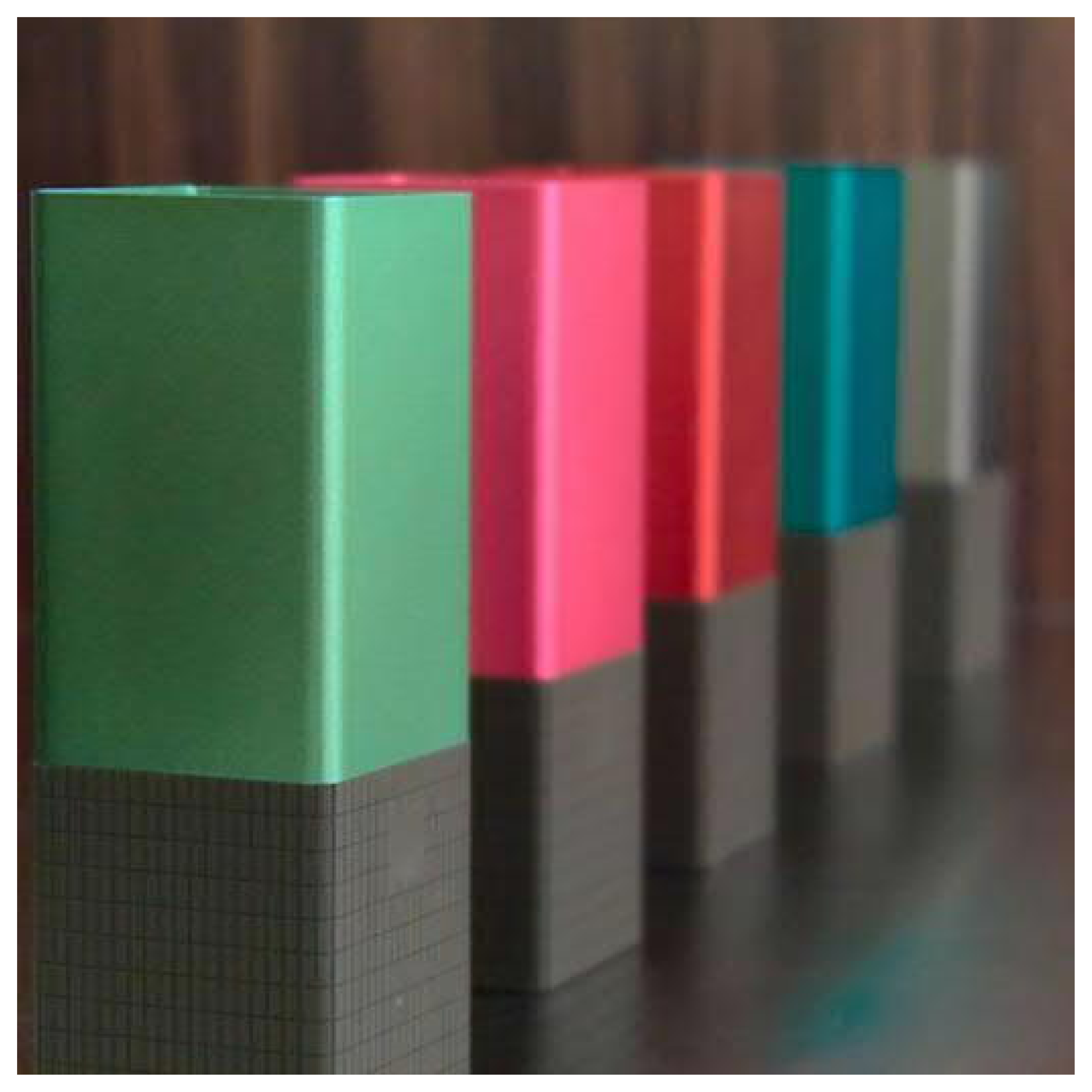}}	
		\end{minipage}%
		\begin{minipage}{0.3\linewidth}
			\centerline{\includegraphics[width=4cm,height=4cm]{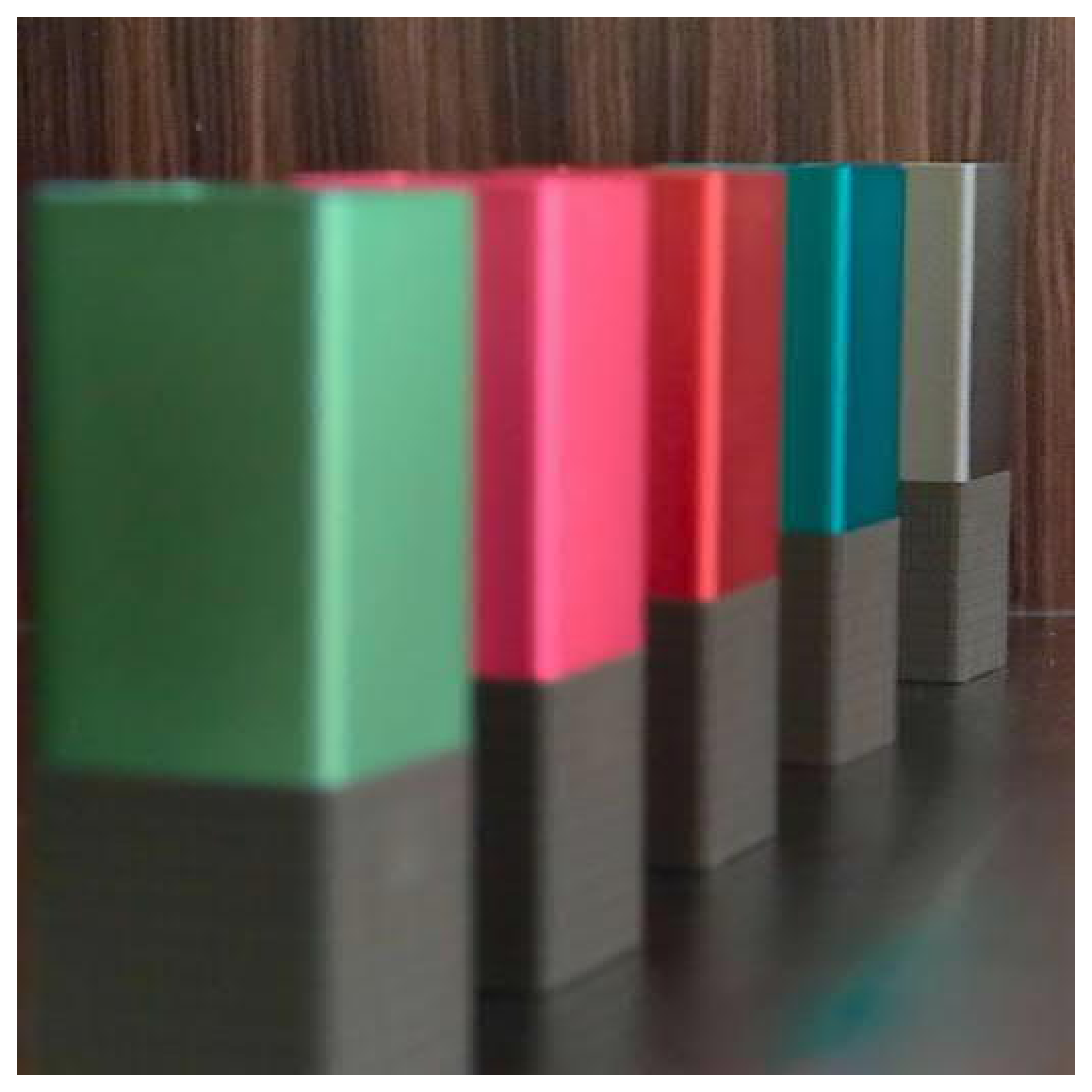}}
		\end{minipage}%
			\vfill
	\medskip
	\begin{minipage}{0.05\linewidth}
		\centerline{d}	
	\end{minipage}%
	\begin{minipage}{0.3\linewidth}
		\centerline{\includegraphics[width=4cm,height=4cm]{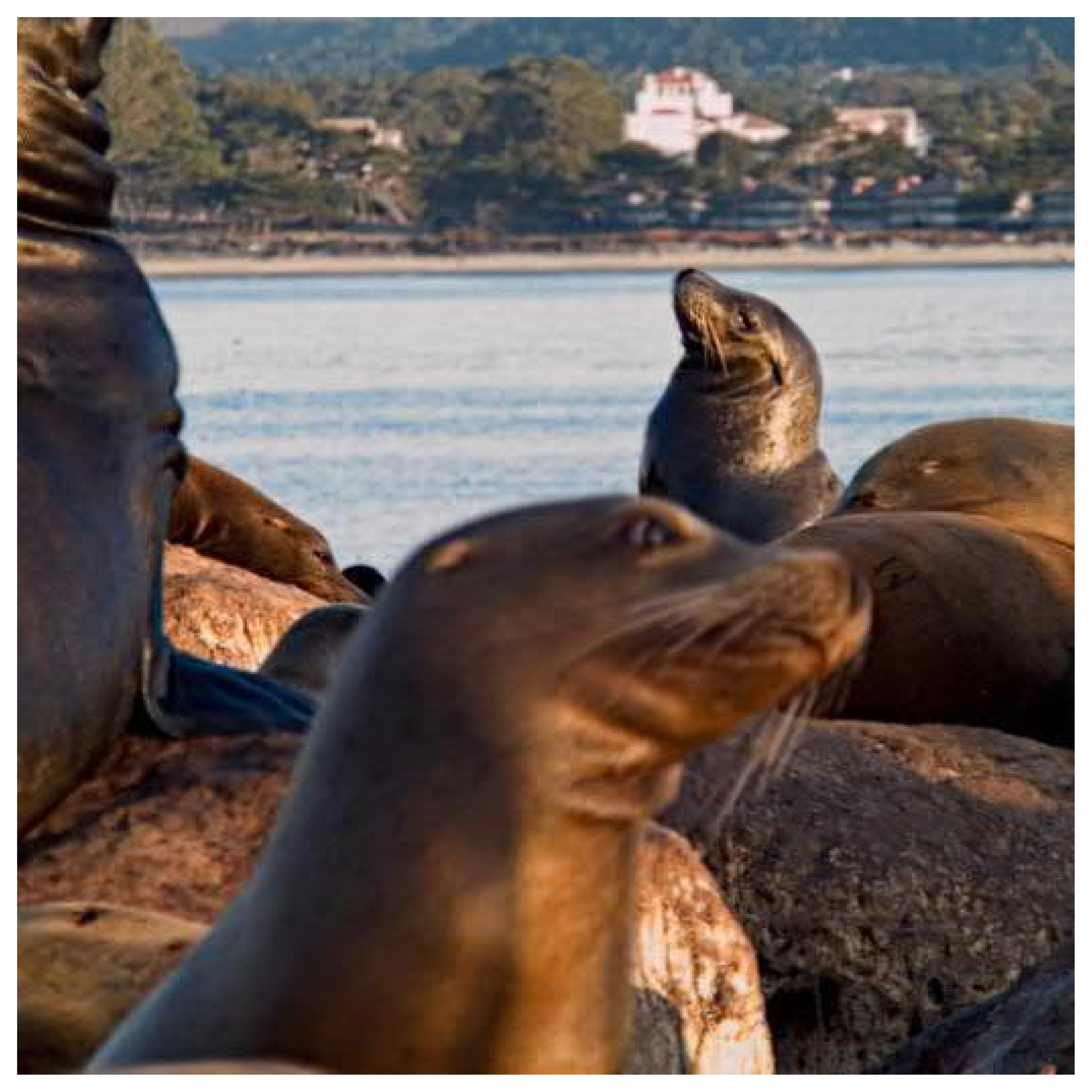}}
	\end{minipage}%
	\begin{minipage}{0.3\linewidth}
		\centerline{\includegraphics[width=4cm,height=4cm]{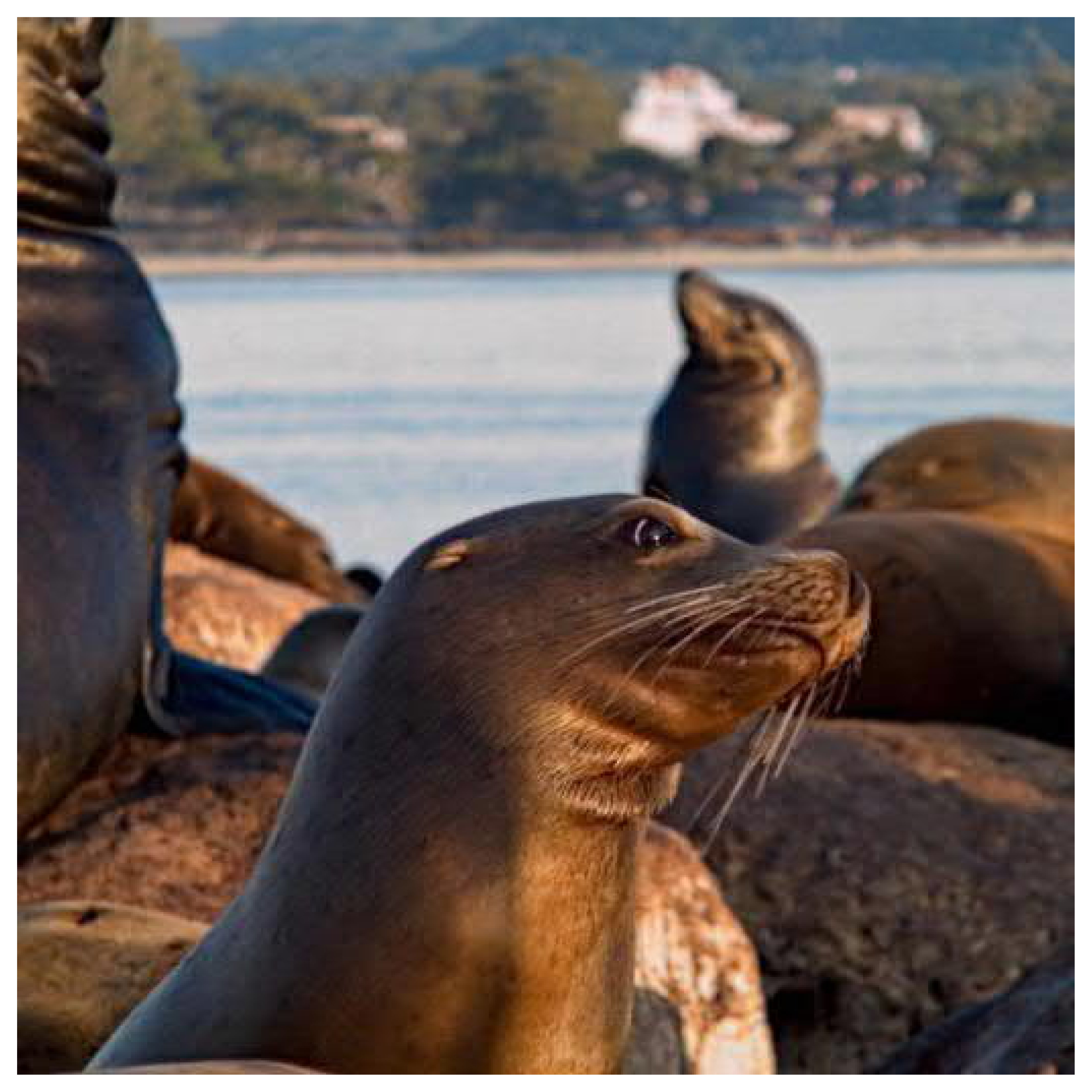}}	
	\end{minipage}%
	\begin{minipage}{0.3\linewidth}
		\centerline{\includegraphics[width=4cm,height=4cm]{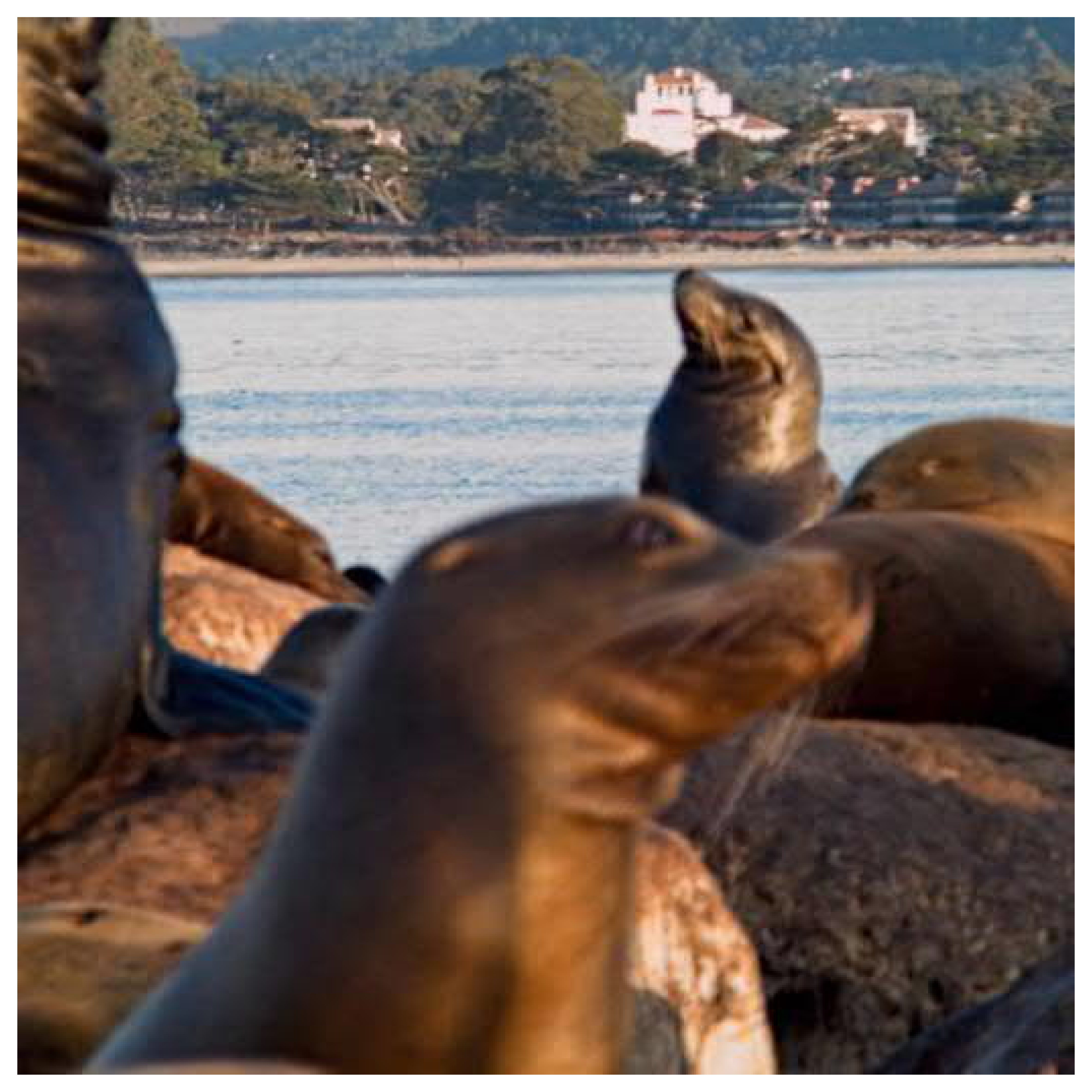}}
	\end{minipage}%
		\centering
		\caption{Input 2: triple color multi-focus images\cite{29}: (a) Diver, (b) Keyboard, (c) Folders, and (d) Seals.}
		\label{fig:sourceimages2}
		\vspace*{-5pt}
	\end{center}	
\end{figure}
\begin{figure}[htb]
	\begin{center}
		\begin{minipage}{0.05\linewidth}
			\centerline{a}	
		\end{minipage}%
		\begin{minipage}{0.45\linewidth}
			\centerline{\includegraphics[width=8cm,height=8cm]{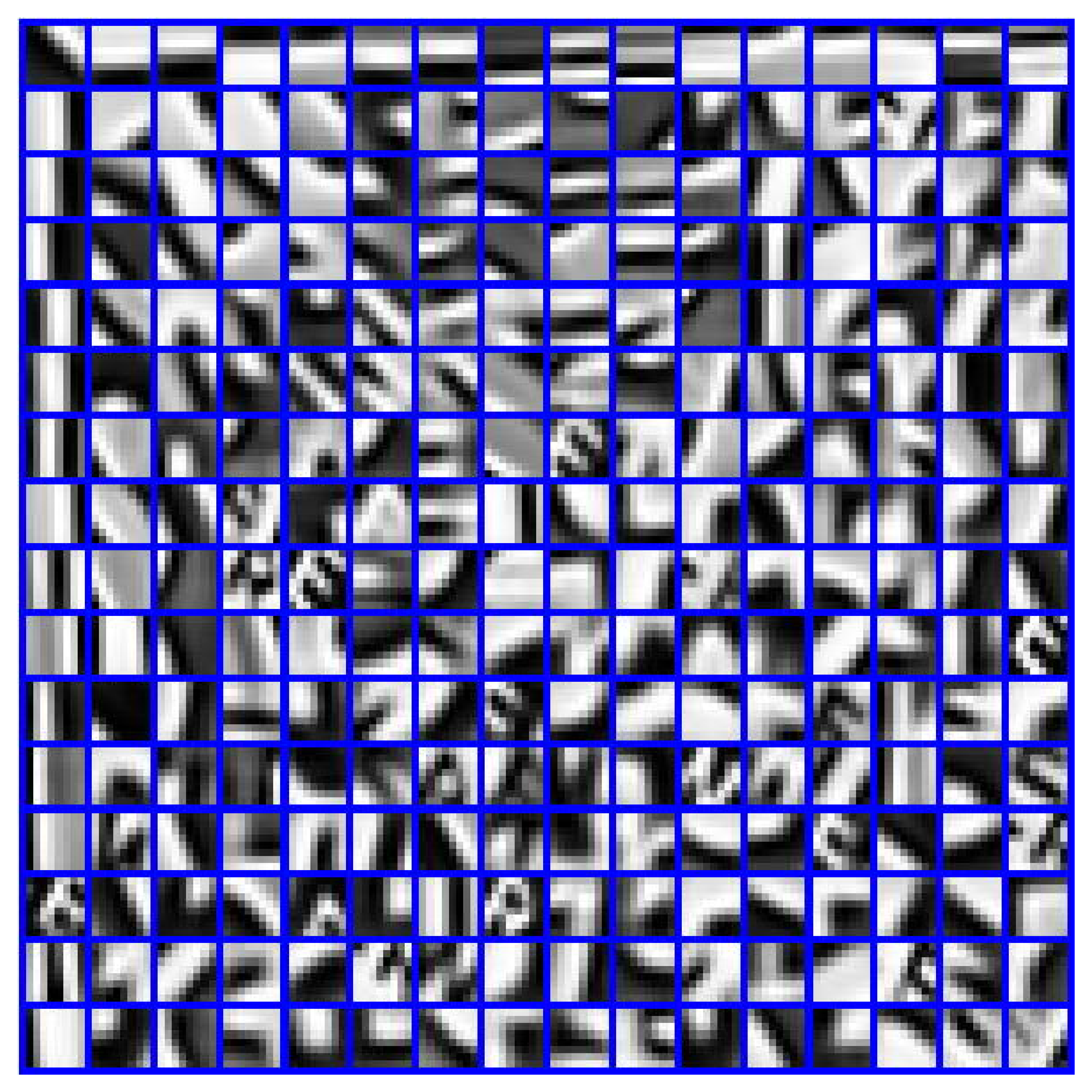}}
		\end{minipage}%
		\begin{minipage}{0.05\linewidth}
			\centerline{b}	
		\end{minipage}%
		\begin{minipage}{0.45\linewidth}
			\centerline{\includegraphics[width=8cm,height=8cm]{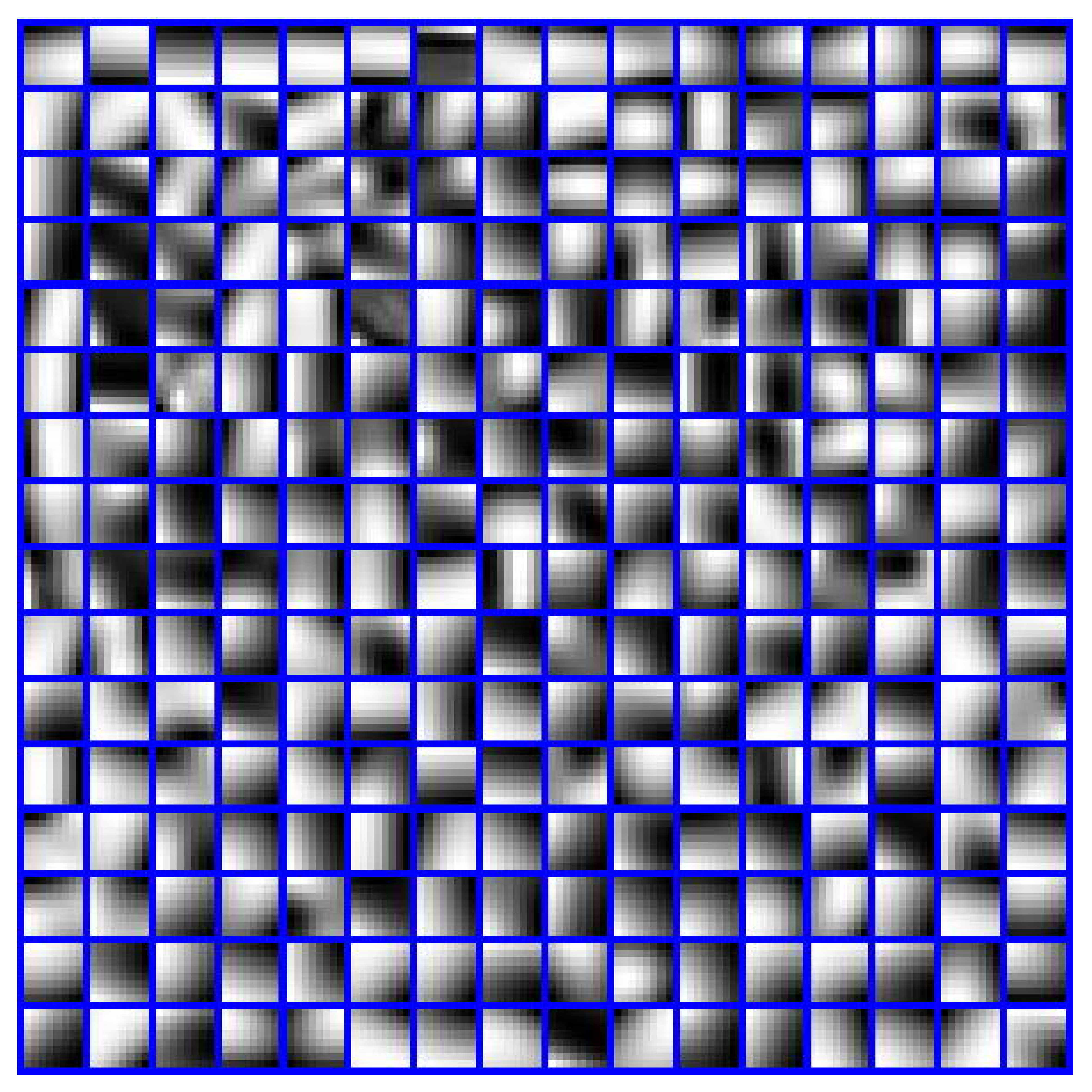}}
		\end{minipage}%
		\vfill
		\begin{minipage}{0.05\linewidth}
			\centerline{c}	
		\end{minipage}%
		\begin{minipage}{0.45\linewidth}
			\centerline{\includegraphics[width=8cm,height=8cm]{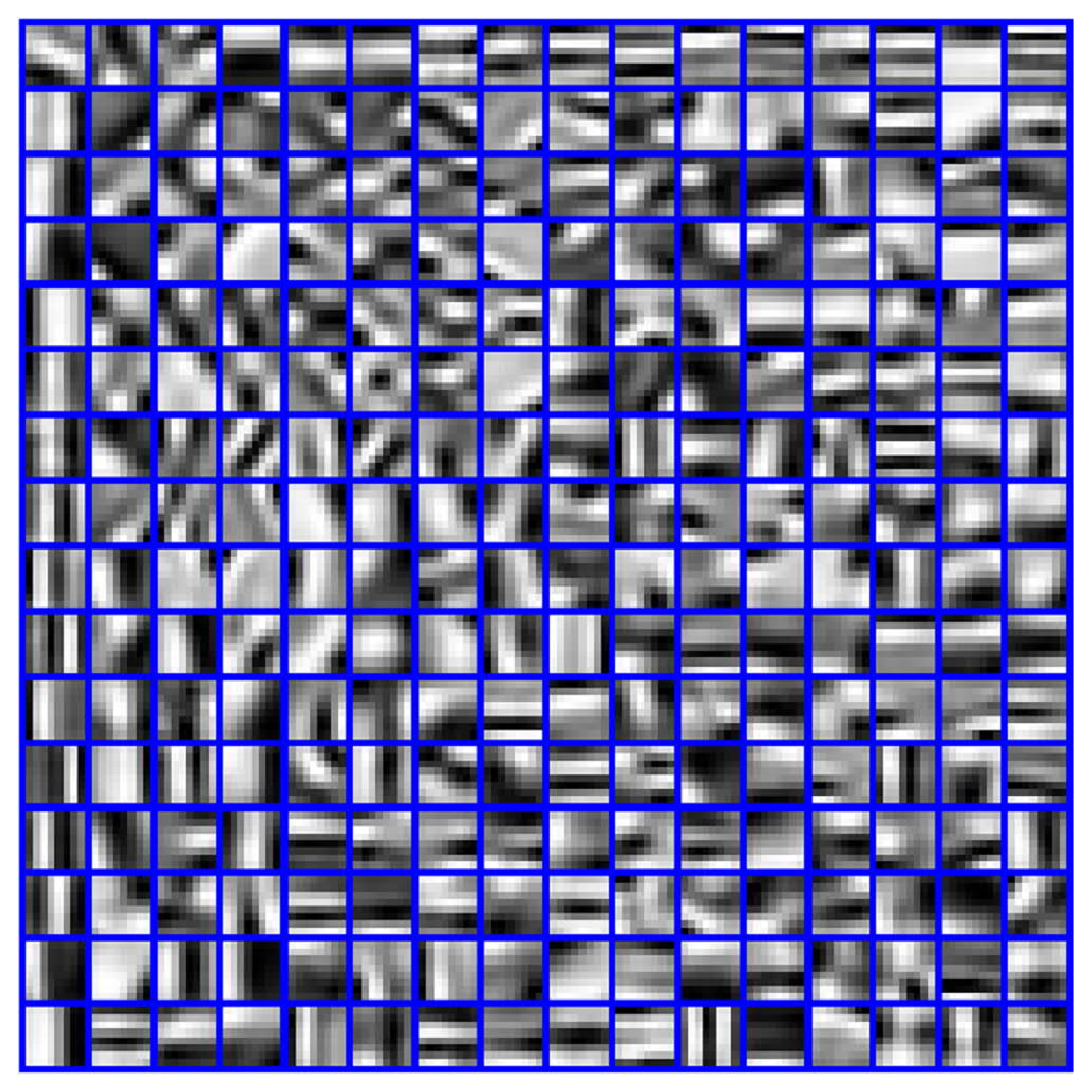}}
		\end{minipage}%
		\begin{minipage}{0.05\linewidth}
			\centerline{d}	
		\end{minipage}%
		\begin{minipage}{0.45\linewidth}
			\centerline{\includegraphics[width=8cm,height=8cm]{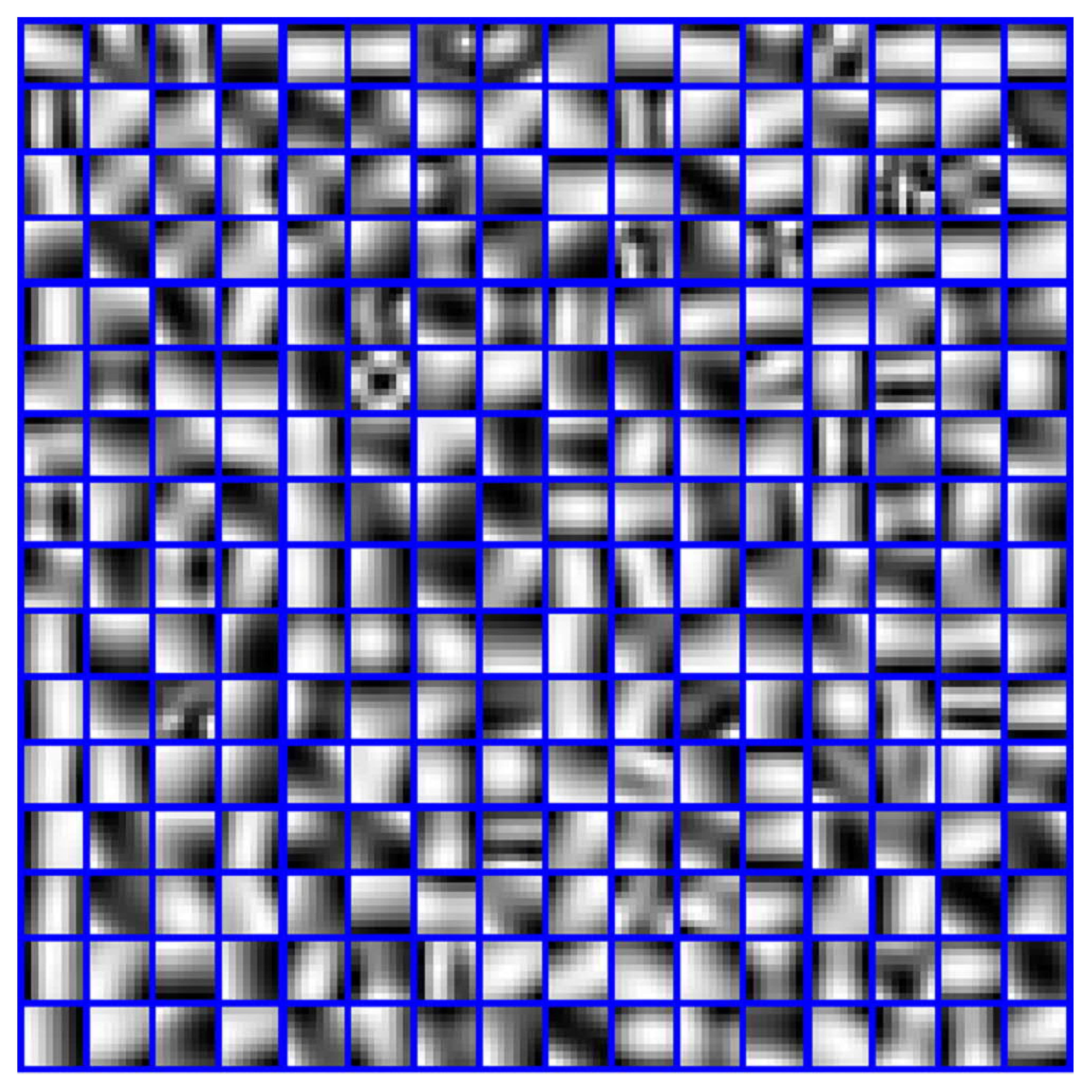}}
		\end{minipage}%
		\centering{
			\caption{Visual comparison between separately learned dictionaries: (a) $\boldsymbol{D}^{\rm F}$ and (b) $\boldsymbol{D}^{\rm B}$, and Coupled dictionary: (c) $\boldsymbol{D}^{\rm F}$ and (d) $\boldsymbol{D}^{\rm B}$.}	\label{fig:CoupledDict}}
	\end{center}
\end{figure}

Throughout all experiments, the parameters used in the methods are set as follows. For DWT method, the source images are decomposed to $3$ levels and the wavelet basis ``db1'' is applied. In DSIFT, an orientation histogram with 8 bins is used for quantizing the gradient information, and the feature vector is of $1\times128$. In implementation of SR-FM method, the Laplacian-energy is calculated as local focus measure and max-pooling is used for feature aggregation. Also the reconstruction phase is performed using overlapping patches, as the segmentation method applied to decision map in the work of~\cite{17} is not the focus of our work. All methods are assessed without applying any post-processing technique, since the problem of refining the decision map is independent from fusion methods. 

For fair comparison, all sparsity based algorithms are implemented using the same patch and dictionary size of $8^2$ and $64\times256$ ($64\times512$ for coupled), $7$ pixel overlap between neighboring patches and the tolerance error of~$\epsilon=0.1$. In addition, for dictionary learning, we execute 10 multiple dictionary update cycles.

For the proposed method, the visual results before and after global reconstruction are given separately. For global reconstruction, the ADMM algorithm of~\cite{27} is used with the regularization parameter $\lambda = 10^{-5}$, updating parameter $\rho = 1$, internal parameter $\gamma =1$. 

The learning data includes 30,000 pairs of patches taken from image parts indicated by rectangles in Fig.~\ref{fig:trainimages}. The images used for learning are taken from Lytro dataset\cite{17}. One coupled dictionary is learned and used for all experiments.

The input data includes six pairs of gray-scale multi-focus images (see Fig.~\ref{fig:sourceimages}) taken from standard multi-focus dataset \cite{17}\footnote{For the gray-scale inputs the available reference images in dataset~\cite{17} are used as perfectly fused all-in-focus images for measuring MSE and SSIM.} and four triple series (see Fig.~\ref{fig:sourceimages2}) of color multi-focus images taken from Lytro dataset. The size of gray-scale image pairs Doll, Clocks and Pepsi is $256 \times 256$, Lab and Disk are $480\times640$ and Jug is $256\times384$. The color images are of size $520 \times 520$. All experiments are performed on a PC running a Intel(R) Xeon(R) 3.40GHz~CPU. 

\subsection{Coupled versus Separately Learned Dictionaries}
\label{CoupledVsSeparate}
The coupled dictionary used in the experiments and two dictionaries separately learned over the same focused and blurred learning data using K-SVD are visualized in Fig.~\ref{fig:CoupledDict} for comparison.

\begin{figure*}[!htb]
	\begin{center}		
		\begin{minipage}{0.5\linewidth}
			\centerline{\includegraphics[width=7.5cm,height=5.5cm]{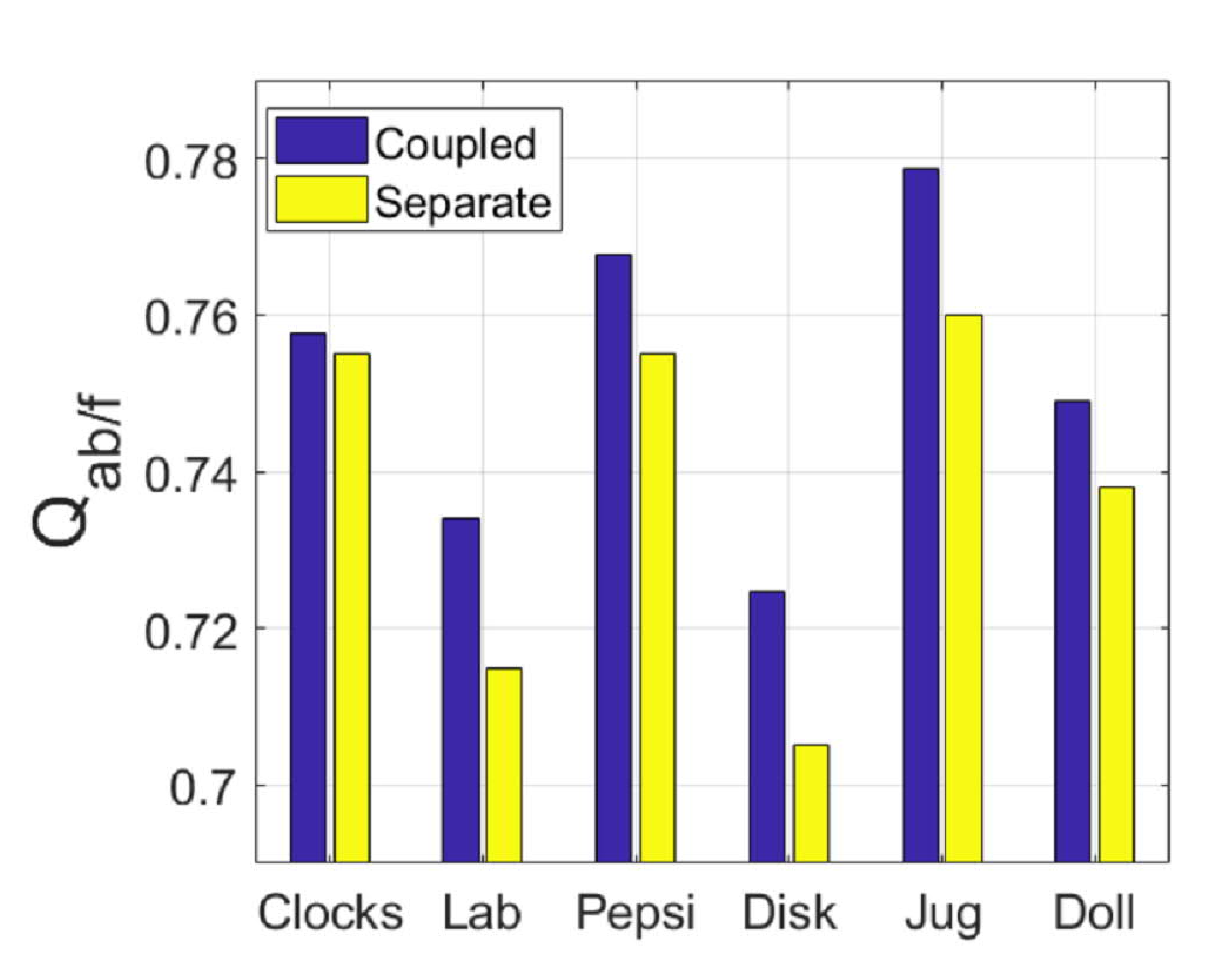}}	
		\end{minipage}%
		\begin{minipage}{0.5\linewidth}
			\centerline{\includegraphics[width=7.5cm,height=5.5cm]{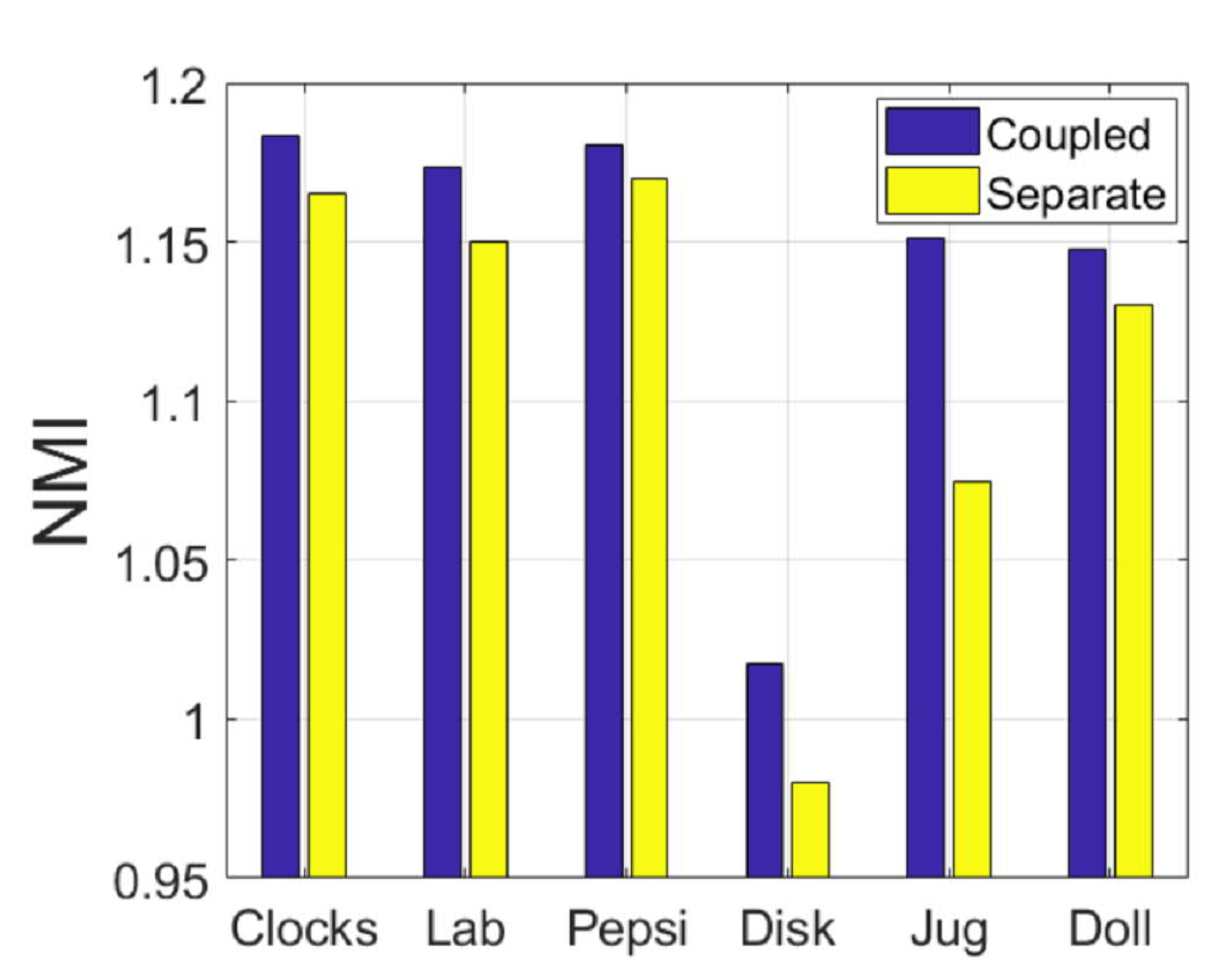}}	
		\end{minipage}%
		\caption{Comparing fusion performances, using proposed method, over coupled dictionary and separately learned dictionaries.}
		\label{fig:comparedic}
	\end{center}
\end{figure*}
\begin{figure*}[htb]
	\begin{minipage}{0.05\linewidth}
		\centerline{a}	
	\end{minipage}%
	\begin{minipage}{0.20\linewidth}
		\centerline{\includegraphics[width=4cm,height=4cm]{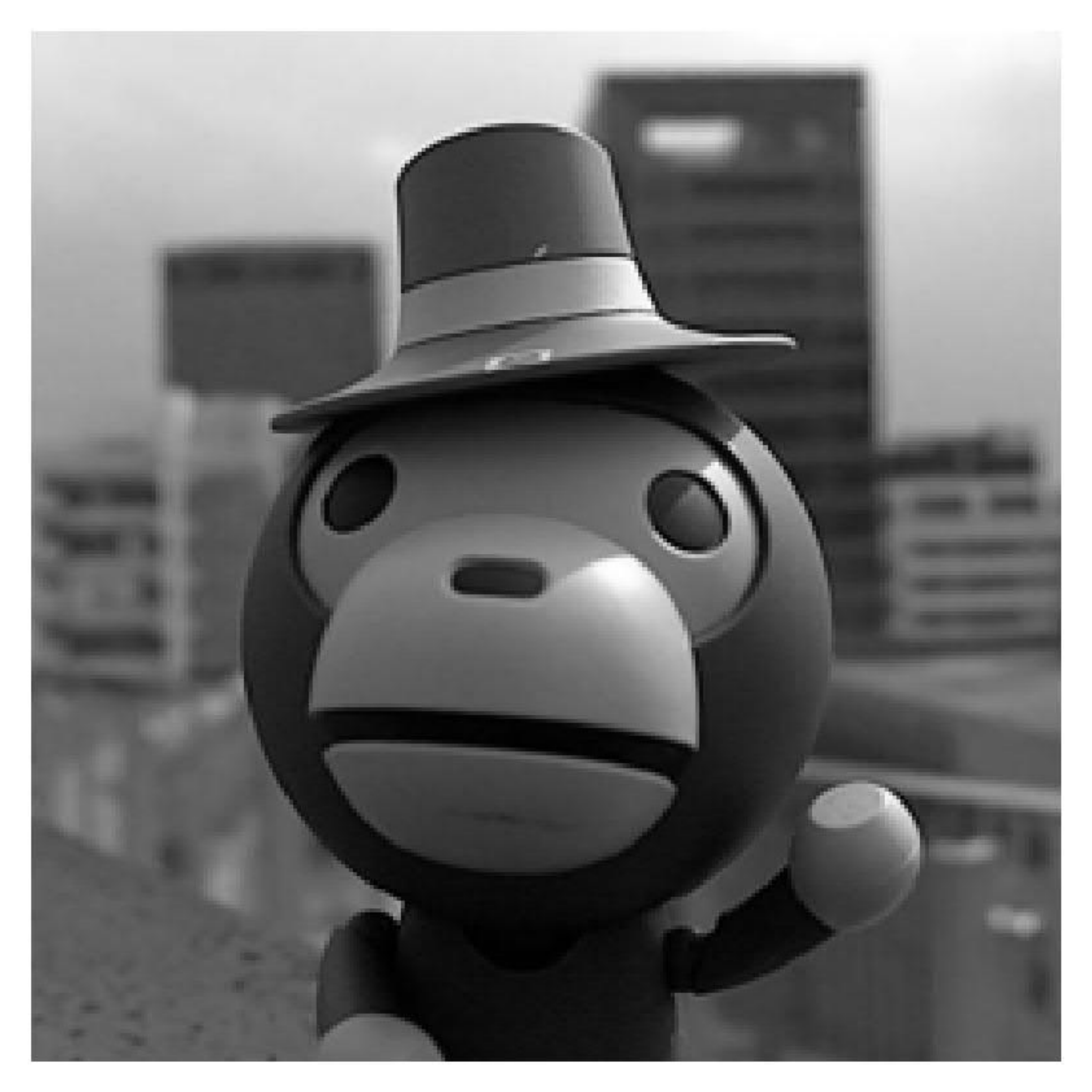}}
	\end{minipage}%
	\begin{minipage}{0.05\linewidth}
		\centerline{b}	
	\end{minipage}%
	\begin{minipage}{0.20\linewidth}
		\centerline{\includegraphics[width=4cm,height=4cm]{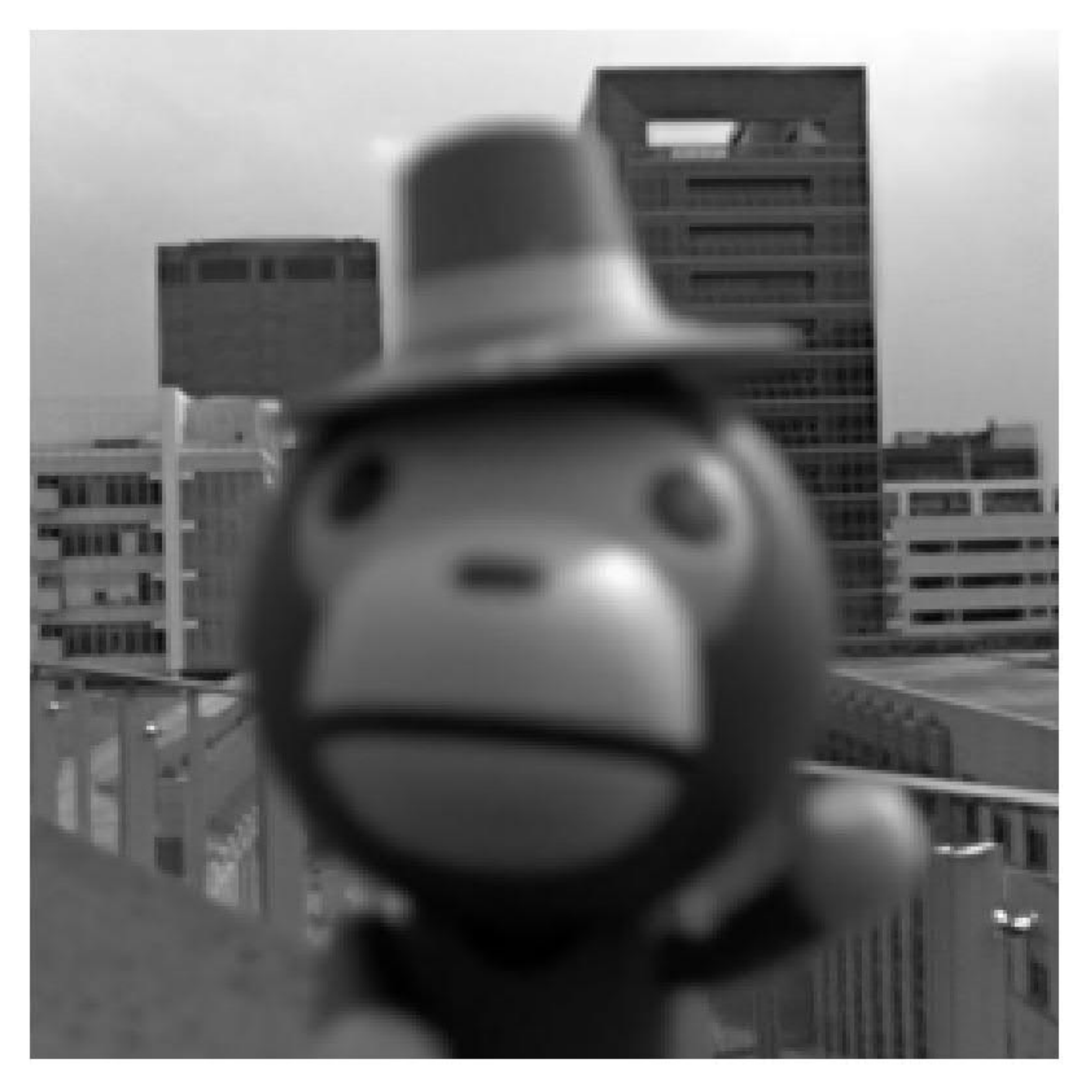}}
	\end{minipage}%
	\begin{minipage}{0.05\linewidth}
		\centerline{c}	
	\end{minipage}%
	\begin{minipage}{0.20\linewidth}
		\centerline{\includegraphics[width=4cm,height=4cm]{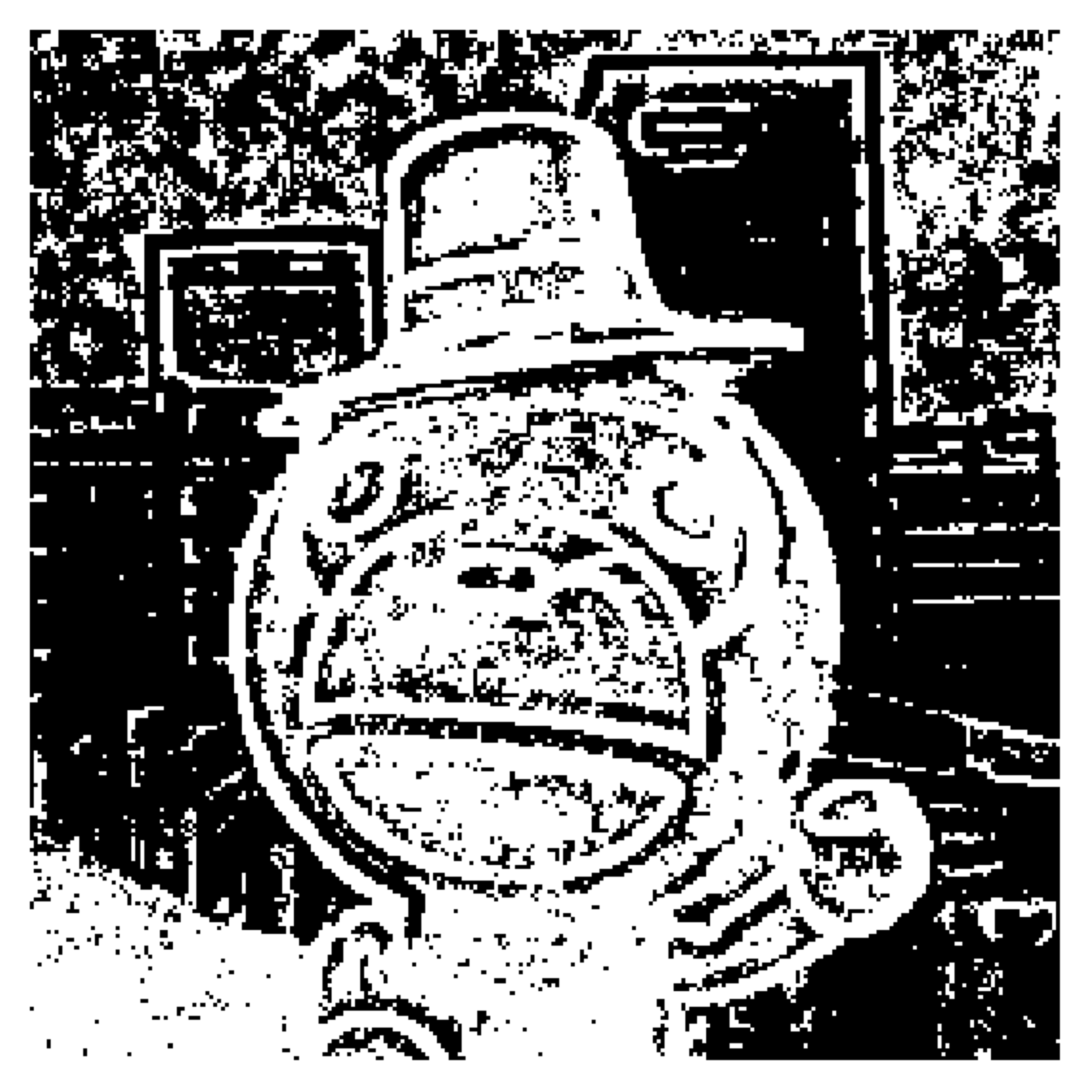}}
	\end{minipage}%
	\begin{minipage}{0.05\linewidth}
		\centerline{d}	
	\end{minipage}%
	\begin{minipage}{0.20\linewidth}
		\centerline{\includegraphics[width=4cm,height=4cm]{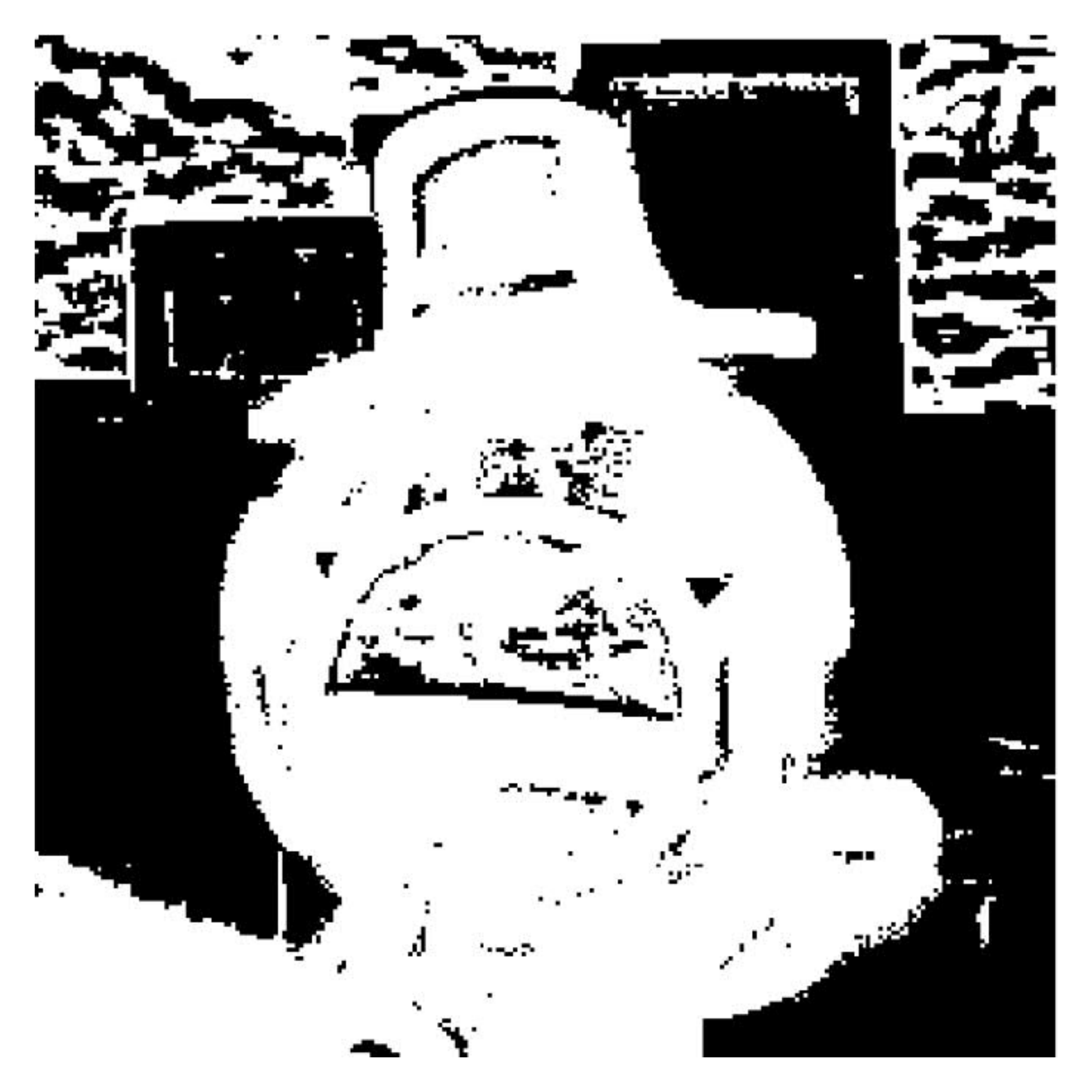}}
	\end{minipage}%
	\centering{
		\caption{Comparing masks obtained using (c) single dictionary (SR-KSVD) and (d) coupled dictionary from source images in (a) and (b). The images are taken from lytro multi-focus dataset~\cite{29}.}	\label{fig:maskcompare}}
	
\end{figure*}
\begin{figure*}[!htb]
	\begin{minipage}{0.05\linewidth}
		\centerline{a}	
	\end{minipage}%
	\begin{minipage}{0.20\linewidth}
		\centerline{\includegraphics[width=4cm,height=4cm]{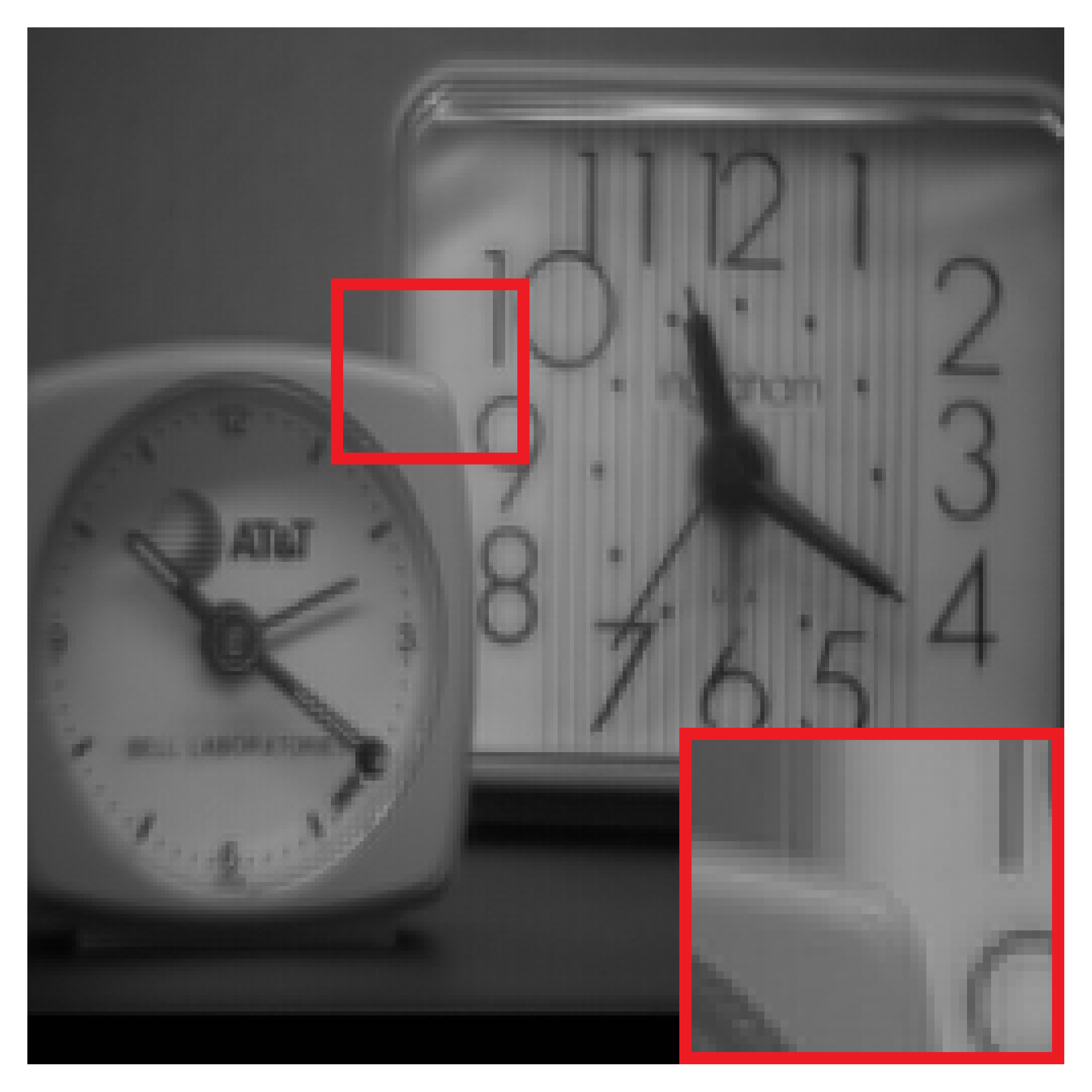}}	
	\end{minipage}%
	\begin{minipage}{0.05\linewidth}
		\centerline{b}	
	\end{minipage}%
	\begin{minipage}{0.20\linewidth}
		\centerline{\includegraphics[width=4cm,height=4cm]{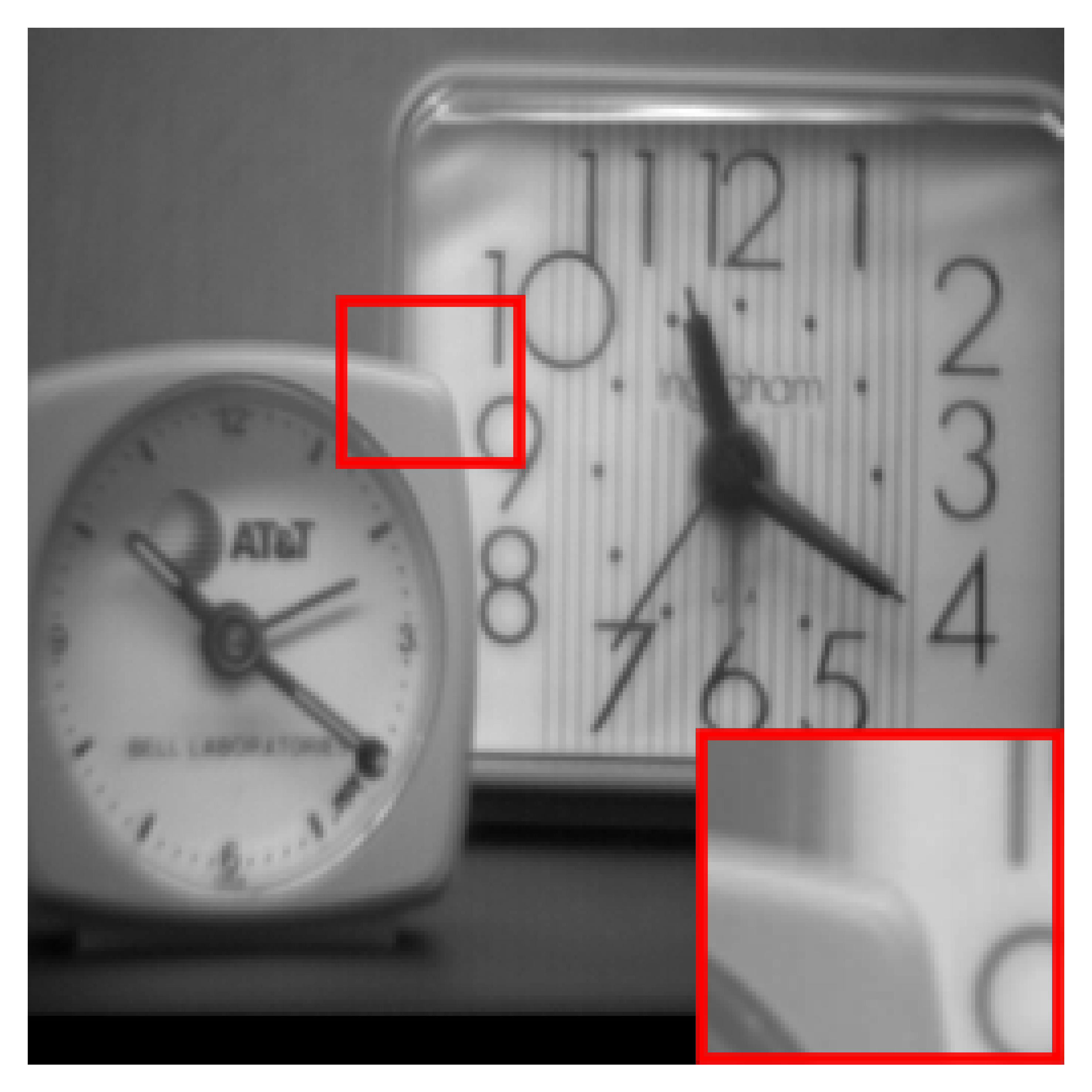}}	
	\end{minipage}%
	\begin{minipage}{0.05\linewidth}
		\centerline{c}	
	\end{minipage}%
	\begin{minipage}{0.20\linewidth}
		\centerline{\includegraphics[width=4cm,height=4cm]{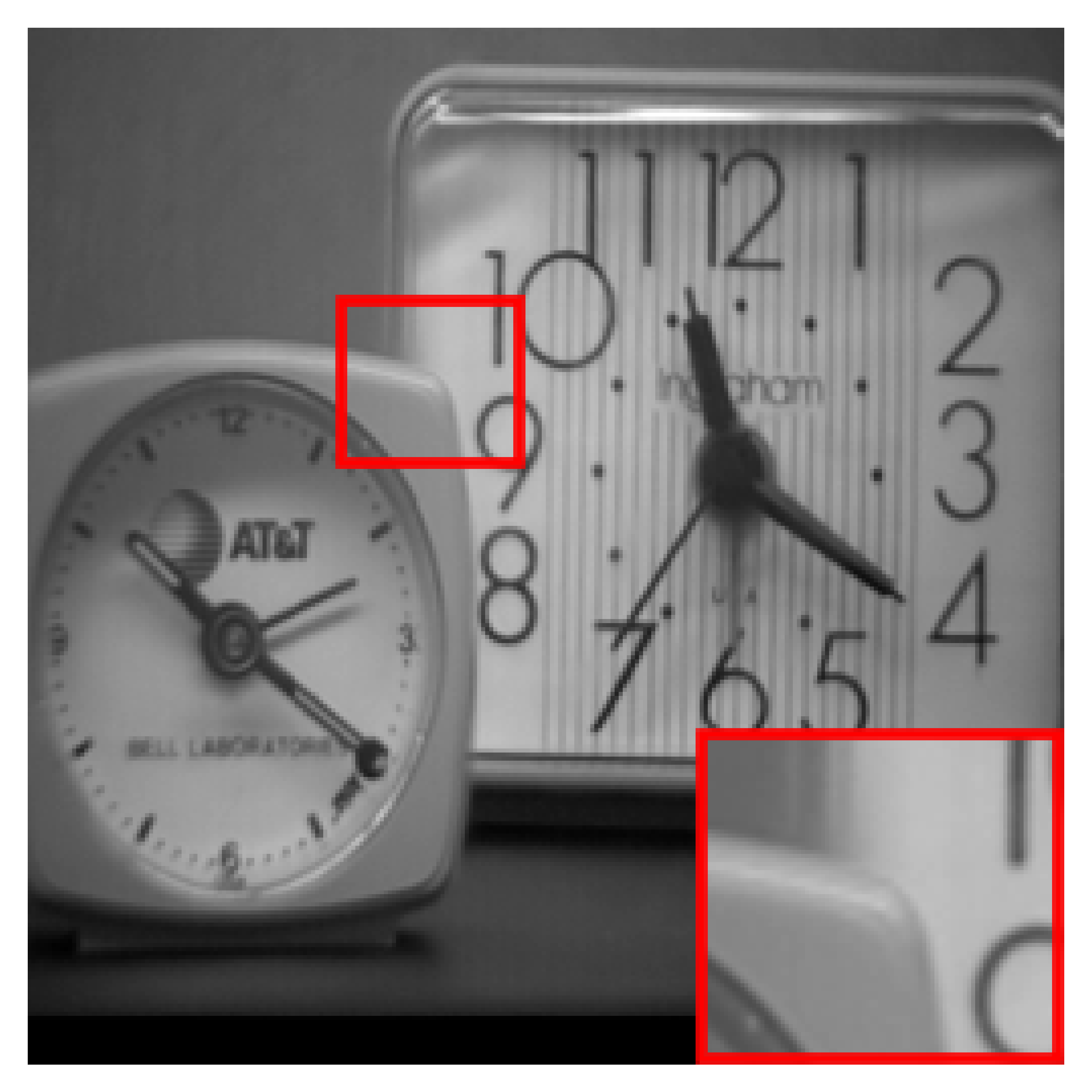}}	
	\end{minipage}%
	\begin{minipage}{0.05\linewidth}
		\centerline{d}	
	\end{minipage}%
	\begin{minipage}{0.20\linewidth}
		\centerline{\includegraphics[width=4cm,height=4cm]{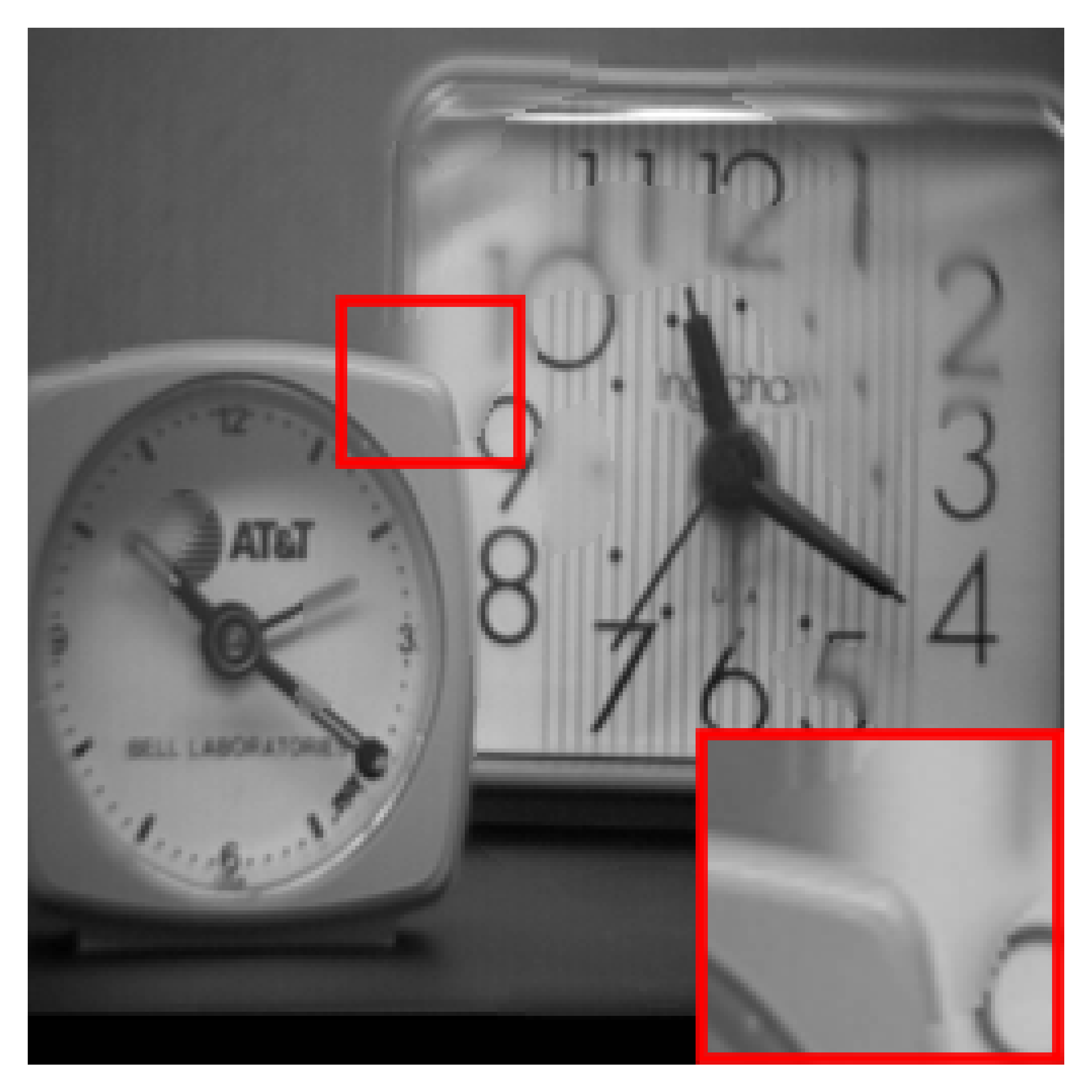}}
	\end{minipage}%
	
	\vfill
	\begin{minipage}{0.05\linewidth}
		\centerline{e}	
	\end{minipage}%
	\begin{minipage}{0.20\linewidth}
		\centerline{\includegraphics[width=4cm,height=4cm]{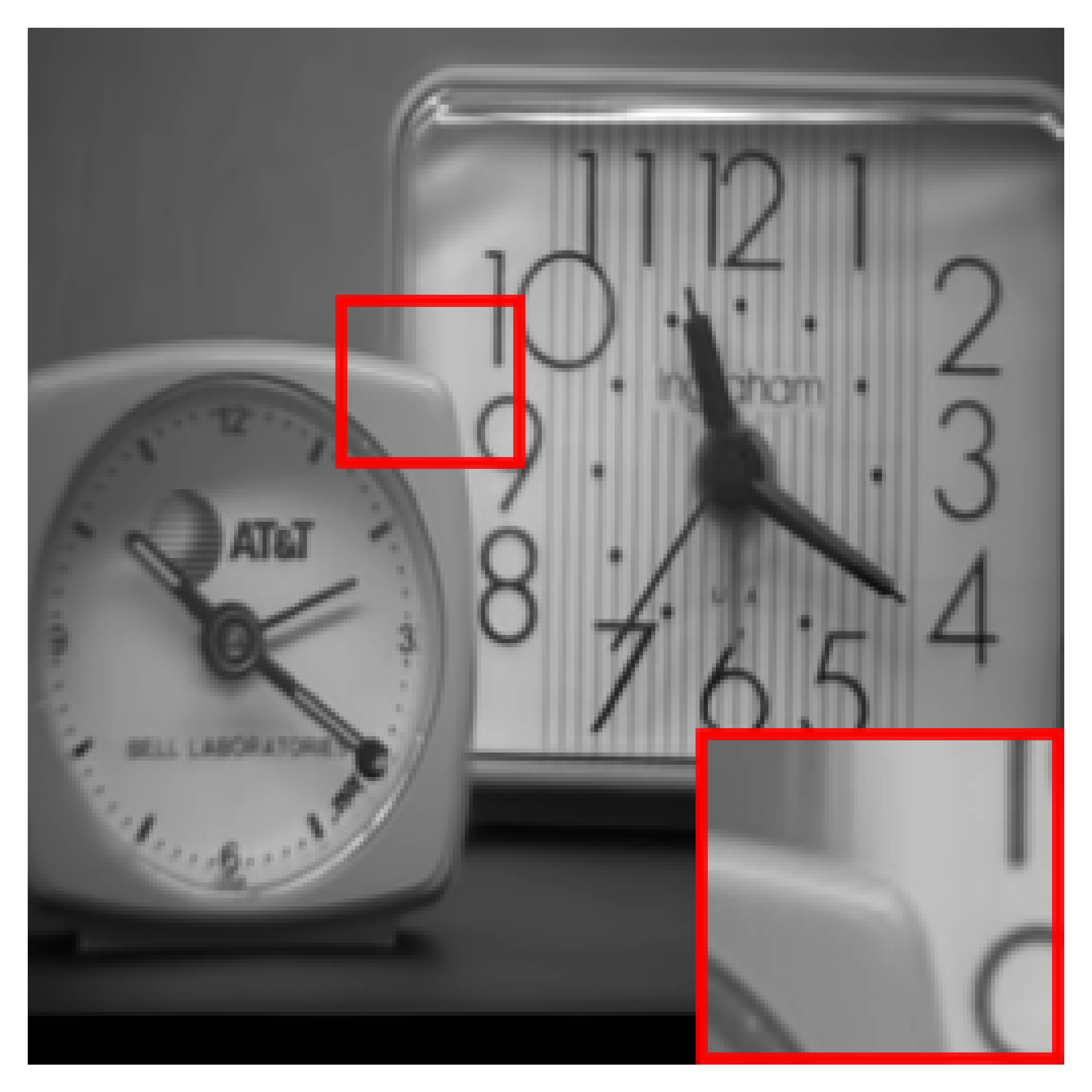}}	
	\end{minipage}%
	\begin{minipage}{0.05\linewidth}
		\centerline{f}	
	\end{minipage}%
	\begin{minipage}{0.20\linewidth}
		\centerline{\includegraphics[width=4cm,height=4cm]{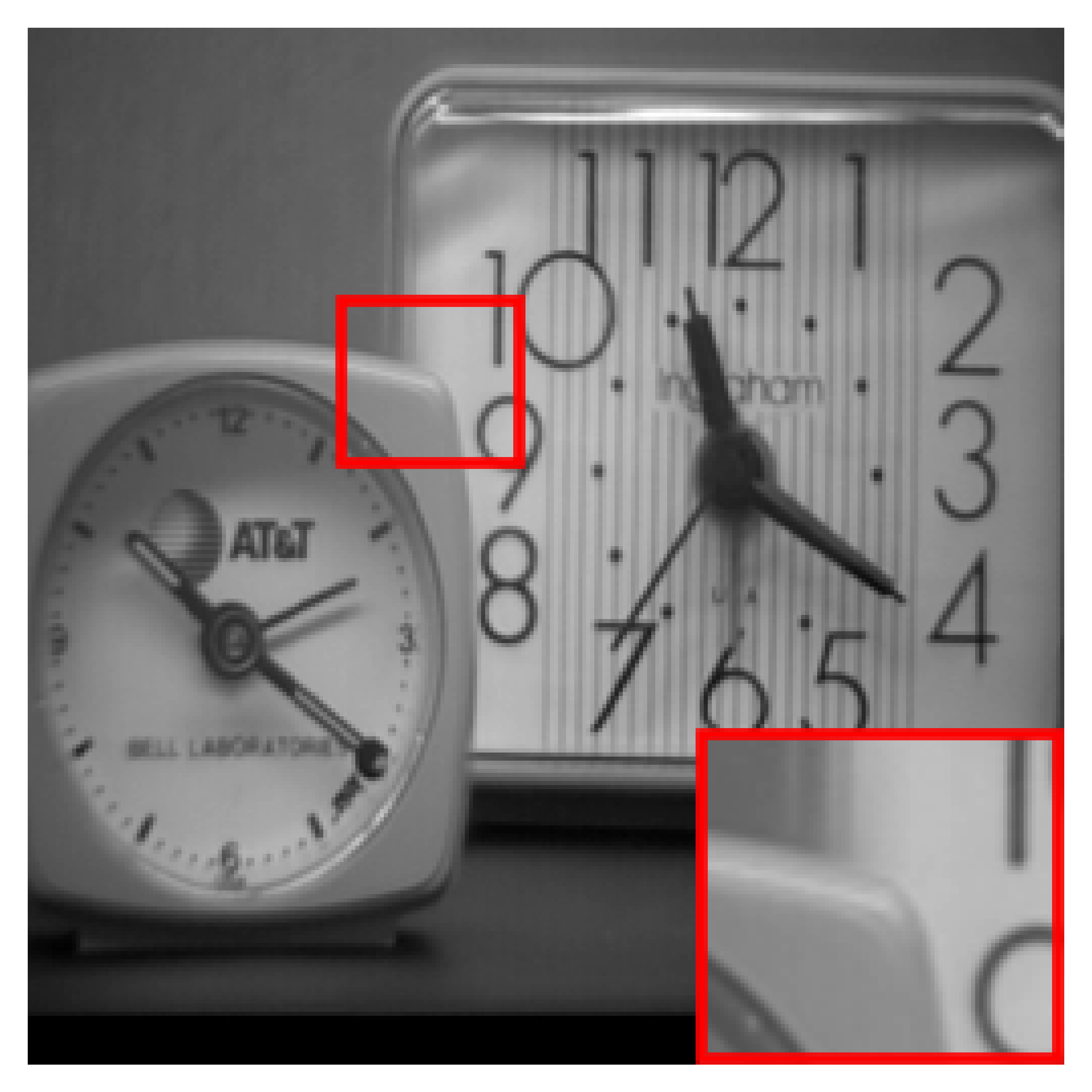}}	
	\end{minipage}%
	\begin{minipage}{0.05\linewidth}
		\centerline{g}	
	\end{minipage}%
	\begin{minipage}{0.20\linewidth}
		\centerline{\includegraphics[width=4cm,height=4cm]{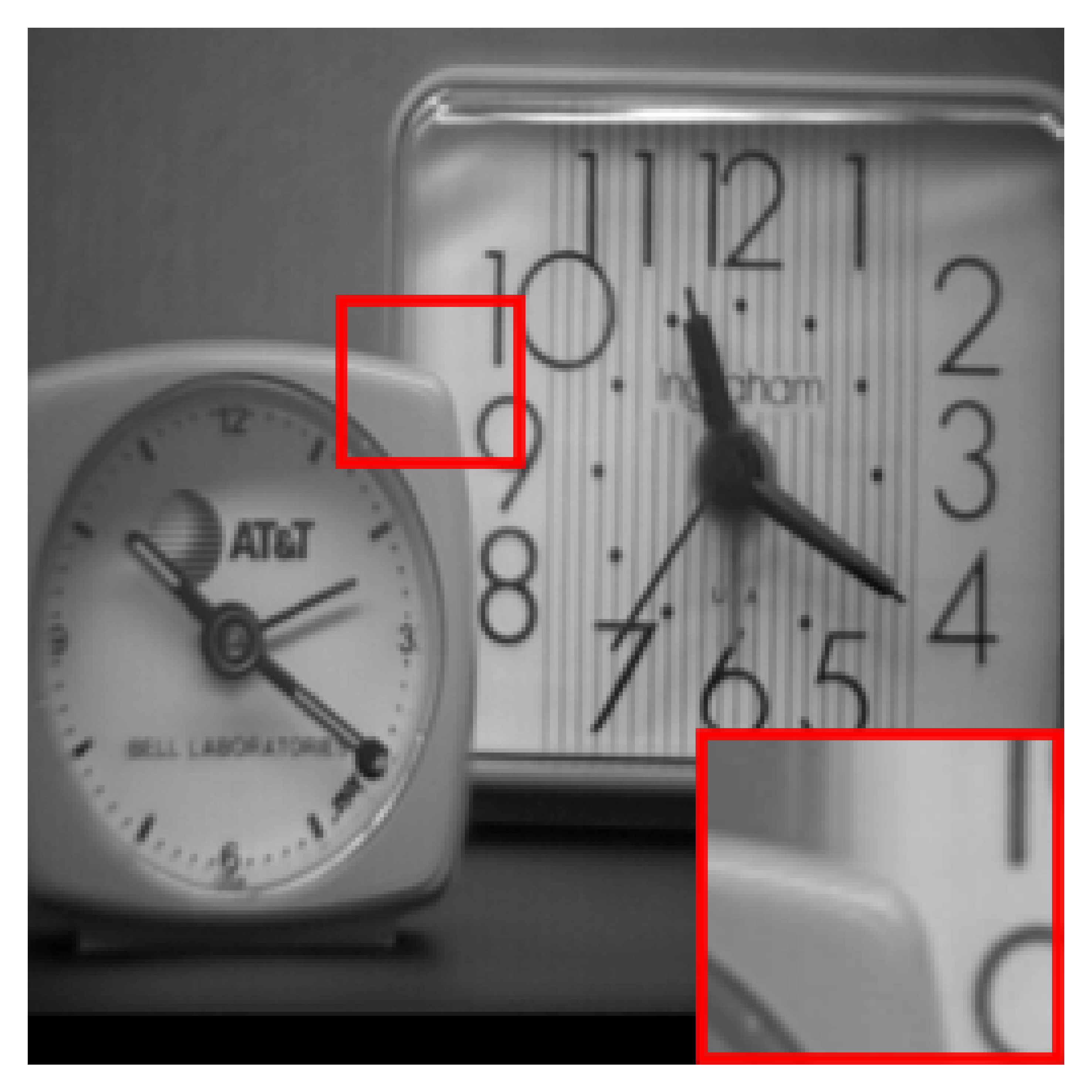}}	
	\end{minipage}%
	\begin{minipage}{0.05\linewidth}
		\centerline{h}	
	\end{minipage}%
	\begin{minipage}{0.20\linewidth}
		\centerline{\includegraphics[width=4cm,height=4cm]{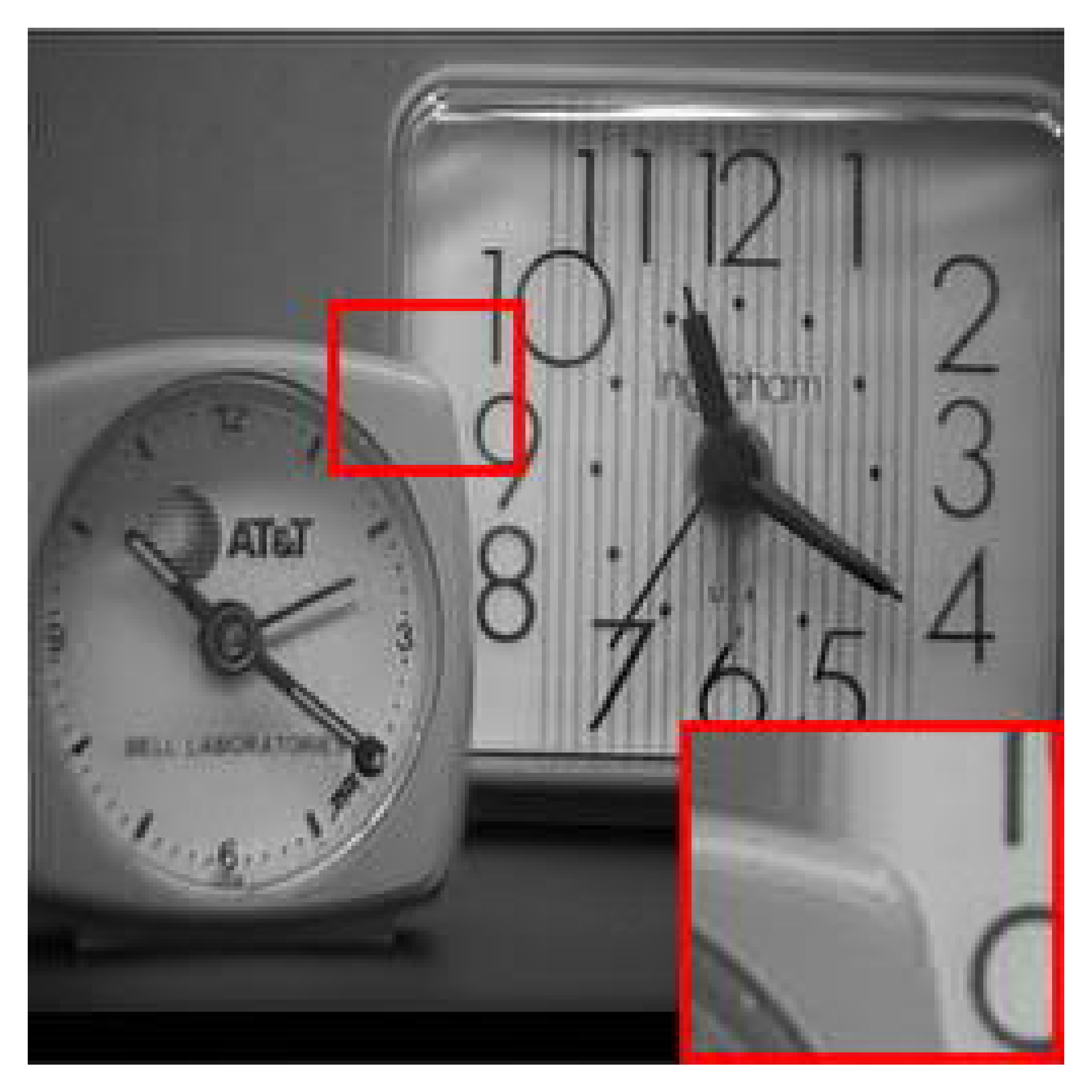}}
	\end{minipage}%
	\vfill
	\centering
	\caption{Fusion result for multi-focus images ''Clocks'', obtained by methods: DWT (a), PCA (b), SR-FM (c), DSIFT (d), SR-CM (e), SR-KSVD (f), proposed method (g), and the proposed after global reconstruction (h)}
	\label{fig:clocks}
	\vspace*{-5pt}
\end{figure*}

The pairwise correlations between the atoms of $\boldsymbol{D}^{\rm F}$ and $\boldsymbol{D}^{\rm B}$ (see Figs.~\ref{fig:CoupledDict}.(c) and (d)) are obtained by enforcing identical sparse representations through the dictionary learning \cite{25}. These pairwise correlations ensure that $\boldsymbol{D}^{\rm F}$ and $\boldsymbol{D}^{\rm B}$ represent corresponding focused and blurred features. 
\begin{figure*}[!htb]
	\begin{minipage}{0.05\linewidth}
		\centerline{a}	
	\end{minipage}%
	\begin{minipage}{0.2\linewidth}
		\centerline{\includegraphics[width=4cm,height=4cm]{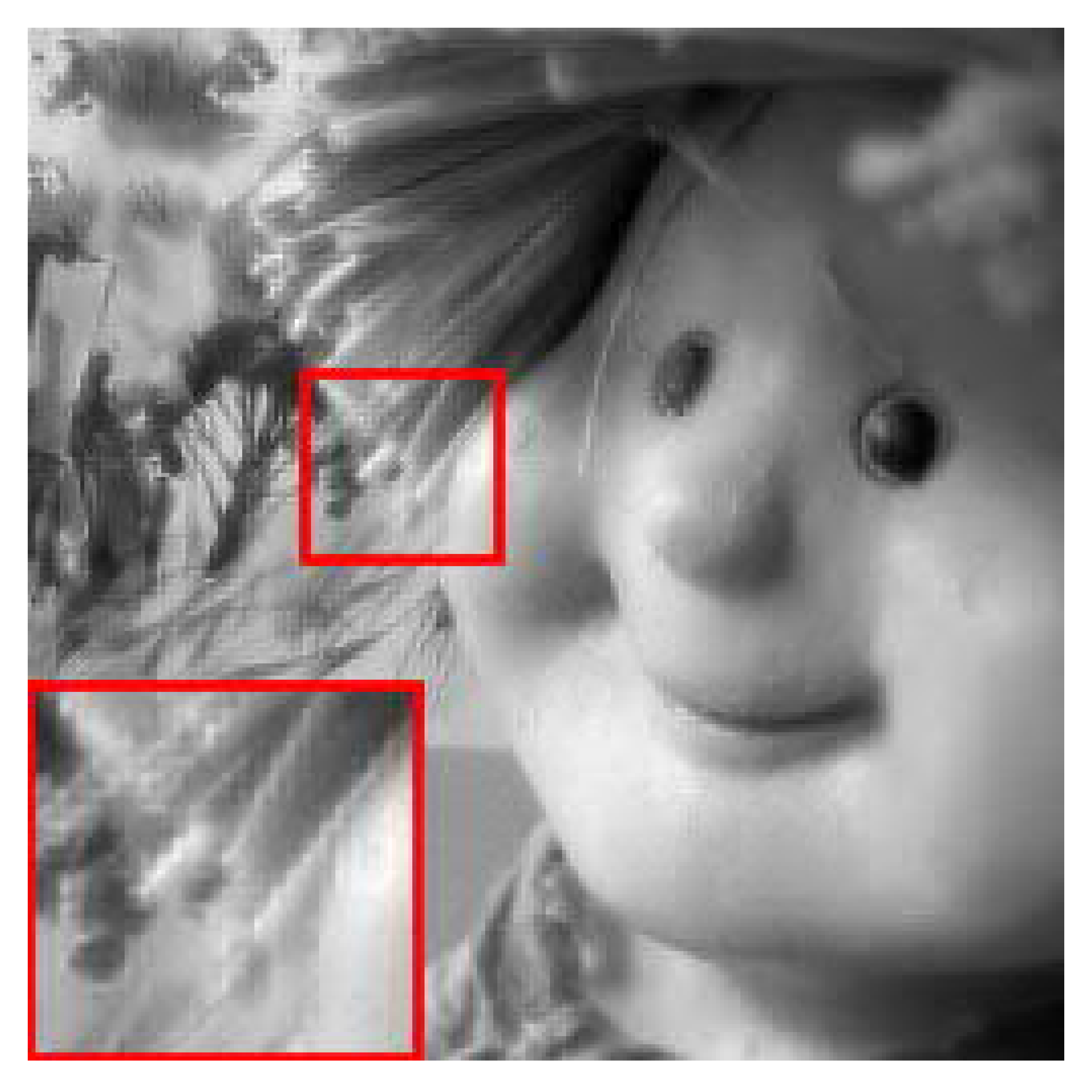}}		
	\end{minipage}%
	\begin{minipage}{0.05\linewidth}
		\centerline{b}	
	\end{minipage}%
	\begin{minipage}{0.2\linewidth}
		\centerline{\includegraphics[width=4cm,height=4cm]{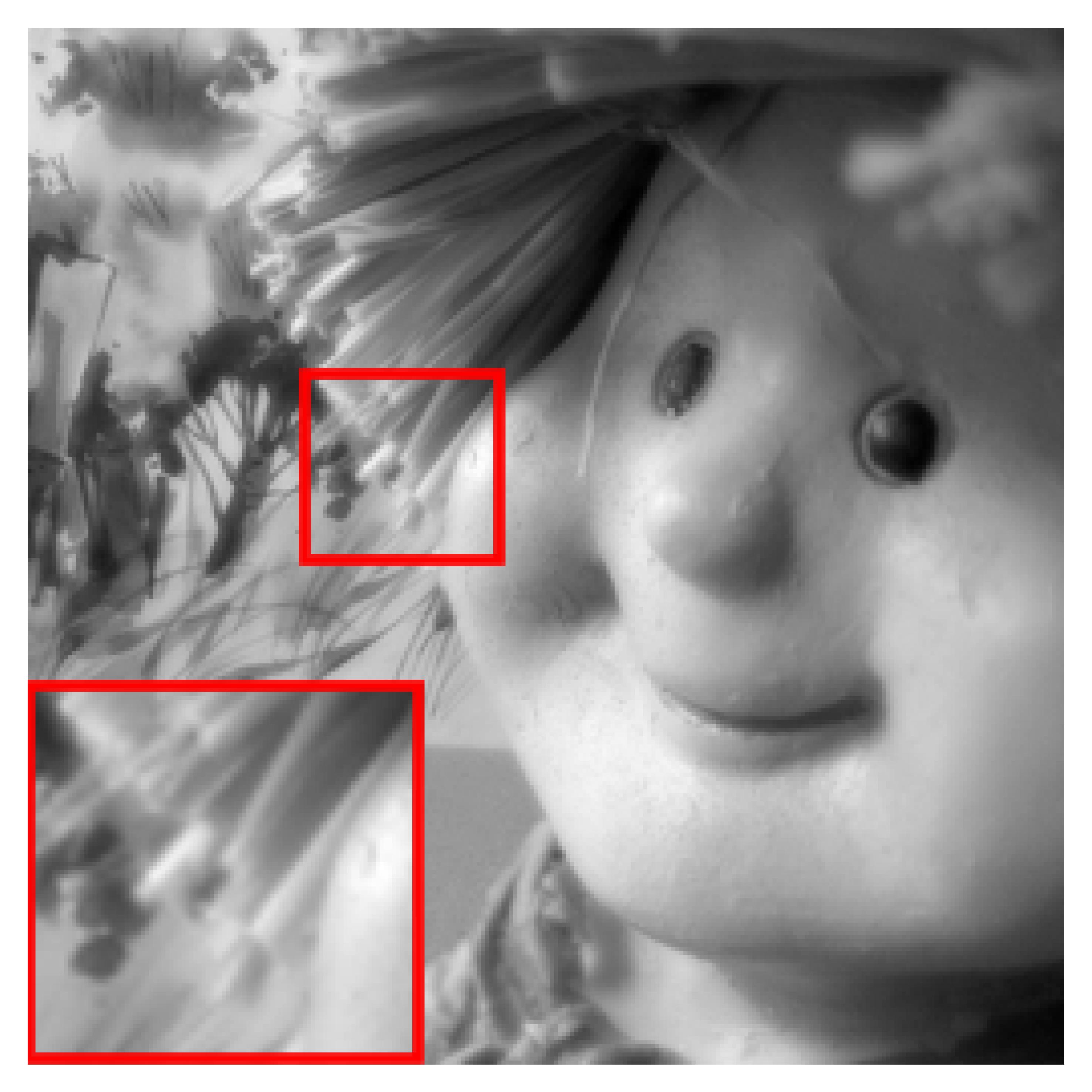}}		
	\end{minipage}%
	\begin{minipage}{0.05\linewidth}
		\centerline{c}	
	\end{minipage}%
	\begin{minipage}{0.2\linewidth}
		\centerline{\includegraphics[width=4cm,height=4cm]{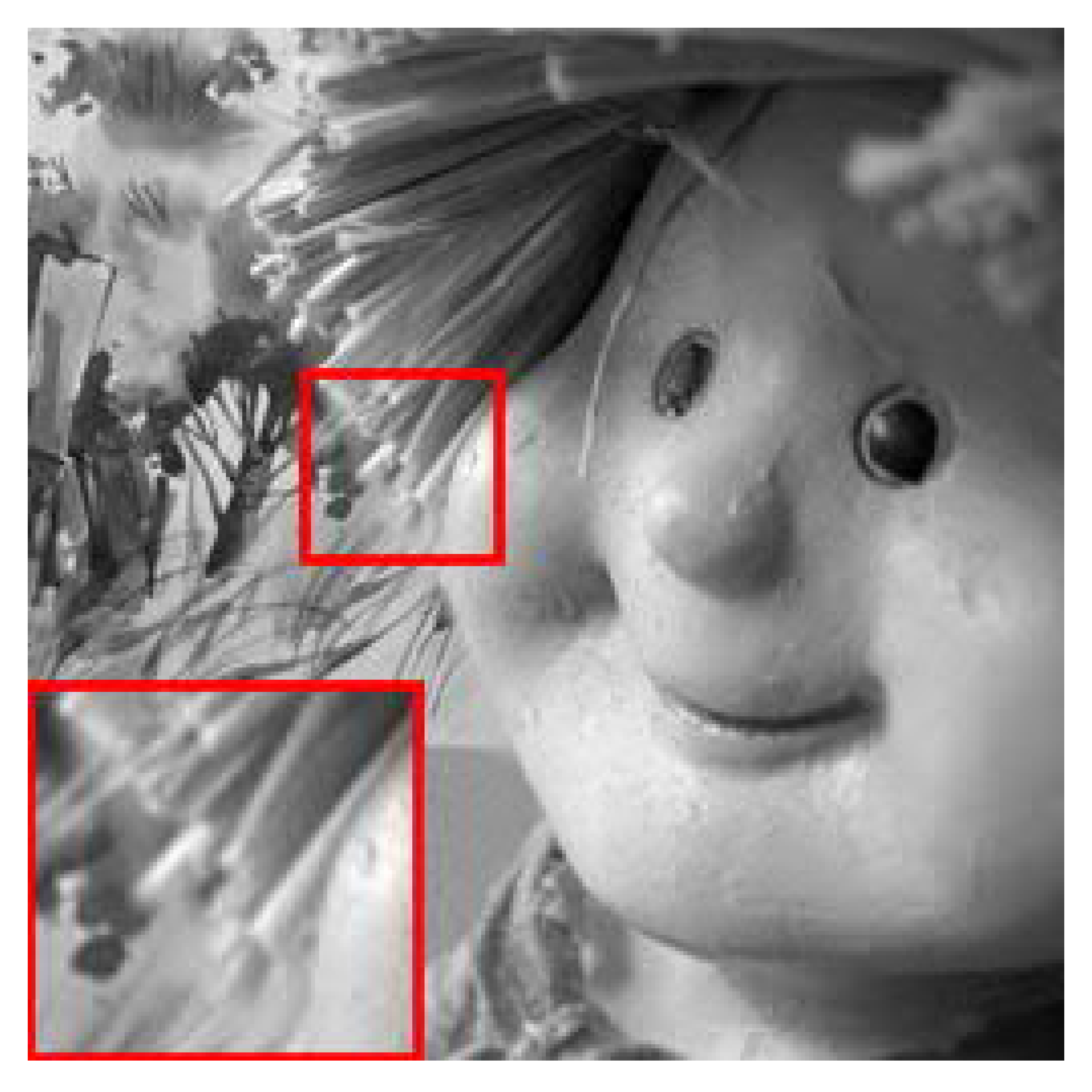}}		
	\end{minipage}%
	\begin{minipage}{0.05\linewidth}
		\centerline{d}	
	\end{minipage}%
	\begin{minipage}{0.2\linewidth}
		\centerline{\includegraphics[width=4cm,height=4cm]{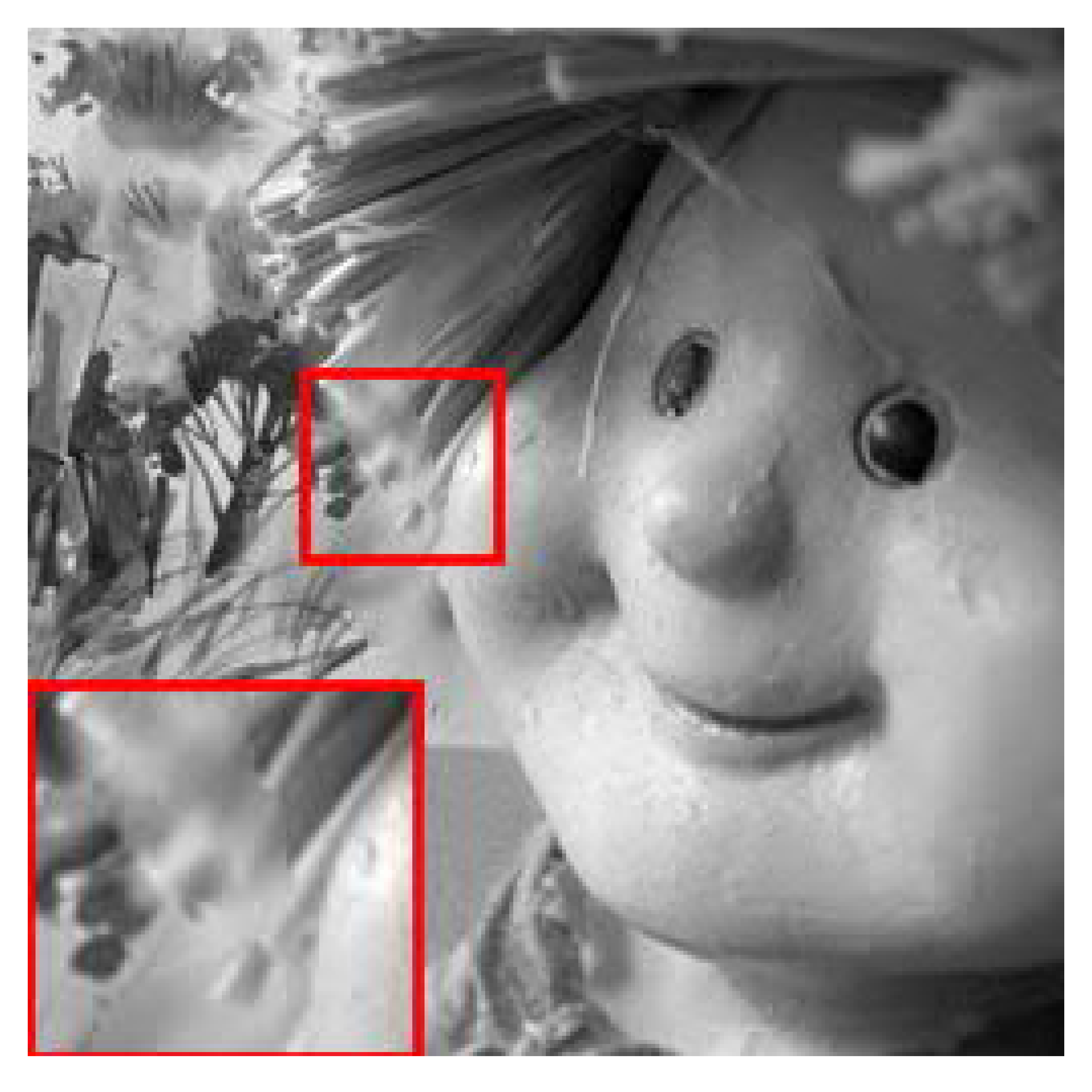}}
	\end{minipage}%
	
	\vfill
	\begin{minipage}{0.05\linewidth}
		\centerline{e}	
	\end{minipage}%
	\begin{minipage}{0.2\linewidth}
		\centerline{\includegraphics[width=4cm,height=4cm]{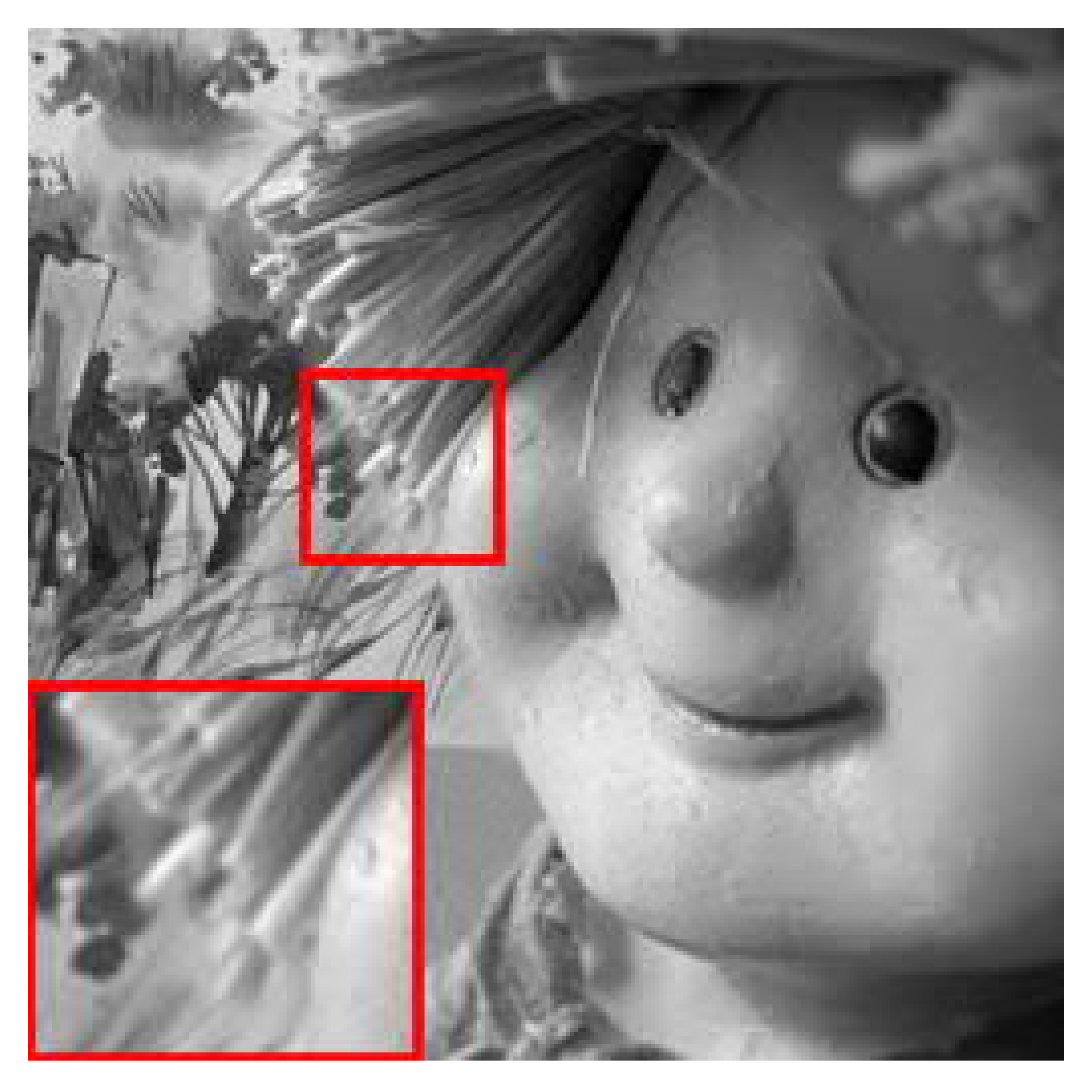}}		
	\end{minipage}%
	\begin{minipage}{0.05\linewidth}
		\centerline{f}	
	\end{minipage}%
	\begin{minipage}{0.2\linewidth}
		\centerline{\includegraphics[width=4cm,height=4cm]{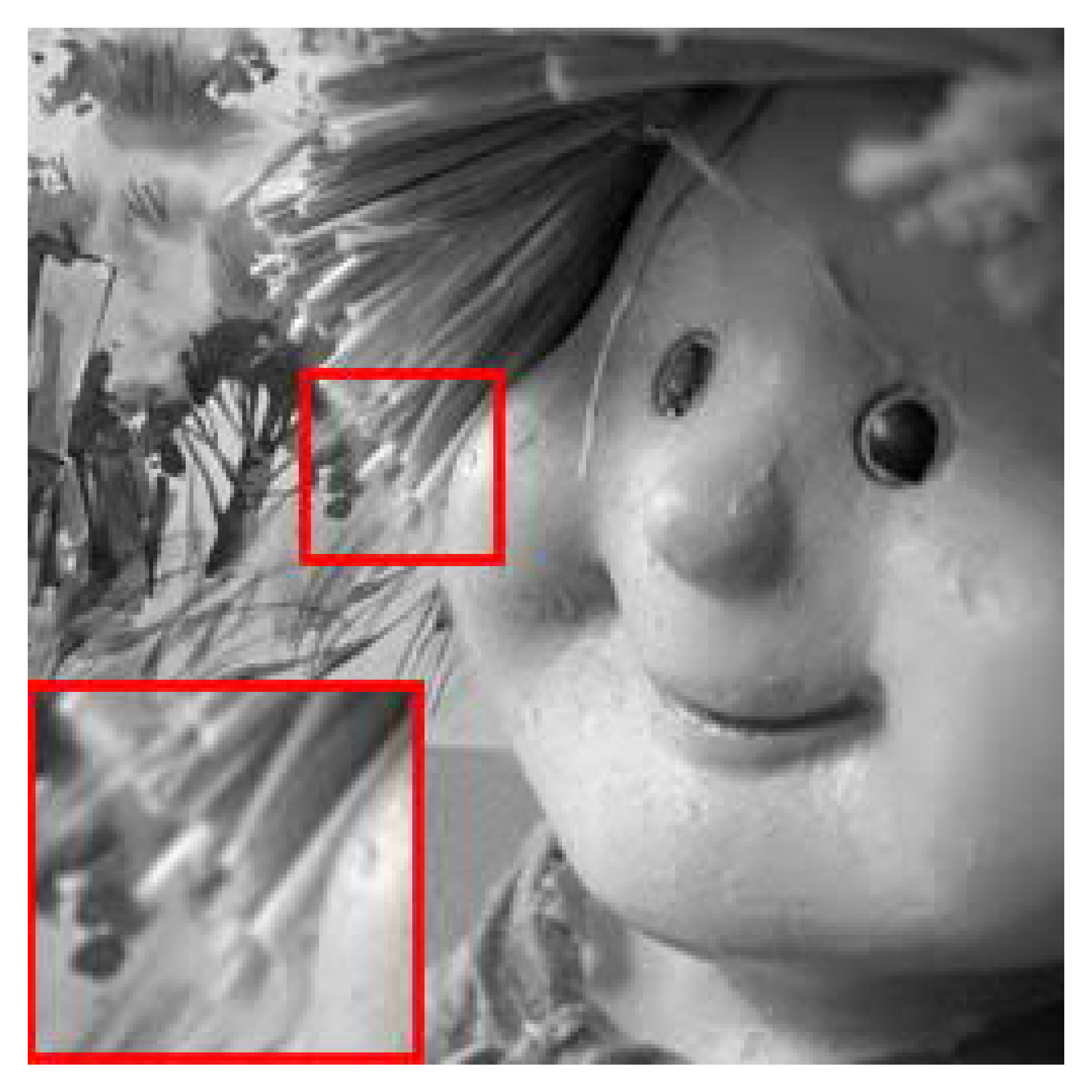}}		
	\end{minipage}%
	\begin{minipage}{0.05\linewidth}
		\centerline{g}	
	\end{minipage}%
	\begin{minipage}{0.2\linewidth}
		\centerline{\includegraphics[width=4cm,height=4cm]{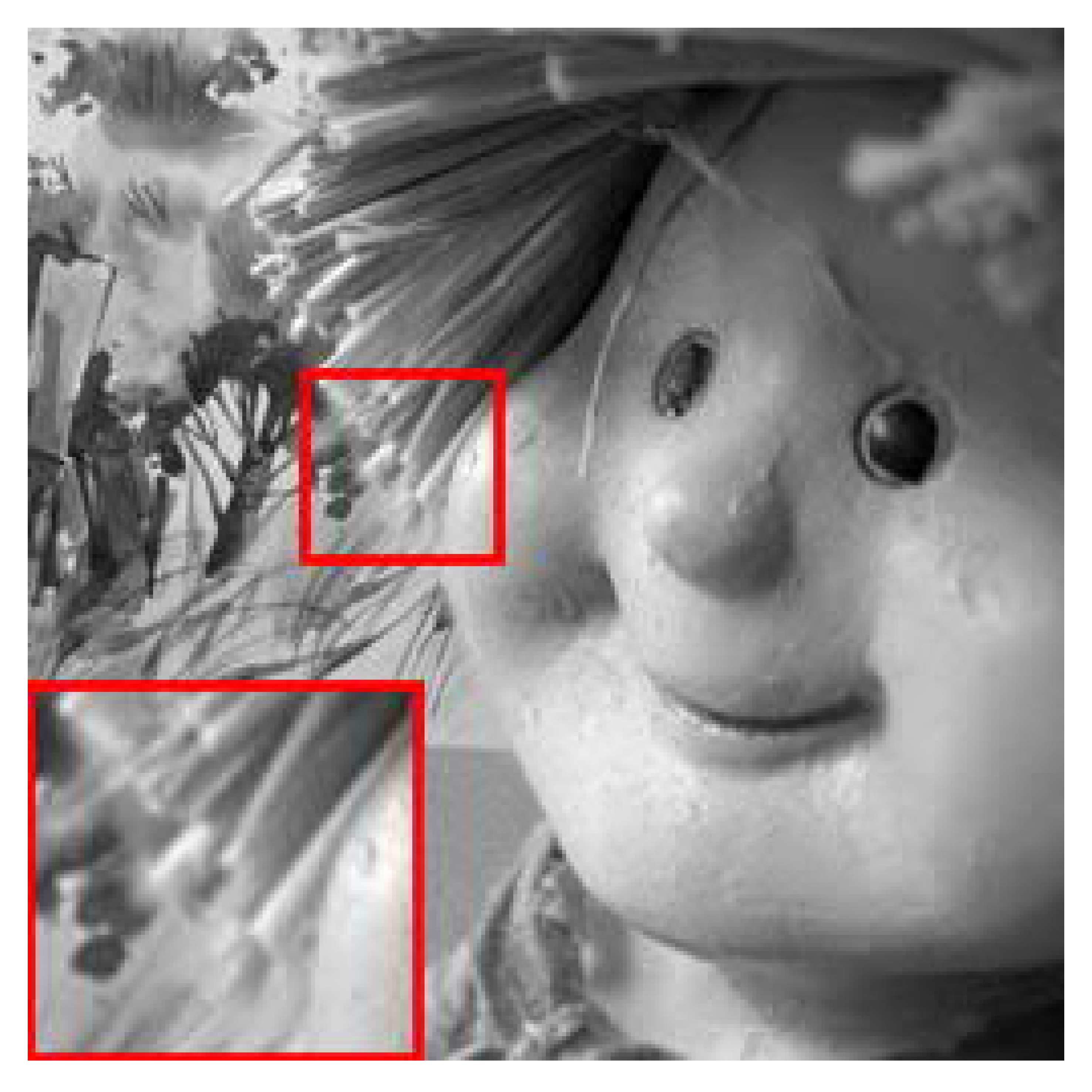}}	
	\end{minipage}%
	\begin{minipage}{0.05\linewidth}
		\centerline{h}	
	\end{minipage}%
	\begin{minipage}{0.2\linewidth}
		\centerline{\includegraphics[width=4cm,height=4cm]{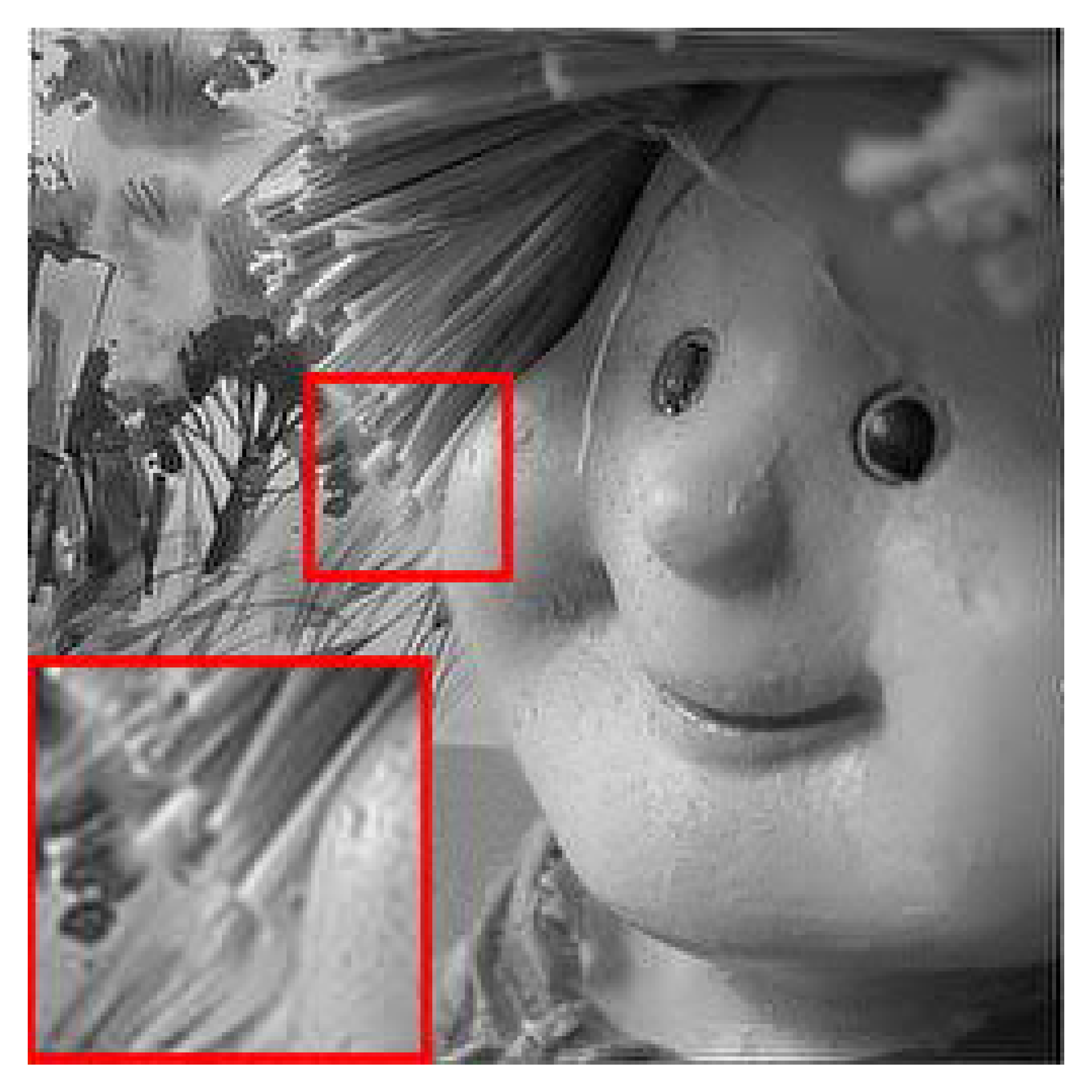}}
	\end{minipage}%
	\vfill
	\centering
	\caption{Fusion result for multi-focus images ''Doll'', the same order as Fig.~\ref{fig:clocks}}
	\label{fig:doll}
	\vspace*{-5pt}
\end{figure*}
\begin{figure*}[!htb]
	\begin{minipage}{0.05\linewidth}
		\centerline{a}	
	\end{minipage}%
	\begin{minipage}{0.2\linewidth}
		\centerline{\includegraphics[width=4cm,height=4cm]{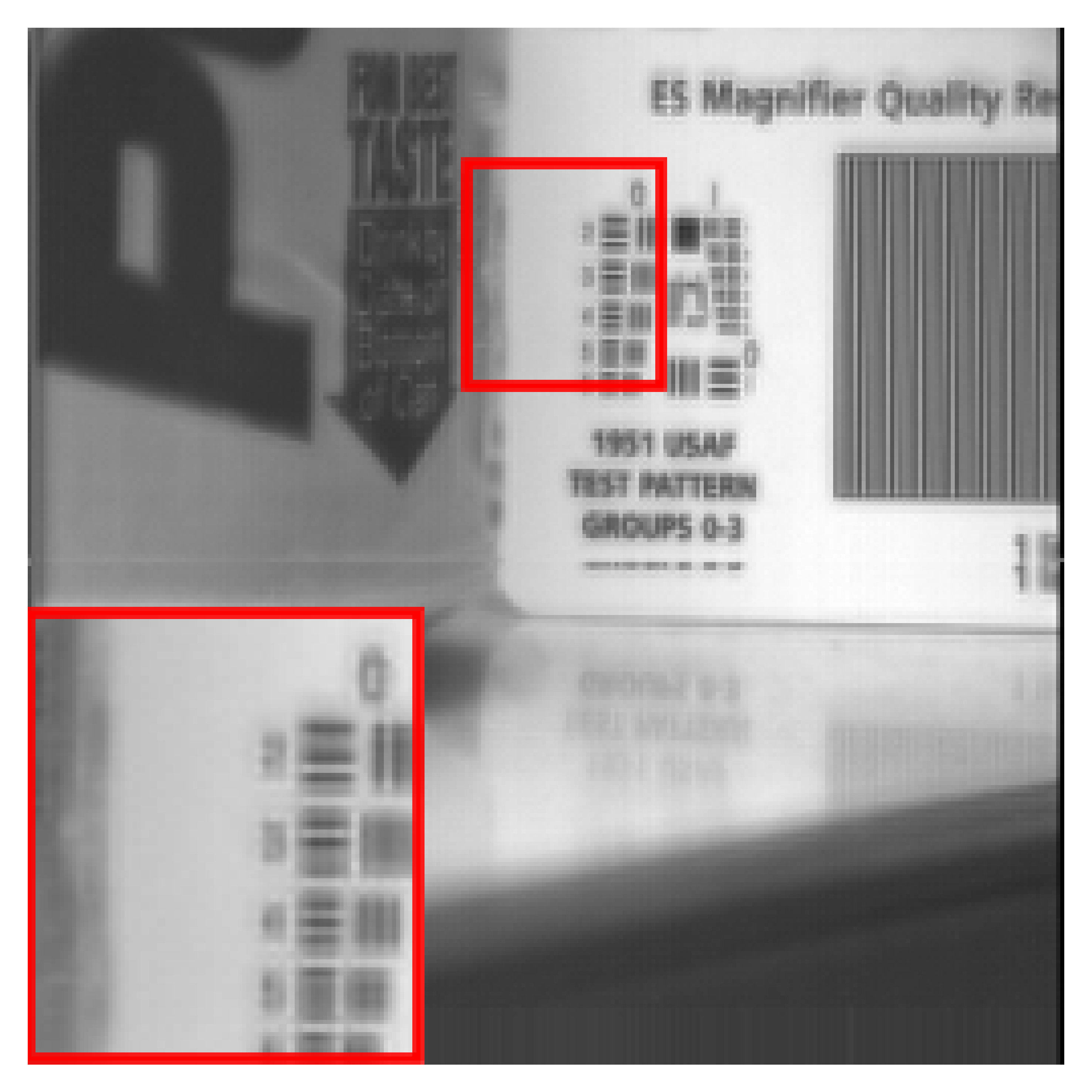}}
	\end{minipage}%
	\begin{minipage}{0.05\linewidth}
		\centerline{b}	
	\end{minipage}%
	\begin{minipage}{0.2\linewidth}
		\centerline{\includegraphics[width=4cm,height=4cm]{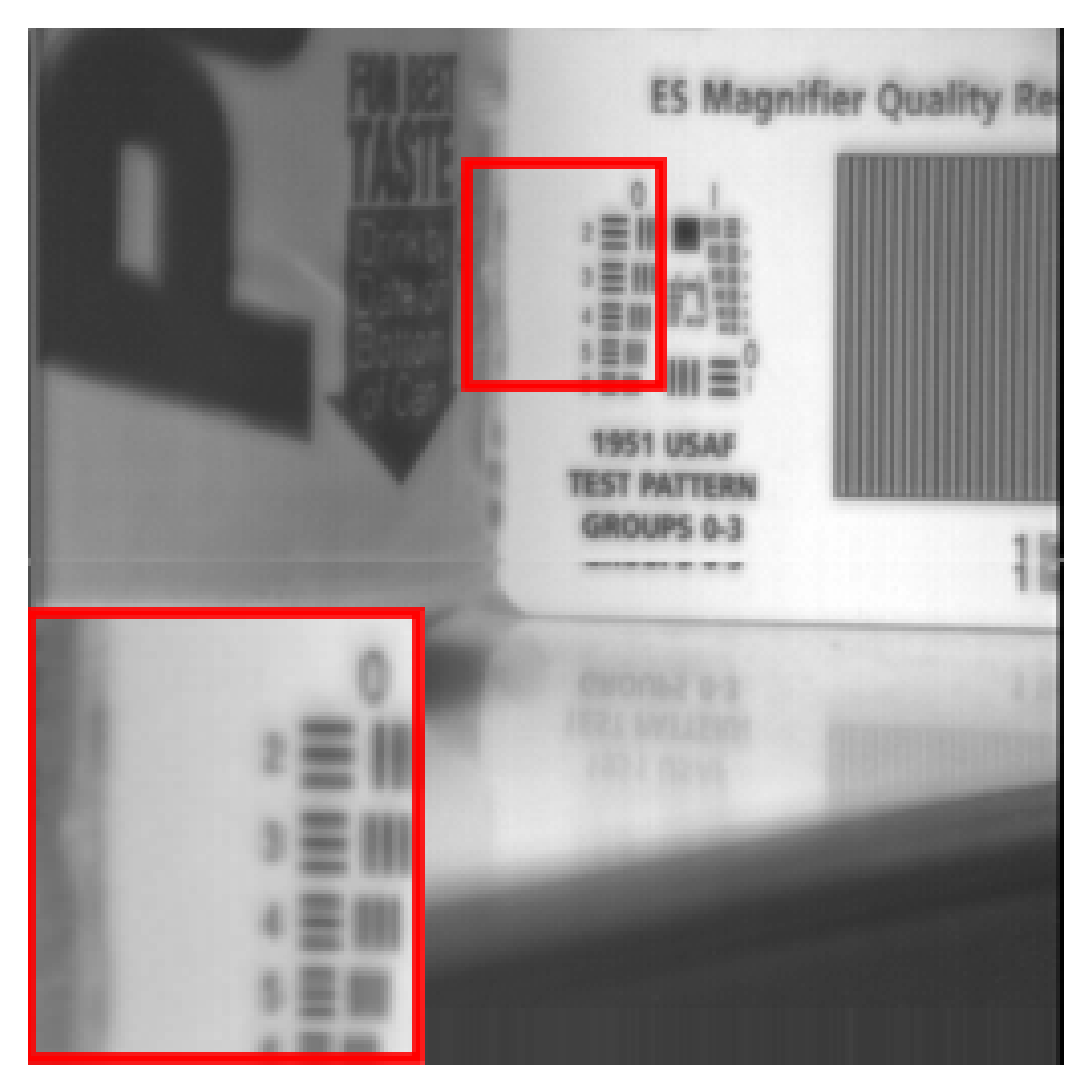}}
	\end{minipage}%
	\begin{minipage}{0.05\linewidth}
		\centerline{c}	
	\end{minipage}%
	\begin{minipage}{0.2\linewidth}
		\centerline{\includegraphics[width=4cm,height=4cm]{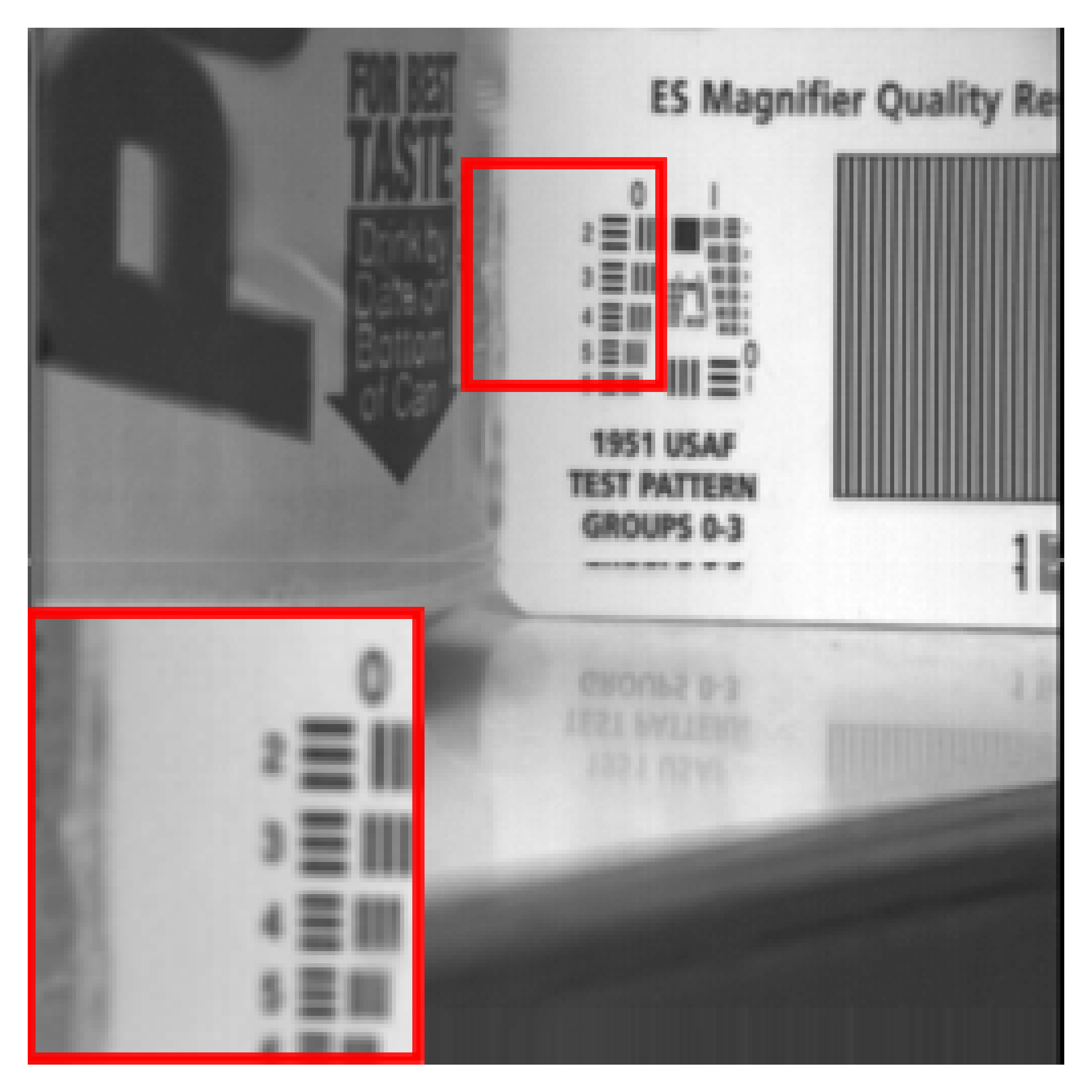}}
	\end{minipage}%
	\begin{minipage}{0.05\linewidth}
		\centerline{d}	
	\end{minipage}%
	\begin{minipage}{0.2\linewidth}
		\centerline{\includegraphics[width=4cm,height=4cm]{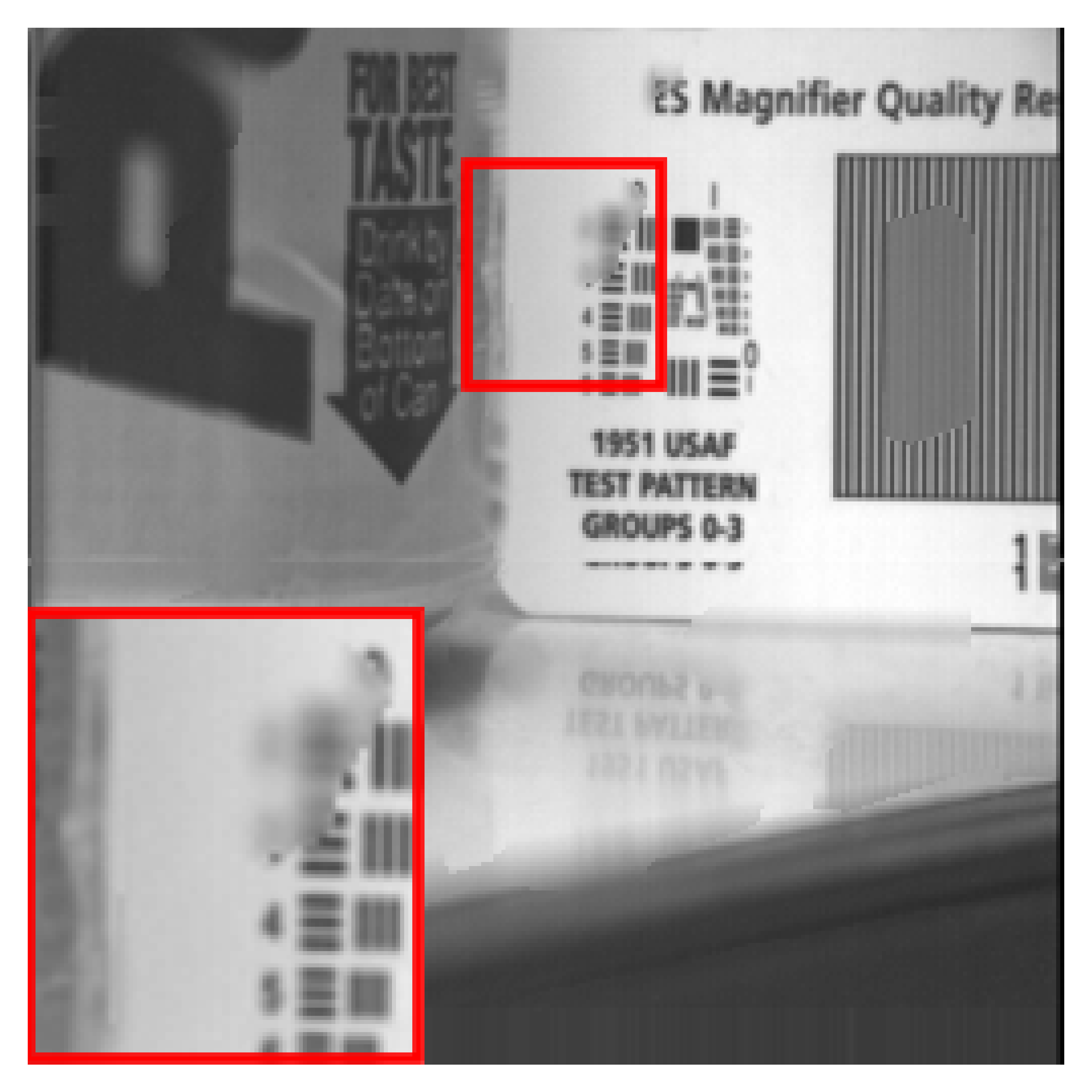}}
	\end{minipage}%
	\vfill
	\begin{minipage}{0.05\linewidth}
		\centerline{e}	
	\end{minipage}%
	\begin{minipage}{0.2\linewidth}
		\centerline{\includegraphics[width=4cm,height=4cm]{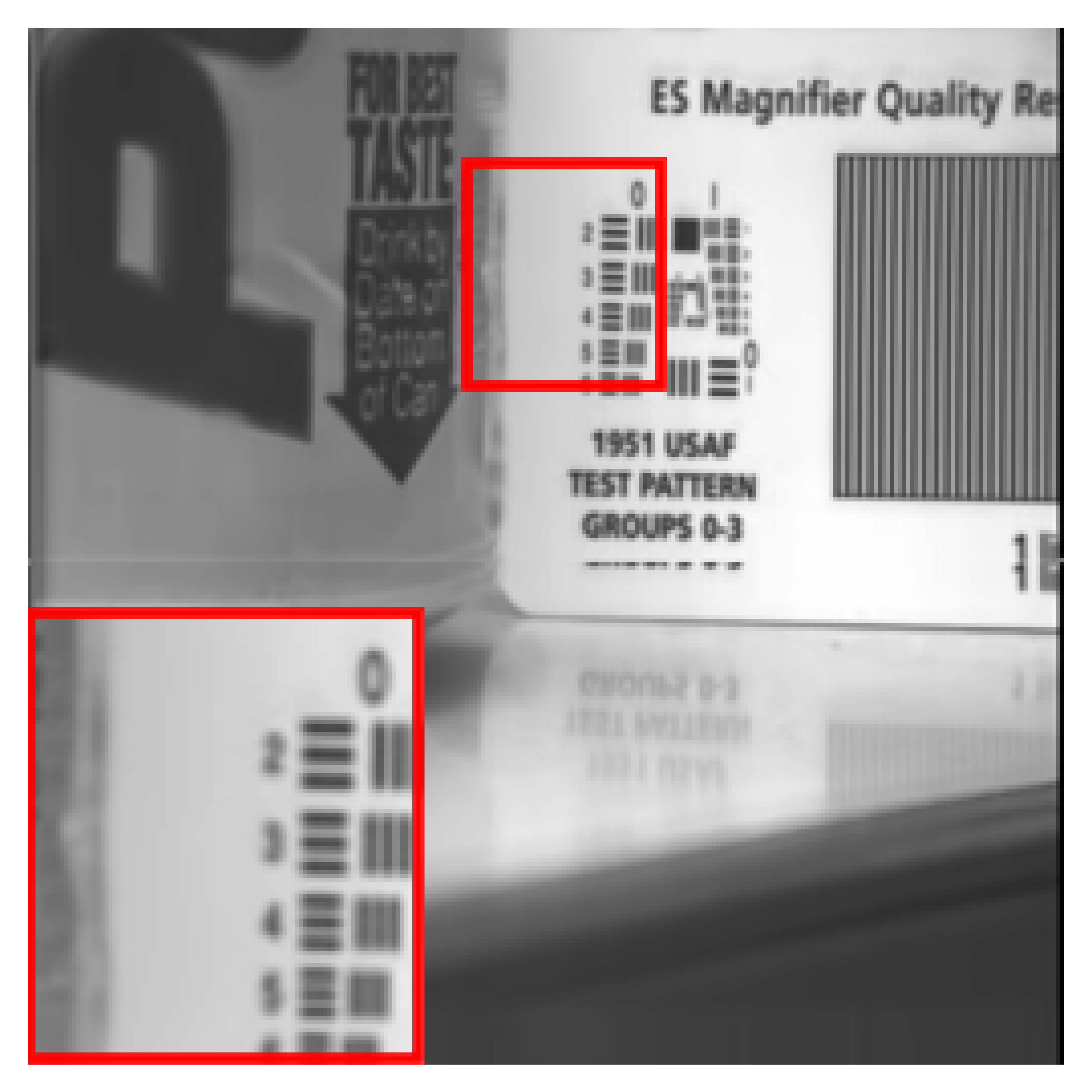}}
	\end{minipage}%
	\begin{minipage}{0.05\linewidth}
		\centerline{f}	
	\end{minipage}%
	\begin{minipage}{0.2\linewidth}
		\centerline{\includegraphics[width=4cm,height=4cm]{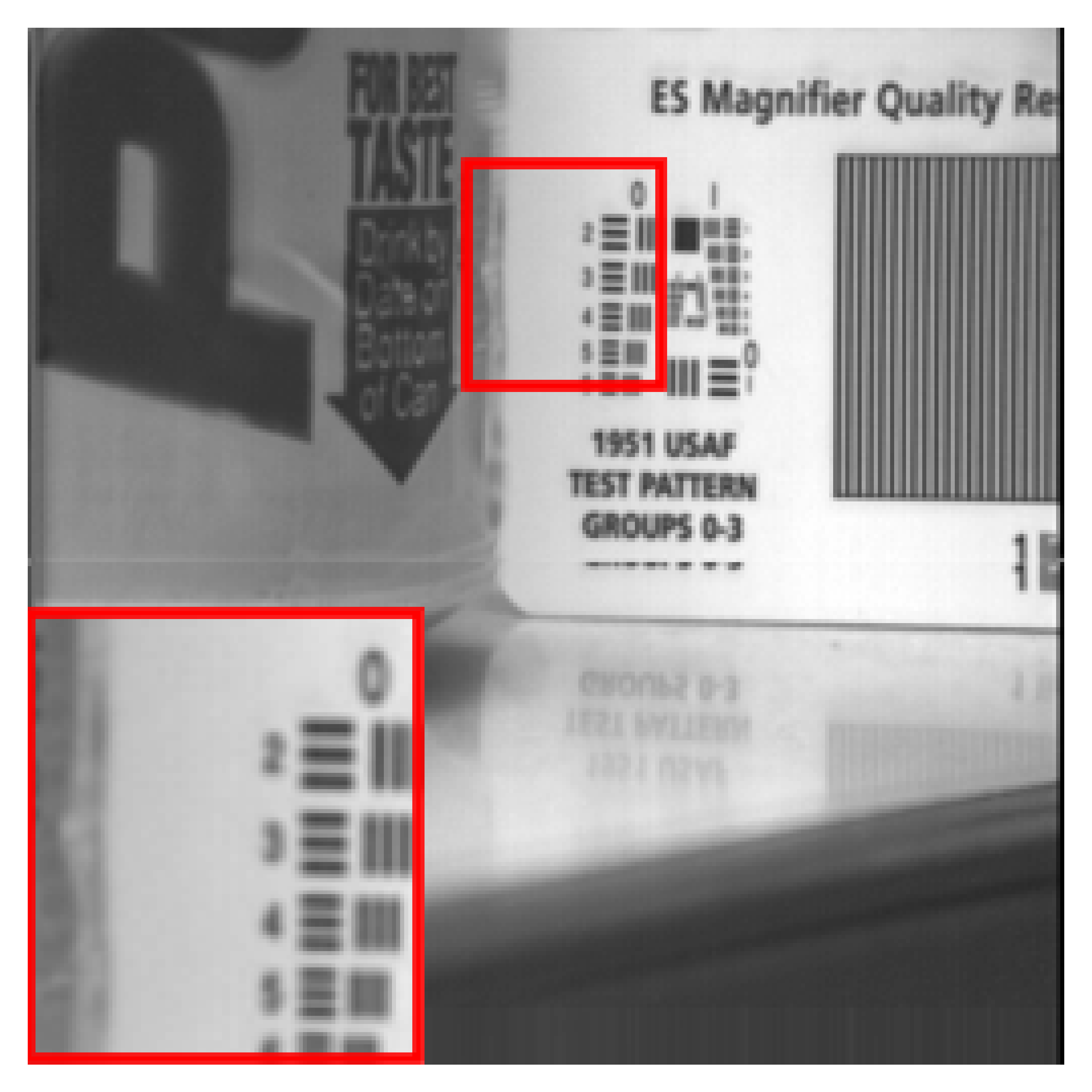}}	
	\end{minipage}%
	\begin{minipage}{0.05\linewidth}
		\centerline{g}	
	\end{minipage}%
	\begin{minipage}{0.2\linewidth}
		\centerline{\includegraphics[width=4cm,height=4cm]{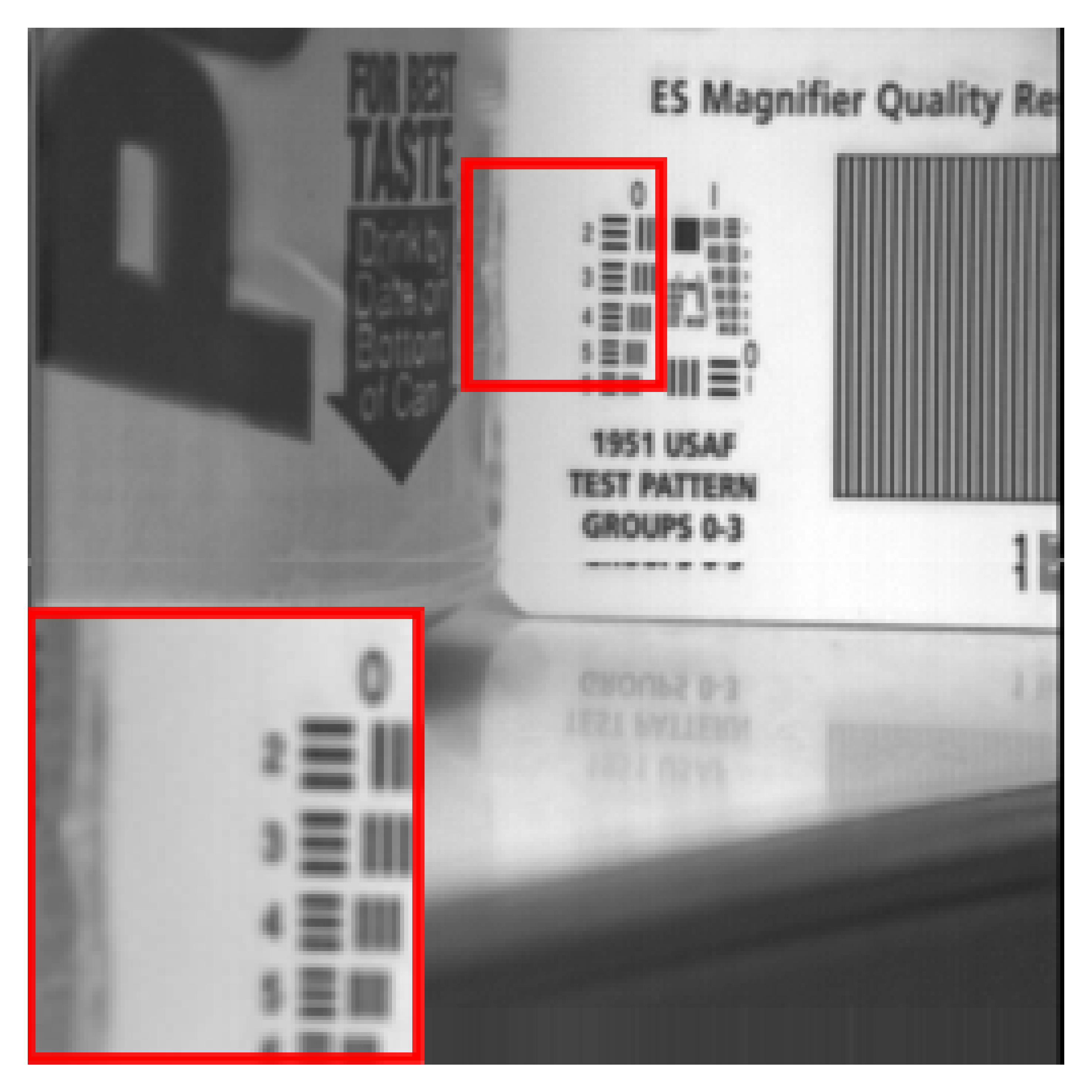}}	
	\end{minipage}%
	\begin{minipage}{0.05\linewidth}
		\centerline{h}	
	\end{minipage}%
	\begin{minipage}{0.2\linewidth}
		\centerline{\includegraphics[width=4cm,height=4cm]{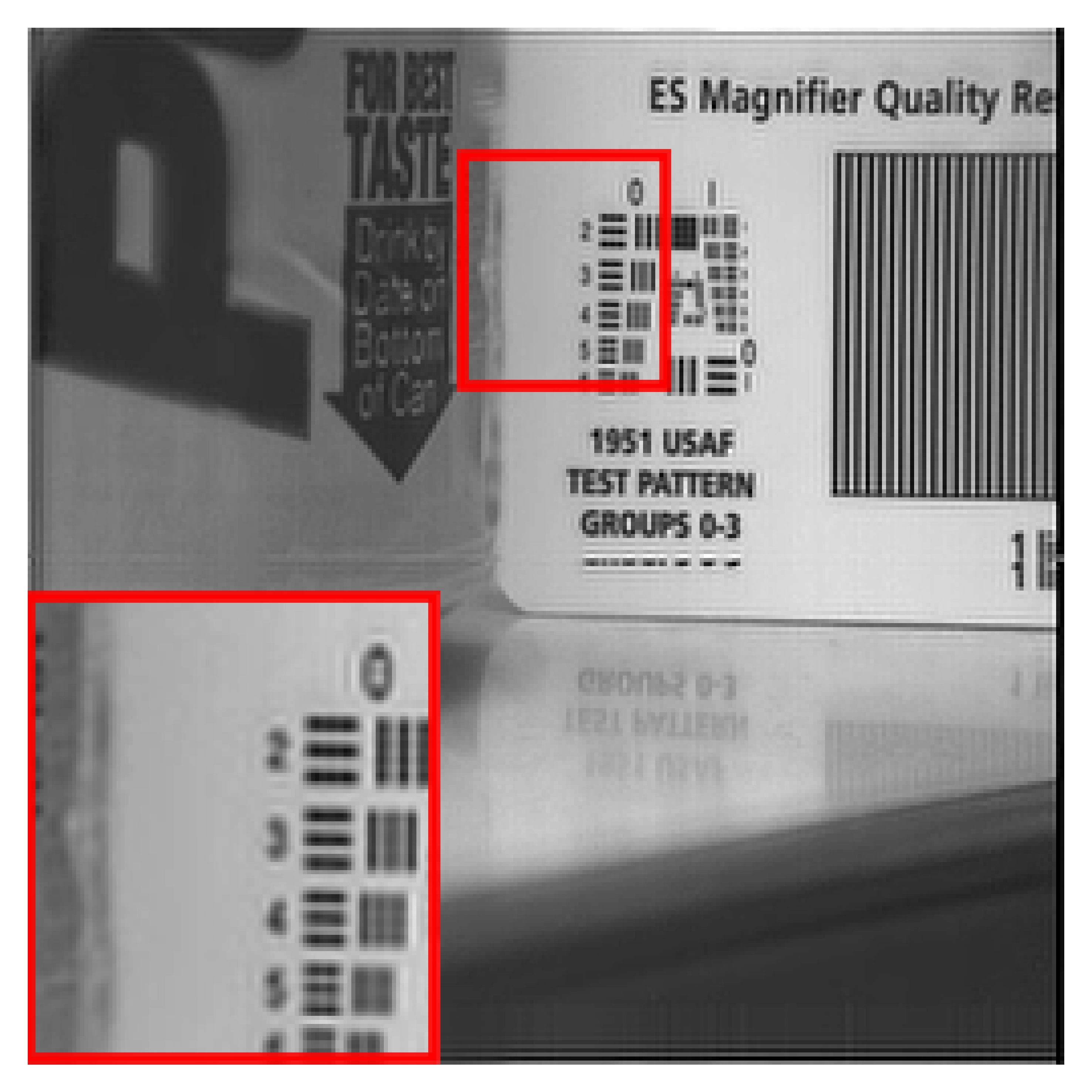}}
	\end{minipage}%
	\vfill
	\centering
	\caption{Fusion result for multi-focus images ''Pepsi'', the same order as Fig.~\ref{fig:clocks}}
	\label{fig:pepsi}
	\vspace*{-5pt}
\end{figure*}

Note that when $\boldsymbol{D}^{\rm F}$ and $\boldsymbol{D}^{\rm B}$ are learned separately, in accordance with max-$l_1$-norm rule~\cite{16} the sparse representations of blurred data $\boldsymbol{X}^{\rm B}$ is sparser, so it contains larger amplitude (but fewer number) non-zero entries (size of focused/blurred patches are normalized). Thus, the correlations between the atoms of $\boldsymbol{D}^{\rm B}$ and their corresponding blurred features are larger than those between the atoms of $\boldsymbol{D}^{\rm F}$ and their corresponding focused features.

To empirically demonstrate the effectiveness of using coupled dictionary learning instead of a single dictionary, we show in Fig.~\ref{fig:comparedic}, the  NMI and $Q_{A\!B/\!F}$ results of fusion using coupled and separately learned dictionaries for all images in the gray-scale dataset. The figure clearly shows that the results obtained using coupled dictionary are superior in all cases.

Moreover, the masks obtained using the proposed and SR-KSVD methods are compared in Fig.~\ref{fig:maskcompare}. It can be seen in the mask resulted by SR-KSVD method (see Fig.~\ref{fig:maskcompare}.(c)) that the excessive bias in selecting patches with the largest $l_1$-norm as the focused one leads to wrong decisions around edges where sub-blurred and sub-focused patches need to be fused. That is because using one dictionary that only represents focused features sets limit on the sparsity of representations of sub-blurred patches, and lead to larger $l_1$-norm for those sparse representations.

However, the proposed method approximates all patches over both the focused and blurred feature spaces ($\boldsymbol{D}^{\rm F}$ and $\boldsymbol{D}^{\rm B}$), then using a weighted $l_1$-norm, it finds the image patch with the highest contribution from the focused features (the atoms of $\boldsymbol{D}^{\rm F}$) in its sparse approximation as the most focused. Moreover, by jointly learning $\boldsymbol{D}^{\rm F}$ and $\boldsymbol{D}^{\rm B}$, two balanced models of two feature spaces are obtained. Thus, the accuracy of the fusion operation is improved. It can be seen form the mask obtained by the proposed method (see Fig.~\ref{fig:maskcompare}.(d)) that the excessive error has been reduced to a high degree. 

\subsection{Comparison Results}
The proposed method can be used for fusion of multifocus image sets $\left\{\boldsymbol{I}_k\right\}_{k=1}^K$,  where $K$ can be any number. The fusion rule~(\ref{eq014}) will take one patch as the most focused at each position to reconstruct the all-in-focus image. Here, the experiments are performed for double and triple input series ($K=2$ and $3$). For color images, the mask is obtained for gray-scale version of the input images and then it is used for fusion of each of the three layers in RGB (red, green and blue) format. 
\begin{figure}[!htb]
	\begin{center}
		\begin{minipage}{0.5\linewidth}
			\centerline{(a)}
			\centerline{\includegraphics[width=4.5cm,height=9cm]{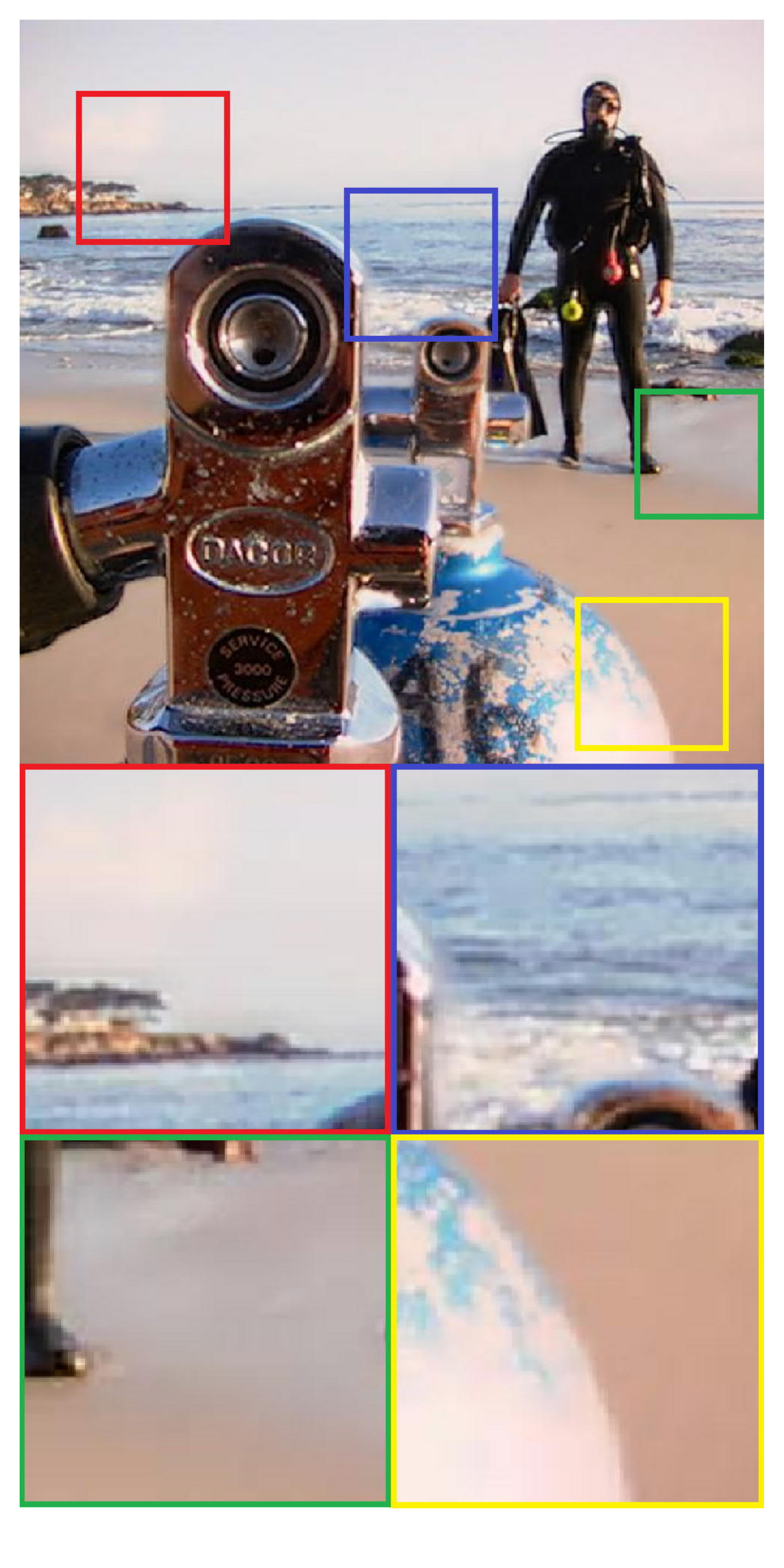}}
		\end{minipage}%
		\begin{minipage}{0.5\linewidth}
			\centerline{(b)}
			\centerline{\includegraphics[width=4.5cm,height=9cm]{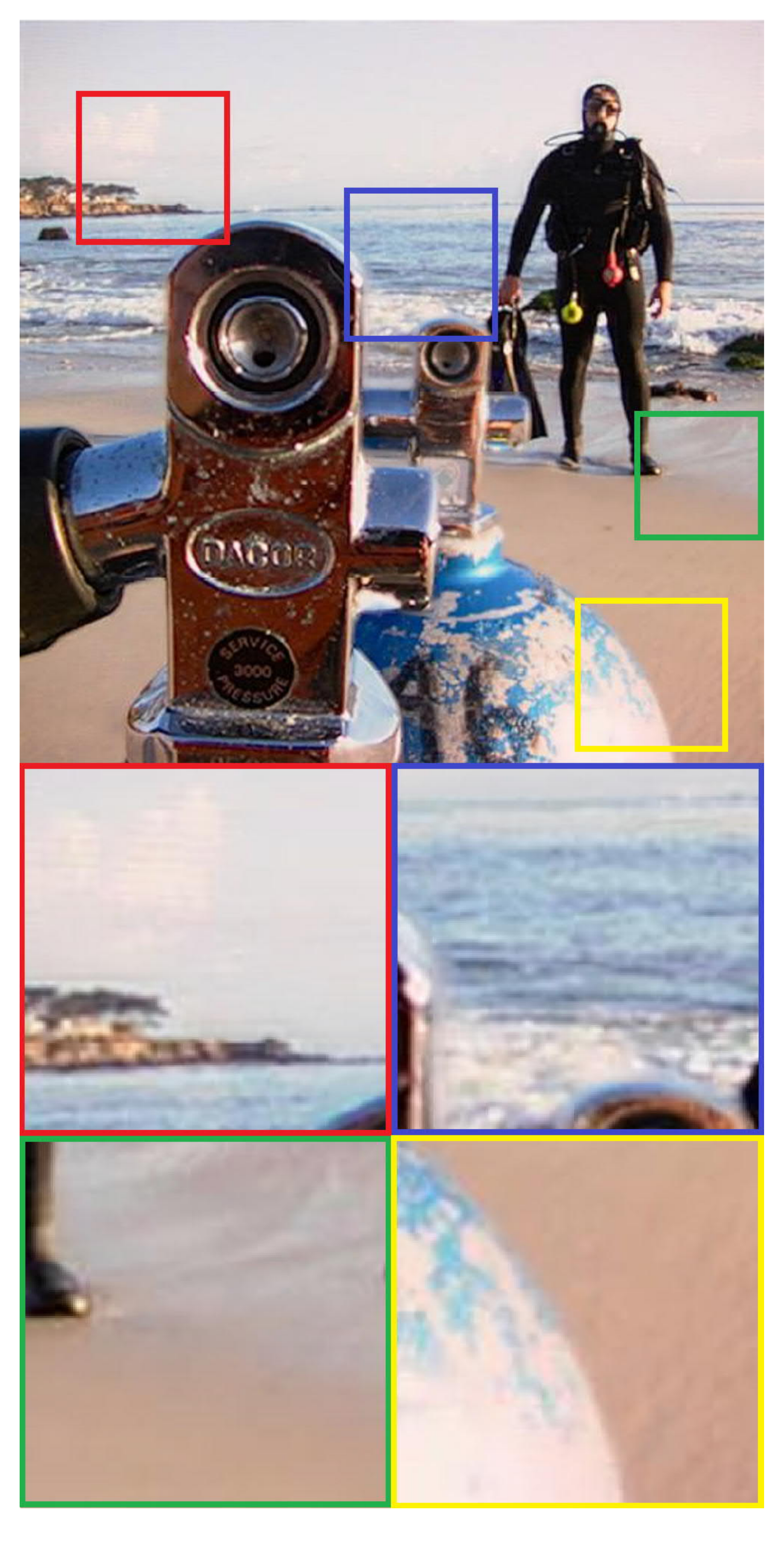}}
		\end{minipage}%
		\centering{
			\caption{Fusion result for multi-focus images ''Diver'': (a) SR-KSVD, (b) proposed.}	\label{fig:triple1}}
	\end{center}
\end{figure}

\begin{figure}[!htb]
	\begin{center}
		\begin{minipage}{0.5\linewidth}
			\centerline{(a)}
			\centerline{\includegraphics[width=4.5cm,height=9cm]{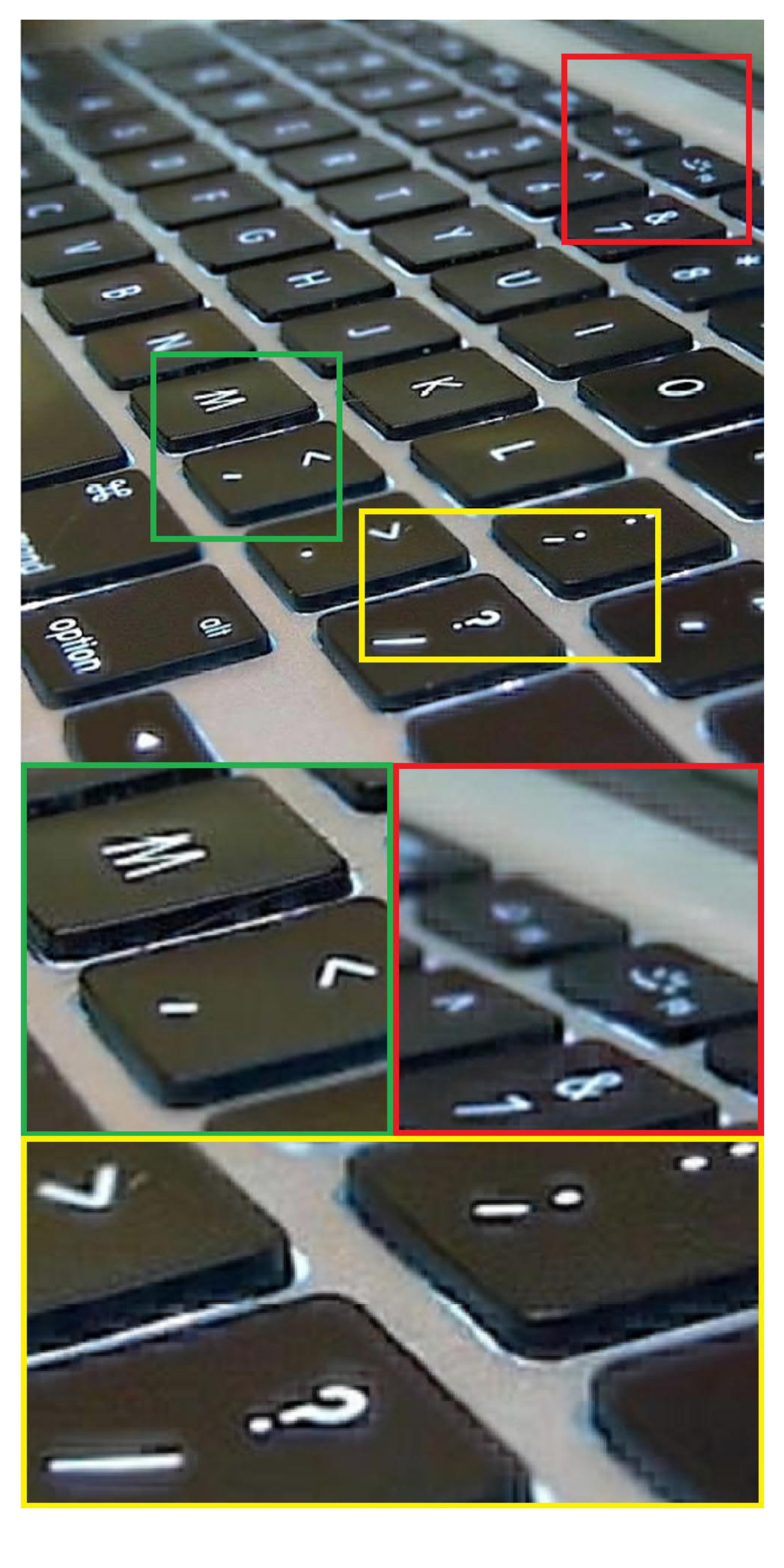}}
		\end{minipage}%
		\begin{minipage}{0.5\linewidth}
			\centerline{(b)}
			\centerline{\includegraphics[width=4.5cm,height=9cm]{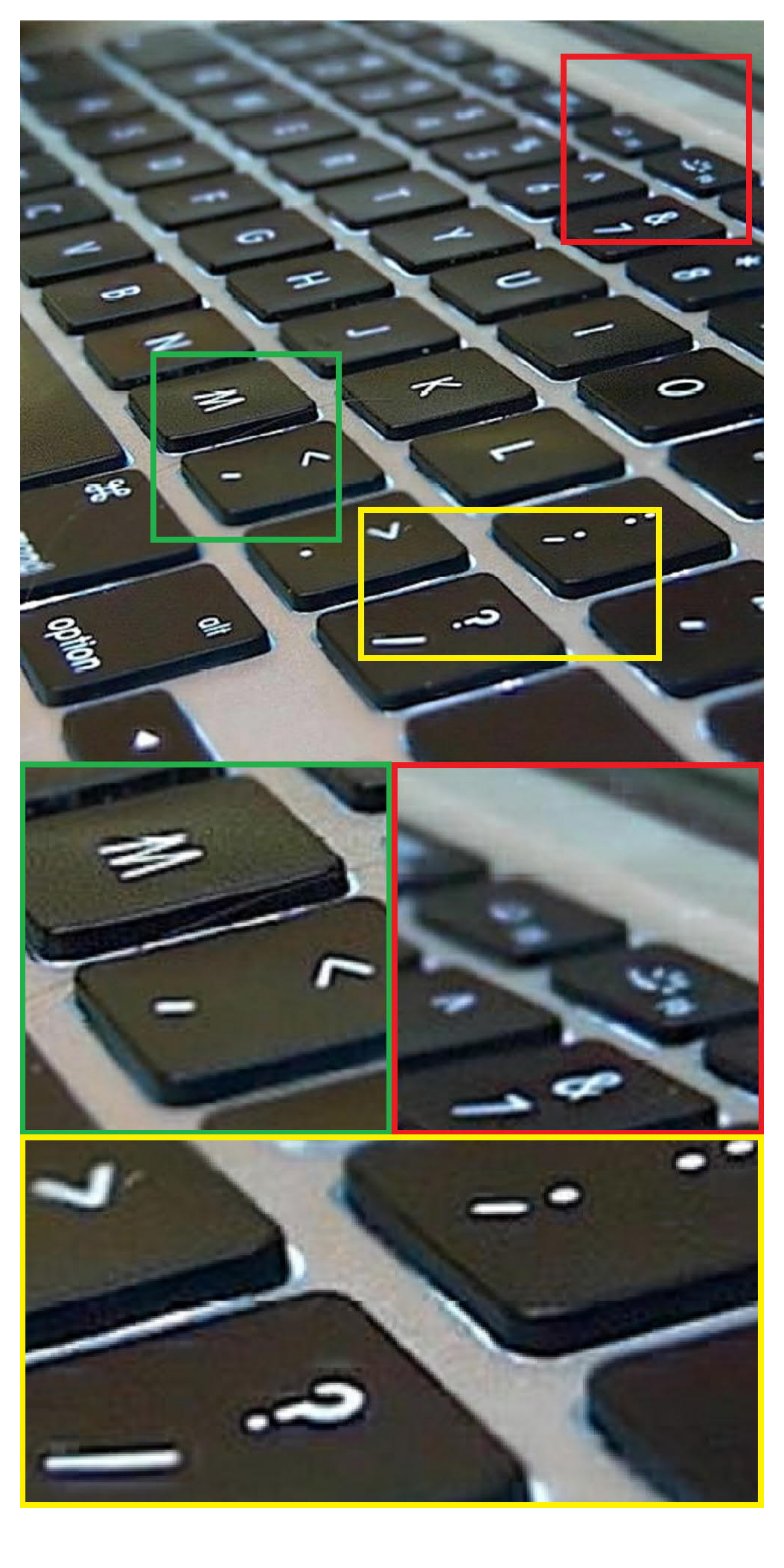}}
		\end{minipage}%
		\centering{
			\caption{Fusion result for multi-focus images ''Keyboard'': (a) SR-KSVD, (b) proposed.}	\label{fig:triple2}}
	\end{center}
\end{figure}

\begin{table}[htb]
	\begin{center}
		
		\begin{tabularx}{15.5cm}{|p{1.5cm}<{\centering}|p{1.5cm}<{\centering}|p{1.5cm}<{\centering}|p{1.5cm}<{\centering}|p{1.5cm}<{\centering}|p{1.5cm}<{\centering}|p{1.5cm}<{\centering}|p{1.5cm}<{\centering}|}
			\hline
			\multirow{1}{*}{\textbf{\tabincell{c} {Methods}}}&\textbf{Measures}&\textbf{Clocks}&\textbf{Lab}& \textbf{Pepsi} 
			& \textbf{Disk}& \textbf{Jug}& \textbf{Doll}\\ 
			\hline
			
			\multirow{4}{*}{\textbf{\tabincell{c} {DWT~\cite{34}}}} 
			&\textbf{$\!N\!M\!I$}&0.9847& 1.0027 & 1.0079 
			& 0.8129 & 0.8497 & 0.8553\\[-2pt] 
			&\textbf{$Q_{A\!B\!/\!F}$}&0.6600& 0.5487 & 0.6587
			& 0.5102 & 0.5048 & 0.6184\\[-2pt]   
			&\textbf{$S\!S\!I\!M$}&0.9403& \textbf{\textcolor{blue}{0.9372}} & \textbf{\textcolor{blue}{0.9362}} 
			& \textbf{\textcolor{blue}{0.9068}} & 0.8871 & 0.9211\\[-2pt]   
			&\textbf{$\!M\!S\!E$}&32.2172& 60.1514 & 43.2703 
			& 94.3113 & 51.6737 & 46.3669\\[-2pt]   
			\hline
			
			\multirow{4}{*}{\textbf{\tabincell{c} {PCA~\cite{40}}}} 
			&\textbf{$\!N\!M\!I$}&1.0276& 1.0270 & 1.0610
			& 0.8372 & 0.8854 & 0.8965\\[-2pt]   
			&\textbf{$Q_{A\!B\!/\!F}$}&0.6939& 0.5651 & 0.6752 
			& 0.5352 & 0.5083 & 0.6355\\[-2pt]  
			&\textbf{$S\!S\!I\!M$}&\textbf{\textcolor{red}{0.9572}}& \textbf{\textcolor{red}{0.9468}} & \textbf{\textcolor{green}{0.9351}} 
			& \textbf{\textcolor{red}{0.9226}} & 0.9048 & 0.9418\\[-2pt]  
			&\textbf{$\!M\!S\!E$}&24.8221& 54.4139 & 29.4576
			& 80.9688 & 45.1700 & 36.2968\\[-2pt] 
			\hline
			
			\multirow{4}{*}{\textbf{\tabincell{c} {DSIFT~\cite{38}}}} 
			&\textbf{$\!N\!M\!I$}&1.0015& 1.0726 &0.9943
			& 0.9217 & 0.8836 & 0.8921\\[-2pt]   
			&\textbf{$Q_{A\!B\!/\!F}$}&0.7017& \textbf{\textcolor{green}{0.7269}} & 0.7243 
			& 0.6938 & \textbf{\textcolor{green}{0.7714}} &\textbf{\textcolor{green}{ 0.7416}}\\[-2pt]   
			&\textbf{$S\!S\!I\!M$}&0.8649& \textbf{\textcolor{green}{0.9003}} & 0.8955 
			& 0.8779 & \textbf{\textcolor{green}{0.9661}} & 0.9817\\[-2pt]   
			&\textbf{$\!M\!S\!E$}&32.1295& 17.55 & 19.1050
			& 45.9702 & 4.3322 & 6.7304\\[-2pt] 
			\hline

			\multirow{4}{*}{\textbf{\tabincell{c} {SR-FM~\cite{17}}}} 
			&\textbf{$\!N\!M\!I$}&1.1100& 1.0573 & \textbf{\textcolor{blue}{1.1764}} 
			& 0.8878 & 0.9490 & \textbf{\textcolor{blue}{1.0935}}\\[-2pt]   
			&\textbf{$Q_{A\!B\!/\!F}$}&\textbf{\textcolor{green}{0.7462}}& 0.6900 & \textbf{\textcolor{green}{0.7577}}
			& 0.6380 & 0.7174 & 0.7380\\[-2pt]   
			&\textbf{$S\!S\!I\!M$}&0.9451& 0.8153 & 0.9296
			& 0.8325 & 0.9490 & \textbf{\textcolor{green}{0.9862}}\\[-2pt]   
			&\textbf{$\!M\!S\!E$}&5.5989& 12.0835 & 5.8016 
			& 30.7394 & 19.3786 & 7.4314\\[-2pt]  
			\hline
			
			\multirow{4}{*}{\textbf{\tabincell{c} {SR-CM~\cite{16}}}} 
			&\textbf{$\!N\!M\!I$}&\textbf{\textcolor{green}{1.1188}}& \textbf{\textcolor{green}{1.1079}} & 1.1063 
			& \textbf{\textcolor{green}{0.9460}} & \textbf{\textcolor{green}{1.0630}} & \textbf{\textcolor{green}{1.0547}}\\[-2pt]   
			&\textbf{$Q_{A\!B\!/\!F}$}&0.7301& 0.7058 & 0.7290 
			& \textbf{\textcolor{green}{0.7052}} & 0.7656 & 0.7402\\[-2pt]   
			&\textbf{$S\!S\!I\!M$}&0.8813& 0.7843 & 0.8229 
			& 0.8367 & 0.9609 & 0.9817\\[-2pt]   
			&\textbf{$\!M\!S\!E$}&\textbf{\textcolor{green}{1.8879}}& \textbf{\textcolor{blue}{7.4700}} & \textbf{\textcolor{green}{3.9962}}
			& \textbf{\textcolor{blue}{11.0090}} & \textbf{\textcolor{green}{3.3700}} & \textbf{\textcolor{green}{3.3617}}\\[-2pt]   
			\hline

			\multirow{4}{*}{\textbf{\tabincell{c} {SR-KSVD}}} 
			&\textbf{$\!N\!M\!I$}&\textbf{\textcolor{blue}{1.1658}}& \textbf{\textcolor{blue}{1.1235}} & \textbf{\textcolor{green}{1.1685}} 
			& \textbf{\textcolor{blue}{0.9821}} & \textbf{\textcolor{blue}{1.1417}} & 1.0517\\[-2pt]   
			&\textbf{$Q_{A\!B\!/\!F}$}&\textbf{\textcolor{blue}{0.7557}}& \textbf{\textcolor{blue}{0.7295}} & \textbf{\textcolor{blue}{0.7613}} 
			& \textbf{\textcolor{blue}{0.7206}} & \textbf{\textcolor{blue}{0.7766}} & \textbf{\textcolor{blue}{0.7454}}\\[-2pt]   
			&\textbf{$S\!S\!I\!M$}&\textbf{\textcolor{green}{0.9527}}& 0.8400 & 0.9258 
			& 0.8667 & \textbf{\textcolor{blue}{0.9925}} & \textbf{\textcolor{blue}{0.9888}}\\[-2pt]   
			&\textbf{$\!M\!S\!E$}&\textbf{\textcolor{blue}{1.6457}}& \textbf{\textcolor{green}{7.7026}} & \textbf{\textcolor{blue}{3.6903}}
			& \textbf{\textcolor{green}{11.0777}} & \textbf{\textcolor{blue}{3.2798}} & \textbf{\textcolor{blue}{3.4843}}\\[-2pt]  
			\hline

			\multirow{4}{*}{\textbf{\tabincell{c} {Proposed}}} 
			&\textbf{$\!N\!M\!I$}&\textbf{\textcolor{red}{1.1833}} & \textbf{\textcolor{red}{1.1733}} & \textbf{\textcolor{red}{1.1803}} 
			& \textbf{\textcolor{red}{1.0176}} & \textbf{\textcolor{red}{1.1513}} & \textbf{\textcolor{red}{1.1475}}\\[-2pt]   
			&\textbf{$Q_{A\!B\!/\!F}$}&\textbf{\textcolor{red}{0.7578}} & \textbf{\textcolor{red}{0.7340}} & \textbf{\textcolor{red}{0.7678}} 
			& \textbf{\textcolor{red}{0.7247}} & \textbf{\textcolor{red}{0.7786}} & \textbf{\textcolor{red}{0.7490}}\\[-2pt]   
			&\textbf{$S\!S\!I\!M$}&\textbf{\textcolor{blue}{0.9565}} & 0.8214 & \textbf{\textcolor{red}{0.9464}} 
			& \textbf{\textcolor{green}{0.8716}} & \textbf{\textcolor{red}{0.9926}} & \textbf{\textcolor{red}{0.9928}}\\[-2pt]   
			&\textbf{$\!M\!S\!E$}&\textbf{\textcolor{red}{1.3048}} & \textbf{\textcolor{red}{5.6800}} & \textbf{\textcolor{red}{2.8415}} 
			& \textbf{\textcolor{red}{6.3516}} & \textbf{\textcolor{red}{1.7500}} & \textbf{\textcolor{red}{2.9864}}\\[-2pt]  
			\hline  						
		\end{tabularx}
		\caption{Objective evaluation of fusion performance for input dataset 1. Results are ranked by colors as follows. Red is the best, blue is the second best, and green, the third best.}\label{table:1}
	\end{center}
\end{table}

\begin{table}[htb]
	\begin{center}
		
		\begin{tabularx}{8.63cm}{|p{1.2cm}<{\centering}|p{1.1cm}<{\centering}|p{0.8cm}<{\centering}|p{1.1cm}<{\centering}|p{1cm}<{\centering}|p{0.8cm}<{\centering}|}
			\hline
			\multirow{1}{*}{\textbf{\tabincell{c} {Methods}}}&\textbf{Measures}&\textbf{Diver}&\textbf{Keyboard}&\textbf{Folders}&\textbf{Seals}\\ 
			\hline
			\multirow{2}{*}{\textbf{\tabincell{c} {SR-KSVD}}} 
			&\textbf{$Q_{A\!B\!/\!F}$}&0.5882&0.5820&0.6123&0.6337\\
			&\textbf{$\!N\!M\!I$}&1.3522&1.0798&1.4685&1.4206\\
			\hline
			\multirow{2}{*}{\textbf{\tabincell{c} {Proposed}}} 
			&\textbf{$Q_{A\!B\!/\!F}$}&\textbf{0.6247}&\textbf{0.6207}&\textbf{0.6544}&\textbf{0.6825}\\
			&\textbf{$\!N\!M\!I$}&\textbf{1.4644}&\textbf{1.1831}&\textbf{1.5984}&\textbf{1.5103}\\
			\hline
		\end{tabularx}
		\caption{Objective evaluation of fusion performance for input dataset 2. Best result are shown in bold.}\label{table:3}
	\end{center}
\end{table}			

\begin{figure}[!htb]
	\centerline{\includegraphics[width=12cm,height=9cm]{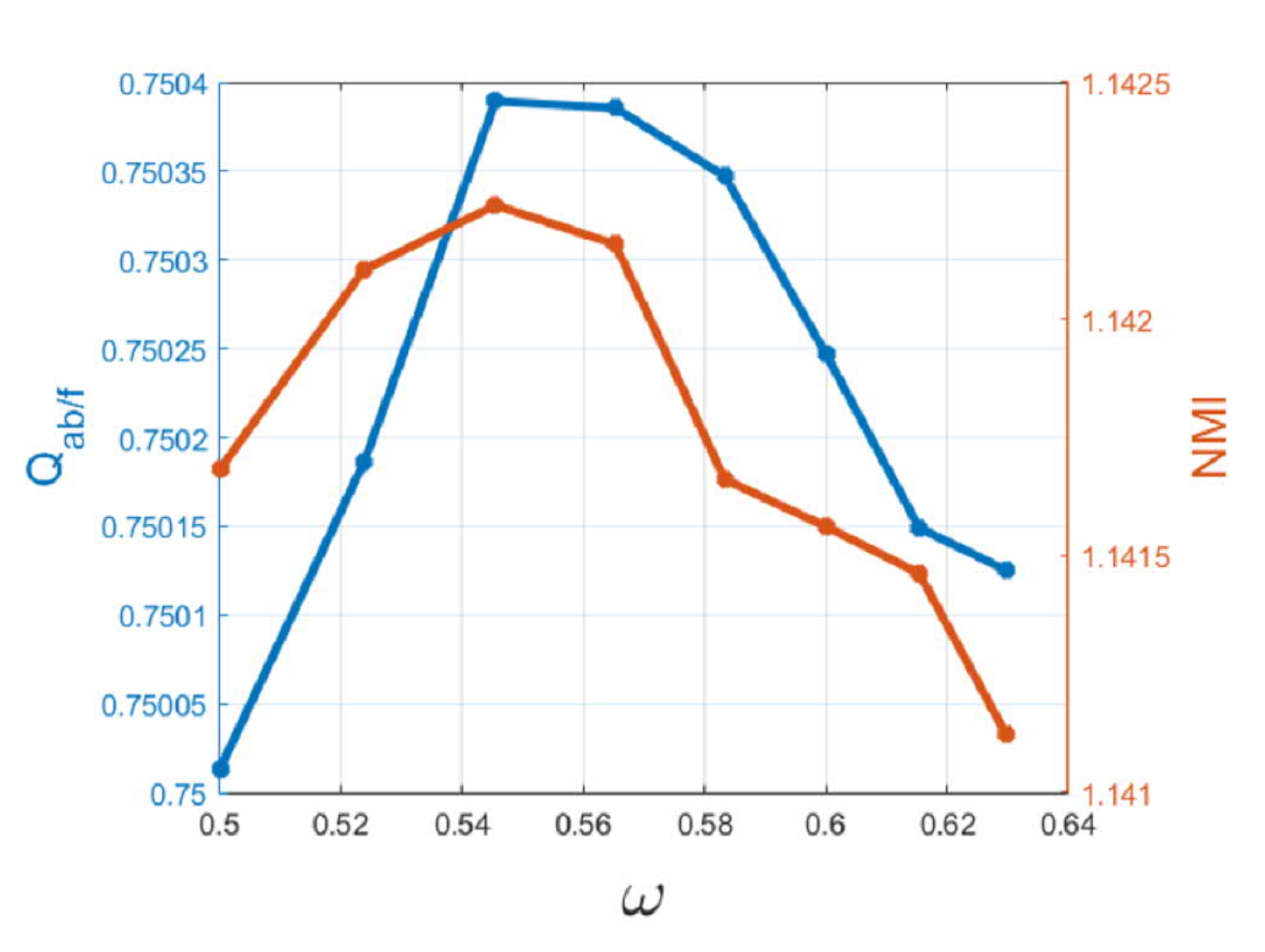}}
	\centering
	\caption{Fusion performance vs. $\omega$ for $\epsilon =0.1$ and $d=8$.}
	\label{fig:gamma}
\end{figure}

\begin{figure}[!htb]
	\centerline{\includegraphics[width=12cm,height=9cm]{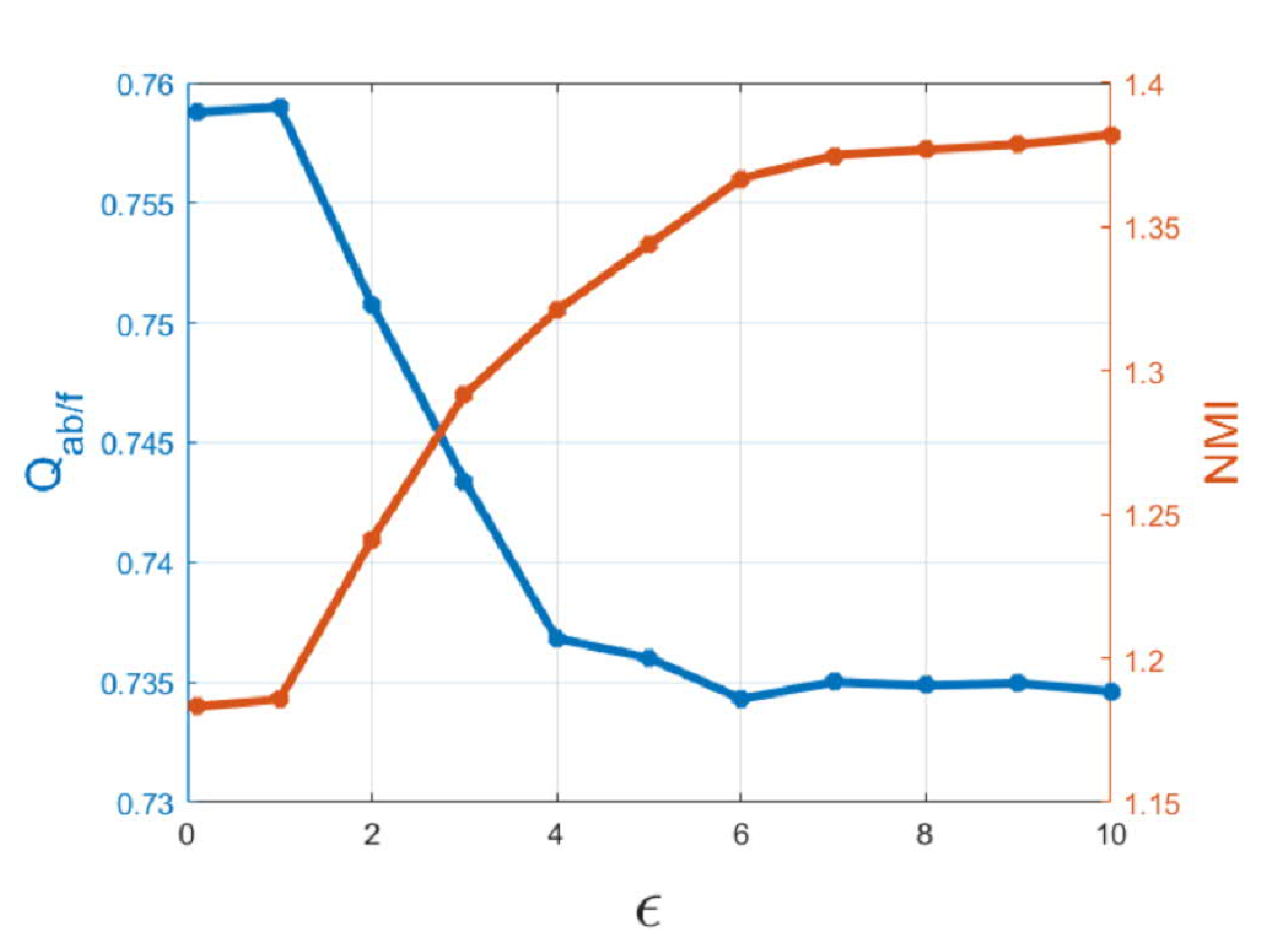}}
	\centering
	\caption{usion performance vs. $\epsilon$ for $d=8$.}
	\label{fig:epsilon}
\end{figure}

\begin{figure}[!htb]
	\centerline{\includegraphics[width=12cm,height=9cm]{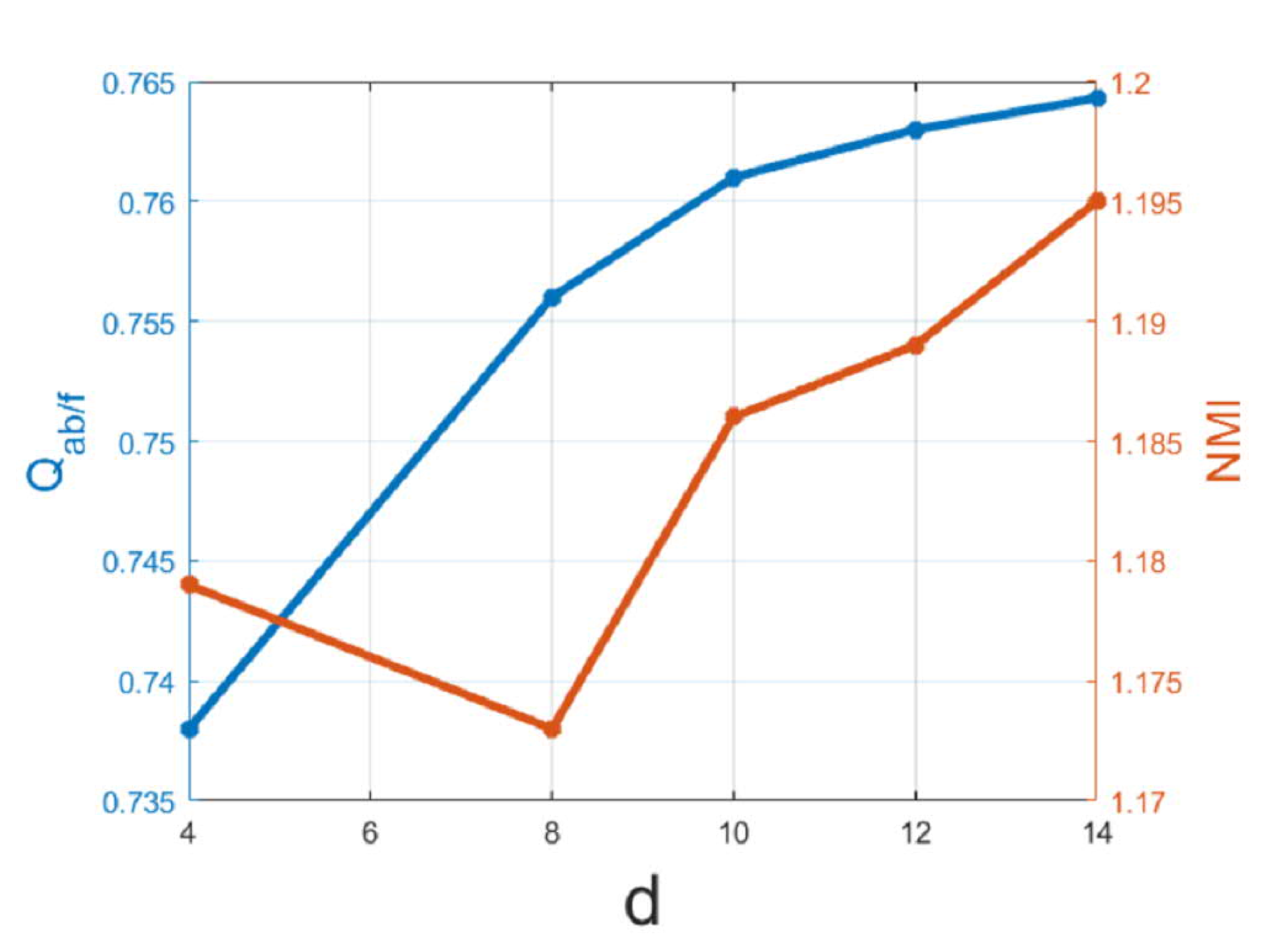}}
	\centering
	\caption{Fusion performance vs. $\epsilon$ for $d=8$.}
	\label{fig:patchSize}
\end{figure}

The representative fusion results for three pairs of images: Clocks, Doll, and Pepsi, and the two triple series: Diver and Keyboard, are shown in Figs.~\ref{fig:clocks}-\ref{fig:triple2}, respectively. All the figures also include the magnified details. 

Visually inspecting the results from gray-scale dataset, we can see that the DWT method results in blocking artifacts, the fusion using PCA method shows excessive blurring effect in all three cases, although the images are relatively smooth. The fused images produced using DSIFT method although have high contrast resolution in all cases, the misaligned decision map results in blocking artifacts that highly reduces the visual quality of the images. These blocking artifacts are less visible in Figs.~\ref{fig:doll} and \ref{fig:pepsi} comparing to Fig.~\ref{fig:clocks}. It can be explained in terms of low robustness of DSIFT method against different levels of blurredness in multi-focus source images. 

The other four methods, namely SR-FM, SR-CM, SR-KSVD, and the proposed method give smooth fused images. Looking more carefully at details and magnified parts (quantitative evaluations are also given later), it can be seen however that the proposed method yields the best results. The effectiveness of the proposed method is more visible in the fusion results for triple series where the inputs are larger and more diverse in terms of focus level. We compared the proposed methods to its closest competitor SR-KSVD. It can clearly be seen from the magnified detailes that the proposed method has a better performance. For example, in the fusion results for the image Diver (see Fig.~\ref{fig:triple1}), the cloud in the red rectangle and the trace of water in the green rectangle, and in the results for Keyboard (see Fig.~\ref{fig:triple2}) the hair strand in the green rectangle are only visible in the results obtained by the proposed method.

It can be also seen in all three visualized gray-scale cases that the image resulted from applying ADMM-based global reconstruction to the fused image is significantly better than the images before applying global reconstruction in terms of contrast resolution and visibility of details.

In addition to the visual comparison, Tables~\ref{table:1} and~\ref{table:3} summarize the quantitative evaluations for the methods tested on the datasets in Figs.~\ref{fig:sourceimages} and~\ref{fig:sourceimages2}. It can be seen from both tables that the proposed approach produces the best quantitative results in term of $Q_{AB/F}$ and NMI in all cases, which means that it reduces the blocking artifacts and artificial distortions and combines a significant edge information into the fused images, while showing the highest fidelity in preserving the pixel intensities of source image. The lowest MSE values obtained in all gray-scale experiments mean that the results are the closest to the reference images, which shows the high accuracy of the proposed selection operator.			

\subsection{Effects of Main Parameters}
\label{Parameters}
The following three main parameters influence the fusion performance: patch size $d^2$, tolerance error $\epsilon$, and weight parameter $\omega$ in \eqref{eq014}. To study the effects of these parameters, we run the proposed method on the whole gray-scale dataset and average the results for NMI and $Q_{AB/F}$. Fig.~\ref{fig:gamma} shows the averaged results for weighting parameter $\omega$. It can be seen that for $\omega$ between 0.54 and 0.56, the best results are achieved. The effect of different tolerance error values $\epsilon$ on fusion performance is shown in Fig.~\ref{fig:epsilon}. As it can be seen, the best $Q_{AB/F}$ results attained for $\epsilon=0.1$, while NMI increases for larger $\epsilon$ values. This observation shows that larger values of $\epsilon$ decrease the scattered error so leads to better NMI results, however it increases the bias (uniformly wrong selected areas in decision map) and reduces the $Q_{AB/F}$. Fig.~\ref{fig:patchSize} shows that with increasing the patch size, the fusion performance slightly improves. However, computation over larger patches increases the run time and computational costs, thus we run all other experiments using $d=8$.

\section{Conclusion}
\label{Conclusion}
We have proposed a fusion algorithm for combing multiple images with different focal settings into one all-in-focus image. We first have formalized the physical process of capturing multi-focus images, and then developed a basic model based on the idea of sparse representation of all-in-focus image using a coupled overcomplete dictionary. 
This approach is straightforwardly extendable to the case of multiple (more than two) dictionaries. Using the coupled dictionary from the focused and blurred feature spaces, we have developed an efficient and accurate fusing approach, and have demonstrated that the proposed approach well preserves the edge and structural information of source images; drastically reduces the blocking artifacts, circle blurring, and artificial distortions; and shows in general better results than the existing fusion methods including state-of-the-art methods.

\section{Acknowledgment}
The authors would like to thank Rui Gao for performing some preliminary studies that finally led to this paper, although the particular results of this paper did not follow from her preliminary studies. 

\bibliographystyle{unsrt}

\end{document}